\documentclass[letterpaper,11pt]{amsart}

\usepackage{enumitem}
\usepackage{caption}
\usepackage[colorlinks]{hyperref}
\usepackage[utf8]{inputenc}
\usepackage{amssymb}
\usepackage{amsthm}
\usepackage{array}
\usepackage{multirow}
\usepackage{mathtools}
\usepackage{mathrsfs}
\usepackage{color,graphicx}
\usepackage{thmtools, thm-restate}
\usepackage{algorithm,algpseudocode}
\usepackage{etoolbox,mathabx}

\graphicspath{{../figures}}

\makeatletter

\newcommand{\cH}{\mathcal{H}}

\newcommand{\cM}{\mathcal{M}}

\newcommand{\cR}{\mathcal{R}}

\newcommand{\cX}{\mathcal{X}}

\newcommand{\bitm}{\begin{itemize}}
\newcommand{\eitm}{\end{itemize}}
\newcommand{\bitme}{\begin{enumerate}[label=(\roman*),leftmargin=0.25in]}
\newcommand{\eitme}{\end{enumerate}}
\newcommand{\beq}{\begin{equation}}
\newcommand{\eeq}{\end{equation}}
\newcommand{\btcb}{\begin{tcolorbox}}
\newcommand{\etcb}{\end{tcolorbox}}
\def\bals#1\eals{\begin{align*} #1 \end{align*}}
\def\bal#1\eal{\begin{align} #1 \end{align}}

\newcommand{\with}{\quad \text{ with }  }

\newcommand{\Ffor}{\quad \text{ for }}

\newcommand\Dom\Omega

\newcommand\EE{\mathbb{E}}

\newcommand\RR{\mathbb{R}}

\newcommand\NN{\mathbb{N}}

\newcommand\Lap\Delta

\newcommand\abs[1]{\left\lvert #1 \right\rvert}

\newcommand\dee{\mathrm{d}}
\newcommand\da{\mathrm{d}a}
\newcommand\db{\mathrm{d}b}

\newcommand\dx{\mathrm{d}x}

\def\bpde#1\epde{\[\left\{\begin{aligned}#1\end{aligned}\right. \]}
\def\inbpde#1\inepde{\left\{\begin{aligned}#1\end{aligned}\right.}
\def\binpde#1\einpde{\left\{\begin{aligned}#1\end{aligned}\right.}

\newcommand\SemiNorm[2]{\lvert { #1 } \rvert_{#2}}
\newcommand\SemiNormlr[2]{\left\lvert { #1 } \right\rvert_{#2}}
\newcommand\Norm[2]{\lVert { #1 } \rVert_{#2}}

\def\bw{\mathbf{w}}

\def\cB{\mathcal{B}}

\def\cL{\mathcal{L}}

\def\cR{\mathcal{R}}
\def\half{\frac{1}{2}}

\def\bu{\mathbf{u}}
\def\bv{\mathbf{v}}
\def\bw{\mathbf{w}}

\def\bfB{\mathbf{B}}

\def\bfD{\mathbf{D}}

\def\bfI{\mathbf{I}}

\def\bfP{\mathbf{P}}

\def\bfS{\mathbf{S}}

\def\bfP{\mathbf{P}}
\def\bfU{\mathbf{U}}
\def\bfV{\mathbf{V}}
\def\bfW{\mathbf{W}}

\def\bfI{\mathbf{I}}

\def\bfU{\mathbf{U}}
\def\bfV{\mathbf{V}}

\def\bfTheta{\mathbf{\Theta}}
\def\bfGamma{\mathbf{\Gamma}}
\def\bfphi{\mbox{\boldmath$\phi$}}
\def\bfPhi{\mbox{\boldmath$\Phi$}}
\def\bfPsi{\mbox{\boldmath$\Psi$}}

\def\bfzeta{\mbox{\boldmath$\zeta$}}

\def\bone{\mathbf{1}}

\def\bx{\mathbf{x}}
\def\by{\mathbf{y}}
\def\bz{\mathbf{z}}
\def\ba{\mathbf{a}}
\def\bb{\mathbf{b}}
\def\bc{\mathbf{c}}

\def\bq{\mathbf{q}}
\def\br{\mathbf{r}}
\def\bs{\mathbf{s}}

\def\bv{\mathbf{v}}

\def\bfXi{\boldsymbol{\Xi}}

\def\bbmat{\begin{bmatrix}[r]}
\def\ebmat{\end{bmatrix}}

\newcommand{\barr}{\begin{array}}
\newcommand{\ea}{\end{array}}
\newcommand{\bea}{\begin{eqnarray}}
\newcommand{\eea}{\end{eqnarray}}
\newcommand{\bt}{\begin{table}}
\newcommand{\et}{\end{table}}

\DeclareMathOperator\Id{Id}

\DeclareMathOperator\diag{diag}

\newtheorem{genericthm}{GENERIC THEOREM ENVIRONMENT}[section]
\newtheorem{thm}[genericthm]{Theorem}

\newtheorem{defi}[genericthm]{Definition}

\numberwithin{equation}{section}

\newcommand\tturl[1]{{\tt \scriptsize [\url{{#1}}]}}

\makeatletter

\patchcmd{\@addmarginpar}{\ifodd\c@page}{\ifodd\c@page\@tempcnta\m@ne}{}{}
\providecommand*{\cupdot}{%
  \mathbin{%
    \mathpalette\@cupdot{}%
  }%
}
\newcommand*{\@cupdot}[2]{%
  \ooalign{%
    $\m@th#1\cup$\cr
    \hidewidth$\m@th#1\cdot$\hidewidth
  }%
}
\providecommand*{\bigcupdot}{%
  \mathop{%
    \vphantom{\bigcup}%
    \mathpalette\@bigcupdot{}%
  }%
}
\newcommand*{\@bigcupdot}[2]{%
  \ooalign{%
    $\m@th#1\bigcup$\cr
    \sbox0{$#1\bigcup$}%
    \dimen@=\ht0 %
    \advance\dimen@ by -\dp0 %
    \sbox0{\scalebox{2}{$\m@th#1\cdot$}}%
    \advance\dimen@ by -\ht0 %
    \dimen@=.5\dimen@
    \hidewidth\raise\dimen@\box0\hidewidth
  }%
}
\makeatother
\reversemarginpar

\newcommand{\remove}[1]{\textcolor{orange}{[Text removed.]}}%
\newcommand\hOmega{\widehat{\Omega}}
\newcommand\hOmegat{\widehat{\Omega}_t}

\newcommand{\otimesd}{\otimes_d}

\makeatother

\ifpdf
  \DeclareGraphicsExtensions{.pdf, .jpg, .tif}
\else
  \DeclareGraphicsExtensions{.eps, .jpg}
\fi

\begin{document}

\title[Learning LRNRs of Hyperbolic Wave Dynamics from Data]{Learning Low Rank Neural Representations of Hyperbolic Wave Dynamics from Data} 

\author[W. Cho]{Woojin Cho}
\address{Telepix, Seoul, South Korea, 07330}
\email{woojin@telepix.net}

\author[K. Lee]{Kookjin Lee}
\address{School of Augmented Intelligence, Arizona State University, Tempe, AZ, USA, 85281}
\email{kookjin.lee@asu.edu}

\author[N. Park]{Noseong Park}
\address{Department of Computer Science, Korea Advanced Institute of Science and Technology, Daejeon, South Korea, 34141}
\email{noseong@kaist.ac.kr}

\author[D. Rim]{Donsub Rim}
\address{Department of Mathematics, Washington University in St. Louis, St. Louis, MO, USA, 63135}
\email{rim@wustl.edu}

\author[G. Welper]{Gerrit Welper}
\address{Department of Mathematics, University of Central Florida, Orlando, FL, USA, 32816}
\email{gerrit.welper@ucf.edu}

\maketitle	

\begin{abstract}
  \noindent
    We present a data-driven dimensionality reduction method that is well-suited
    for physics-based data representing hyperbolic wave propagation. The method
    utilizes a specialized neural network architecture called low rank neural
    representation (LRNR) inside a hypernetwork framework. The architecture is
    motivated by theoretical results that rigorously prove the existence of
    efficient representations for this wave class. We illustrate through
    archetypal examples that such an efficient low-dimensional representation
    of propagating waves can be learned directly from data through a combination
    of deep learning techniques. We observe that a low rank tensor
    representation arises naturally in the trained LRNRs, and that this reveals
    a new decomposition of wave propagation where each decomposed mode
    corresponds to interpretable physical features. Furthermore, we demonstrate
    that the LRNR architecture enables efficient inference via a compression
    scheme, which is a potentially important feature when deploying LRNRs in demanding performance regimes.
\end{abstract}


\section{Introduction}

Hyperbolic waves arise in models of inviscid fluids such as gases or
depth-integrated surface flows like tsunamis
\cite{Liepmann-Roshko,Whitham1974,Ser99,LeVeque2002,Dafermos2010,leveque11}.
These waves are characterized by long-range propagation, shock formation, and
nonlinear interactions, all occurring without dissipation or dispersion.  The
defining attributes of these waves are the placement of sharp features along
smooth curves and the continuous evolution of these curves with respect to time
and geometry \cite{VanDyke1982}.  More generally, similar patterns are
encountered in systems that do not explicitly model hyperbolic waves. The smooth
placement of edges and curves is widely observed in data across a range of
scientific and engineering domains
\cite{Natterer,Mallat2009,DMDbook2016,Kutyniok2012}.  The routine manifestation
of wave-like structure in data underscores the fundamental importance of
computational methods that can learn efficient representations of such
structures in a data-driven manner.

The main subject of this paper is an efficient representation of hyperbolic wave
dynamics based on neural networks with a specialized architecture, one that is
supported by theory and can be learned directly from data alone through deep
learning techniques.  The representation is based on a new notion of low
dimensionality and is built on an implicit neural representation family called
the low-rank neural representation (LRNR) introduced in recent works
\cite{cho2023hypernetworkbased,rim2025lrnr,cho2025}.  We aim to establish that
LRNRs exhibit several desirable properties in the data-driven setting.  First,
LRNRs are efficient representations in the sense that they provide a
low-dimensional encoding of the data by mapping it to a low-dimensional latent
or coefficient space in a stable manner.  Second, they induce a new
decomposition of wave phenomena into interpretable features.  Third, the
dynamics governing the latent variables are highly regular, that is, they vary
smoothly in time; this key property, first observed in the theoretical
construction \cite{rim2025lrnr}, is also observed empirically in LRNRs trained
from data as we shall demonstrate.  Fourth, the LRNR architecture enables
accelerated evaluation and backpropagation computations; this is challenging to
achieve with generic architectures.  These properties make LRNRs promising
candidates for reduced or surrogate models for solving partial differential
equations (PDEs), as proposed in \cite{cho2023hypernetworkbased,cho2025}. 

Traditionally, sparsity has been one prominent measure of efficiency for data
representations, and there is a substantial body of literature devoted to
finding sparse representations of wave propagation using dictionaries or frames
such as wavelets, curvelets, and shearlets
\cite{wickerhauser1996adapted,DeVore1998,Mallat2009,candes2005,Kutyniok2012,Dahmen2018}.
These approaches are highly relevant to the study of digital images,
which also display
sharp features aligned along smooth curves, such as the edges of rigid objects
in photographic images; these features are sometimes modeled by so-called
cartoon-like functions, indicator functions over spatial regions with smooth
boundaries \cite{Petersen2018,Bolcskei2019,devore_hanin_petrova_2021,
MarcatiOpschoorPetersenSchwab2022,BernerGrohsKutyniokPetersen2022}.  Another
class of important constructions arises from a slightly different motivation, in
works from numerical analysis that seek to devise efficient representations
compatible with numerical methods for solving PDEs, such as finite volume
methods or discontinuous Galerkin methods \cite{LeVeque2002,dgbook}. In adaptive
mesh refinement (AMR) methods, the grid representation of a propagating wave is
refined to have higher resolution locally near shock discontinuities
\cite{Berger1984,Berger1998}.  These two perspectives, the
approximation-theoretic and the numerical-analytic, have also been studied in
conjunction, for example in the development of various discretizations of
Fourier integral operators \cite{candes2007,candes2009}.  

\vskip8pt
\textbf{Low rank neural representation and compositional low-rankness.}
The LRNR is based on a different theoretical notion of efficiency, one that
incorporates the compositional structure seen in deep learning models. To
illustrate the basic idea, we consider a simple example involving the wave
equation $(\partial_{tt}  - c^2 \partial_{xx}) u = 0$ in one spatial dimension.
The solution is given by the d'Alembert
formula \cite{Folland-pde1995,evans10}
\begin{equation} \label{eq:dalembert}
  u(x, t) = f(x - ct) + g(x + ct),
\end{equation} 
for some one-dimensional functions $f$ and $g$.  Intuitively, one observes that
four spatial functions are used in this construction: the pair $f(\cdot)$ and
$g(\cdot)$ describing the wave profiles, along with the pair $\Id(\cdot)$ and
$1(\cdot)$ (the identity map and the constant function, respectively) which
comprise the expression for the translation \begin{equation}
\label{eq:dalembert_inner}
  x \pm ct = \Id(x) \pm ct \cdot 1(x),
\end{equation}
where the scalars $\pm ct$ play the role of time-dependent coefficient
variables. Since the set of four functions $\{ f, g, \Id, 1 \}$ determines the
solution to the wave equation, one may view the solution as a four-dimensional
object. This perspective highlights a fundamental property of waves that is
already familiar to the reader: their solutions can be written as a composition
of a small number of interpretable components, some related to the shape of the
wave profile and others related to the propagation of the wave.

The LRNR architecture was devised to exploit this notion of dimensionality,  
relying on a generalization referred to as a compositional low-rank structure. 
This structure is
motivated by structures in nonlinear waves satisfying scalar conservation laws
that are much more complicated than the d'Alembert solution \cite{rim2025lrnr}.
The LRNR architecture is central to describing this structure, so we introduce
it here informally and defer the technical definition of LRNRs until 
Sec.~\ref{sec:lrnr_arch}. 

We begin with a feedforward neural network that maps spatial inputs to physical
outputs, that is, an implicit neural representation (INR)
\cite{Park2019,Sitzmann2020}.  A LRNR is defined as a family of feedforward
neural networks with depth $L$ where each weight matrix $\bfW^\ell$ and bias
vector $\bb^\ell$
is expressed as a linear combination of rank-1 weights $\bfW_i^\ell$
and biases $\bb_i^\ell$. 
Suppose these rank-1 weights and biases are provided to us and are
fixed; then $\bfW^\ell$ and $\bb^\ell$ take linear combinations of the fixed
matrices and vectors and only their coefficients are allowed to vary.  If we
denote these coefficients by a vector $\bs = (s^\ell_{p, i})_{p, i, \ell}$ whose
entries are selected through indices $p, i, \ell$ then we can write the weights
and biases as, for example,
\begin{equation} \label{eq:lr_weights_biases}
  \bfW^\ell(\bs) = \sum_{i=1}^{r^\ell_1} s_{1,i}^\ell \bfW_i^\ell, 
  \qquad
  \bb^\ell(\bs) = \sum_{i=1}^{r^\ell_2} s_{2,i}^\ell \bb_i^\ell.
\end{equation}
We denote the vector of layer-wise
ranks $\br = (r^\ell_p)_{p, \ell} \in \RR^{2L}$, where each entry
represents
the rank of a specific weight $r^\ell_1$ or bias $r^\ell_2$ at a certain layer
$\ell$.  The total number of coefficients in this case is given by the $1$-norm
of $\br$, that is, $n = \Norm{\br}{1} = \sum_{p, \ell}  r^\ell_p$.    
The width $M$ of a LRNR is given by the largest matrix row dimension of
the weight matrices $\bfW^\ell$ over all layers.

We denote by $\cH_{\br, M}$ the set of all LRNRs of rank $\br$ and width $M$.
Given a LRNR
$\cH_{\br,M}$, its member $h \in \cH_{\br,M}$ can be viewed as an INR $h =
h(\cdot; \bs)$ parametrized by its coefficients $\bs \in \RR^n$.  This is a
slightly more general version of LRNR introduced in previous works
\cite{cho2023hypernetworkbased,rim2025lrnr,cho2025}. 

\begin{figure}
\centering
\includegraphics[width=1.0\textwidth]{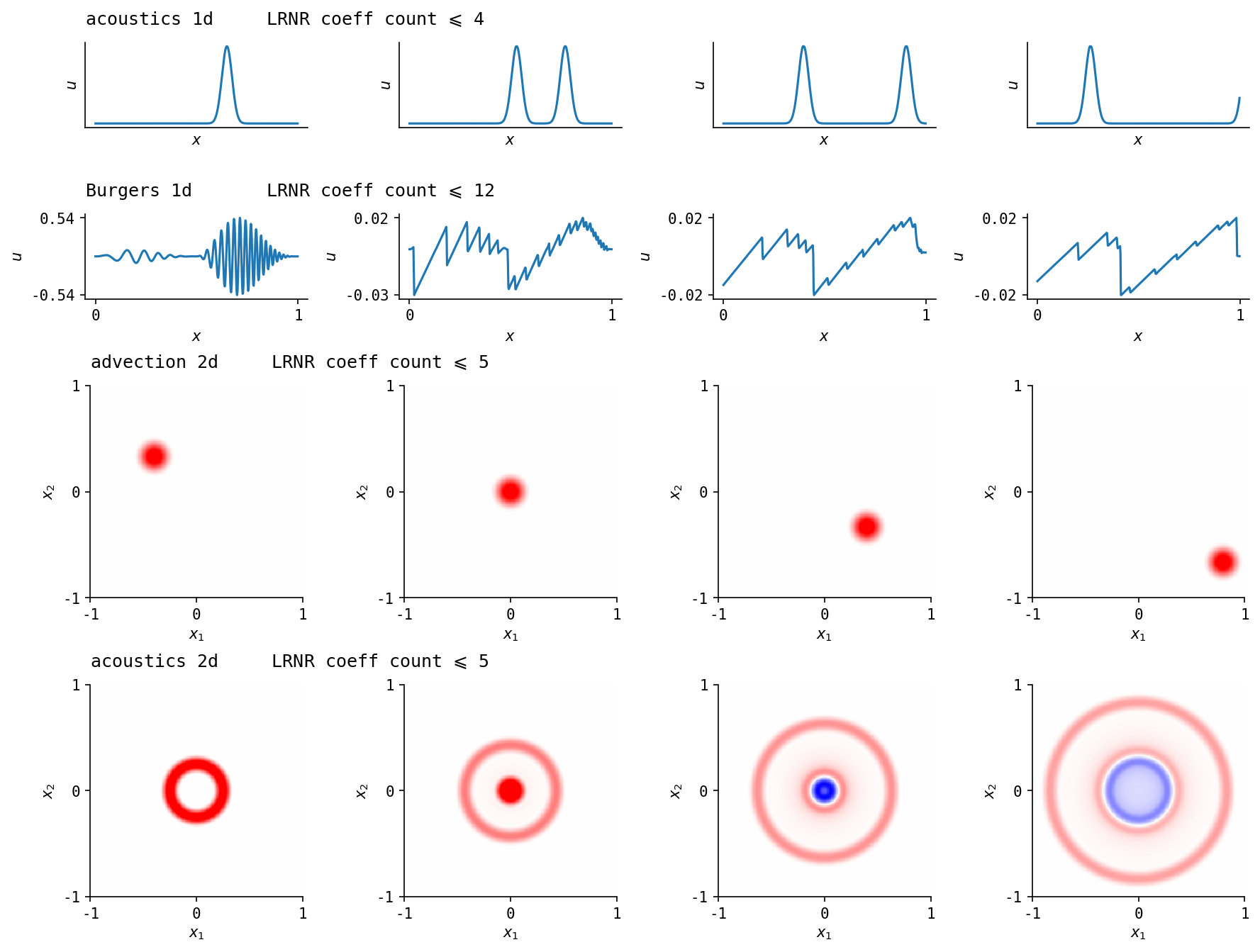}
\caption{ Various solution manifolds of time-dependent problems $\cM = \{
u(\cdot, t) \mid t \in [0, T]\}$ and their LRNR coefficient counts $\mathscr{C}$
(technical definition below in Def.~\ref{def:lrnr_cc}). The solution of 1d and
2d wave equations, as well as the 1d and 2d advection equations,
have a coefficient count of at most five (Thms.~\ref{thm:wave} and \ref{thm:advection}). 
Solution manifolds of scalar conservation laws
(e.g., the 1d Burgers' equation) have a coefficient count of at most twelve
(Thm.~\ref{thm:claw}).}
\label{fig:lrnr_manifolds}
\end{figure}

\vskip8pt
\textbf{The LRNR coefficient count.}
We can make a connection between the total number of LRNR coefficients $n$
discussed above and the notion of dimensionality discussed earlier in connection
with the d'Alembert solution. To do so, we introduce an approximation theoretic 
notion in terms of the number of coefficients. Informally, given a collection of
spatial
functions $\cM$, its \emph{LRNR coefficient count} $\mathscr{C}$ is defined as
the minimal number of LRNR coefficients $n$ required to construct a convergent
approximation of the functions in $\cM$ with uniform approximation rate $\alpha$
with respect to the width $M$ (precise definition in Sec.~\ref{sec:lrnr_cc}).
In this work, we consider the collection $\cM$
made up of solutions to initial value problems in a given time window
$\cM = \{u(\cdot, t) \mid t \in [0, T]\}$.

For solutions to representative hyperbolic problems, upper
bounds can be established for their coefficient counts; see the gallery shown in
Fig.~\ref{fig:lrnr_manifolds}.  For example, the d'Alembert solution discussed
above has a LRNR coefficient count of at most four, which matches the count of
four functions noted earlier in (\ref{eq:dalembert}-\ref{eq:dalembert_inner}). 
The entropy solution to the scalar conservation law $\partial_t u + \partial_x
F(u) = 0$ where the flux function is smooth and strictly convex ($F'' > 0$) is
significantly more complex than the wave equation, and is characterized by
nonlinear shock formation and interaction. Recent work \cite{rim2025lrnr} yields
an upper bound on the coefficient count of an entropy solution: it is at most
twelve.

For the wave equation in higher spatial dimensions, the analogue of the
d'Alembert solution is the Lax-Phillips translational representation
\cite{Lax64scattering} based on the planar wave decomposition of the
solution expressed in terms of the Radon transform.  In this work, we show that
the LRNR coefficient count of the solution to the $d$-dimensional wave equation
is at most $d+3$ via a plane wave decomposition in the form of neural networks.
This low dimensionality of the wave equation was previously
discussed in \cite{Rim18split,Rim18mr,Long2025}, and the construction is a type
of barycenter in the sliced Wasserstein metric \cite{Bonneel15slice,cuturi2014}.
On the other hand, lower bounds or optimal constructions have not yet been
established, and they appear particularly challenging to prove, especially for
deeper networks with more than two layers ($L >2$); new insights appear
necessary for lower bounds.

\vskip8pt
\textbf{Training LRNR from data.}
This work presents the first results on training LRNRs directly from data,
particularly for data that consist of solutions to hyperbolic problems.  The
theoretical results confirm the existence of efficient LRNRs of
hyperbolic waves, but they do not provide computational methods for training
LRNRs directly from data.  We aim to show that the training can be performed
by standard gradient-based learning
algorithms together with a combination of techniques including
hypernetworks, learning rate scheduling, regularization, and adaptively weighted
loss functions.  We note that hypernetworks
\cite{ha2016hypernetworks} have been widely used in conjunction with families of
INRs to represent continuous fields representing a collection of physical objects. 
They serve as an alternative to directly learning latent variables as inputs
\cite{Park2019},
and they were used successfully in conjunction with well-established
INRs like SIREN \cite{Sitzmann2020}. Our approach uses hypernetworks to
represent the
dependence of neural network parameters on physical parameters like time when
such data are available.  The Neural Implicit Flow (NIF) \cite{Pan2023} is based
on a similar physical-to-network parameter hypernetwork approach, although the
precise form of dependence is similar to that of SIRENs, but distinct from
our approach.  Some operator learning approaches, such as DeepONet
\cite{lu2021learning} or PCA-Net \cite{Hesthaven2018,Bhattacharya2021}, can be
viewed as hypernetwork frameworks as well.

We report an important property of LRNRs trained on data, namely the regularity
in the dynamics of the coefficients. This was an important property seen in the
theoretical constructions. In the d'Alembert solution discussed above, the
time-dependent coefficients $\bs(t)$ are linear $t \mapsto \pm ct$ with respect to $t$
\eqref{eq:dalembert_inner} and hence have highly regular dependence; in the
theoretical LRNR construction for entropy solutions \cite{rim2025lrnr}, it was
emphasized that the coefficients $\bs$ vary linearly with respect to time; in the
theoretical construction for the wave equation in all dimensions below, the
time-dependent coefficients are also linear in time. While these theoretical
constructions do not necessarily imply that the LRNRs trained on data will have
such properties, this work provides empirical evidence that LRNRs trained on
data via gradient descent also display temporal regularity in the learned
coefficient variables. Notably, this regularity is observed even when the wave
data contain jump discontinuities.

The LRNR has another distinguishing property that can potentially be important
for performance: The low-rank structure built into its architecture enables
it to be compressed. The compressed versions are referred to as FastLRNRs and
serve as accurate surrogates \cite{cho2025}. Here we demonstrate that
LRNRs trained on wave data can be evaluated efficiently using this compression
scheme, with the computational complexity scaling in terms of the rank
$\Norm{\br}{\infty}$ rather than the width $M$ which can potentially be large.
This property is useful in practice,
as it allows LRNRs with large widths to approximate complex
wave profiles, while simultaneously allowing LRNRs to be evaluated efficiently.
To the best of our knowledge, this property has not been discussed for other
proposed INRs (see discussions of performance in \cite{muller2022,Kerbl2023}). A
previous work by the authors \cite{cho2025} has applied this compression scheme
to approximate backpropagation in physics-informed neural networks, leading to
accelerated gradient descent iterations.  From a deep learning perspective, this
compression can be interpreted as a form of student-teacher knowledge transfer,
or knowledge distillation \cite{caruana2007,hinton2015}.  

In this work we report the following training results.
We train LRNRs on solution data from hyperbolic problems such as the
compressible Euler equations, variable speed advection equation, Burgers'
equation, and acoustic wave equations.  We show that an adaptation of the
meta-learning approach discussed
in \cite{cho2023hypernetworkbased,cho2025} leads to an additional form of
dimensionality reduction, appearing as low-rank tensors in the architecture;
this falls beyond the existing modes of dimensionality reduction. We observe
that this new reduction induces interpretable behavior when LRNRs are
extrapolated along what we call \emph{hypermodes}: it enables accurate and
efficient tangent space representations and displays remarkably coherent
extrapolation. We also demonstrate that FastLRNR compression \cite{cho2025} can
be used to evaluate these models efficiently at fixed spatial points. Finally,
we discuss the spatial extrapolation behavior of trained LRNR models, which
exhibits wave propagation consistent with space-time causality in hyperbolic
systems.

\vskip8pt
\textbf{Other related works.} 
There are relevant approximation theoretical results relating transport
equations, reduced models, and neural networks.  Results regarding parametric
linear transport equations were proved for smooth convective fields in
\cite{Laakmann2021}, where the initial condition and the smooth characteristic
curves are approximated \cite{Yarotsky2017,Petersen2018}. Neural networks were
shown to match the efficiency of reduced basis methods
\cite{Geist2021,Kutyniok2022}. The notion of compositional sparsity was
introduced in \cite{dahmen2023compositional}. These results mainly concern
generic feedforward architectures. We mention another architecture inspired by
solutions to Hamilton-Jacobi equations \cite{Darbon2021}.

More broadly, the LRNR-based framework is closely related to model reduction in
scientific computing, which aims to construct efficient representations of
solutions to partial differential equations in order to accelerate computation.
Classical approaches in this domain rely on linear approximations, such as a
reduced basis approximation \cite{Hesthaven2016,benner15}, but their
applicability is limited in many practical settings.  This is especially true
for hyperbolic wave problems, where sharp features propagate and interact in
complex ways \cite{rowley00,Ohlberger16,Welper2017,amsallem2016,rim17reversal}.
The challenge is referred to as the \emph{Kolmogorov barrier}, reflecting the
fundamental limitations of linear approximation as quantified by the Kolmogorov
$n$-width \cite{pinkus12}.  Many articles proposed approaches that aim to
overcome the Kolmogorov barrier: Template-fitting \cite{rowley00}, method of
freezing \cite{Beyn04,Ohlberger13}, shock reconstruction \cite{Constantine2012},
approximated Lax-Pairs \cite{Gerbeau14}, advection modes \cite{iollo14},
transported snapshot interpolation (TSI) \cite{Welper2017}, shifted proper
orthogonal decomposition (sPOD) \cite{schulze18},  calibrated manifolds
\cite{CagniartMadayStamm2019}, Lagrangian basis method \cite{Mojgani17},
transport reversal \cite{rim17reversal}, registration methods \cite{taddei2020},
Wasserstein barycenters \cite{ehrlacher19}, Front Transport Reduction (FTR)
\cite{Krah2023}, quadratic manifolds \cite{GeelenWrightWillcox2023}, or
Transformed Generative Pre-Trained Physics-Informed Neural Networks (TGPT-PINN)
\cite{Chen2024}.  Some of these proposed methods can be related to
dynamical low-rank approximations \cite{koch07,Sapsis2009}. 

The LRNR architecture itself can be viewed as a deeper generalization of
transported subspaces, a class of reduced models developed to overcome
theoretical approximation bounds while enabling efficient computation through
online-offline decomposition \cite{rim2023mats}. Their connection arises through
what is called the inverse-bias trick \cite{rim2025lrnr}.  In LRNRs, these ideas
are further extended to higher dimensions using neural network architectures,
which in turn generalize planar decompositions based on the Radon transform
\cite{Rim18split,Rim18mr}. Due to these connections to model reduction, the
results in this work represent a substantial step forward in overcoming the
Kolmogorov barrier. Nevertheless, we note that certain high-dimensional
hyperbolic problems were found to remain subject to other fundamental
performance constraints \cite{rim2023performance}.

In the general deep learning literature, there has been much interest in
incorporating low-rankness in the model architecture both for better performance
and better generalization: Introducing low rank factorizations of convolution
kernels \cite{jaderberg2014speeding}; Widely known low rank adaptation (LoRA)
techniques for large language models incorporate such techniques
\cite{hu2022lora}; Modification of LoRA to physics-informed settings
\cite{majumdar2023pihlora}; Low rank factorizations of components in the
transformer model \cite{deepseekai2025}. However, in these approaches specific
components of the architecture (e.g. convolution kernels or the adaptation) are
made low-rank, whereas the hidden states themselves are generally not low-rank.
This is
an important distinction from the LRNR architecture.  The LRNR training approach
is also distinct from train-then-truncate strategies considered in the past
\cite{cheng2017survey,Cheng2018}.

\section{Low Rank Neural Representation}

In this section, we introduce the LRNR architecture. For a more extensive
discussion of its motivation and derivation, we refer the reader to the
theoretical work \cite{rim2025lrnr}.  A more concise introduction, which is
perhaps more useful for practitioners, appears in
\cite{cho2023hypernetworkbased, cho2025}.

\subsection{The LRNR architecture} \label{sec:lrnr_arch}

We first define feedforward neural networks (NNs) in a standard manner
\cite{devore_hanin_petrova_2021,BernerGrohsKutyniokPetersen2022}.  Given the
dimensions $M_0, ...\,M_L \in \NN$, and an input $\bx \in \RR^{M_0}$, we
define the corresponding output $\by \in \RR^{M_L}$ of the NN $h(\bx) = \by :=
\bz^L$ by setting $\bz^0 := \bx$ and computing the sequence
$(\bz^\ell)$ via
\begin{equation} \label{eq:NN}
  \begin{aligned}
    \bz^\ell &= \sigma( \bfW^\ell \bz^{\ell - 1} + \bb^\ell ) \in \RR^{M_\ell}
    \quad \Ffor \ell = 1, ...\,, L-1, \\
    \bz^L &= \bfW^L \bz^{L - 1} + \bb^L \in \RR^{M_L},
  \end{aligned}
\end{equation}
where $\sigma$ is a nonlinear activation function, $\bfW^\ell \in \RR^{M_{\ell}
\times M_{\ell-1}}$ and $\bb^\ell \in \RR^{M_\ell}$.  Throughout, the
superscript $\ell$ is used as an index and not as an exponent.

We refer to a NN as an implicit neural
representation (INR) if its inputs are spatio-temporal coordinates and the goal
is to approximate a target spatio-temporal function. In our case, the inputs
$\bx \in \RR^d$ will be the spatial variables of the physical system, so the
input dimension $M_0$ is set as the spatial dimension $M_0 := d$.  In
particular, we are interested in representing a family of spatial functions, and
to this end we consider a family of neural networks with a corresponding family
of weights and biases. Suppose these weights and biases are determined by a set
of coefficient variables represented by a vector $\bs$; then one may write
\begin{equation}
  \bfW^\ell = \bfW^\ell (\bs),
  \qquad
  \bb^\ell = \bb^\ell (\bs),
  \qquad
  h = h(\cdot\,; \bs).
\end{equation}

The distinguishing feature of a LRNR is the enforcement of a low-rank structure 
on the weights and biases. Consider enforcing a constraint that the family of
weight matrices $\{\bfW^\ell(\bs) \mid \bs\}$ and bias vectors $\{\bb^\ell(\bs)
\mid \bs\}$, defined over all coefficient parameters $\bs$, share global column
and row spaces of certain rank. 
A simple formulation would express $\bfW^\ell(\bs)$ and $\bb^\ell(\bs)$ as
linear combinations \eqref{eq:lr_weights_biases}
of rank-1 matrices $\bfW_i^\ell$ and basis vectors $\bb_i^\ell$.
More abstractly, we are imposing on the weights and biases the constraints
\begin{equation} \label{eq:rowcolrank}
  \dim\, \bigcup_\bs \text{colspan} (\bfW^\ell(\bs)),\,
  \dim\, \bigcup_\bs \text{rowspan} (\bfW^\ell(\bs)) \le r^\ell_1,
  \qquad
  \dim \bigcup_\bs \text{colspan} (\bb^\ell(\bs)) \le r^\ell_2.
\end{equation}
This is the basic setup for the LRNR definition.  The full definition of the
LRNR architecture given below is modestly more general, in that 
we allow $\bfW_i^\ell$ to be sparse.

\begin{defi}[Low rank neural representation (LRNR)] \label{def:LRNR}
A LRNR of depth $L$ and rank $\br \in \RR^{2L}$ is a
family of feedforward neural networks \eqref{eq:NN} whose layer-wise weights
$\bfW^\ell$ and biases $\bb^\ell$ depend on a set of coefficient variables $\bs \in
\RR^n$ with $n = \Norm{\br}{1}$. Denoting the
individual entries of $\br$ and $\bs$ using the indexing
\begin{equation} \label{eq:coeff}
  \br =(r_p^\ell:  p = 1,2\,; \ell = 1, ...\,, L),
  \qquad
  \bs = (s^\ell_{p, i} : p = 1,2\,; i = 1, ...\,, r_p^\ell\,; \ell = 1, ...\,, L),
\end{equation} 
the $\bs$-dependent weights and biases are defined as 
\begin{equation} \label{eq:lowrank}
  \bfW^\ell(\bs) = \sum_{i=1}^{r^\ell_1} s_{1,i}^\ell \bfW_i^\ell
      \,\, \in \RR^{M_\ell \times M_{\ell-1}},
  \qquad
  \bb^\ell(\bs) = \sum_{i=1}^{r^\ell_2} s_{2,i}^\ell \bb_i^\ell
      \,\, \in \RR^{M_\ell},
  \qquad
  \ell = 1, 2, ... , L,
\end{equation}
in which $\bb_i^\ell$ are given vectors, and each of $\bfW_i^\ell$ 
are either given rank-1 or sparse matrices. Furthermore, we
\begin{itemize}
  \item denote by $M := \max_\ell M_\ell$ the width of the network,
  \item define $\cH_{\br,M}$ to be the collection of all LRNRs of depth $L$,
  rank $\br \in \RR^{2L}$, and width $M$, 
  \item set $\cH_\br := \bigcup_{M} \cH_{\br,M}$,
  \item for each LRNR $h \in \cH_{\br,M}$, the specific NN corresponding to the
  coefficient values $\bs$ is denoted by $h(\cdot\,; \bs)$, 
  \item denote by $\bs_1^\ell \in \RR^{r^\ell_1}$ the coefficients that appear in
  the weights in layer $\ell$, and $\bs_2^\ell \in \RR^{r^\ell_2}$ the
  coefficients that appear in the biases in the layer $\ell$.
\end{itemize}
\end{defi}

Given a LRNR $h \in \cH_{\br, M}$,
suppose the corresponding $(\bfW^\ell_i)$ and $(\bb^\ell_i)$ are learned first and are fixed.
Then the full parameter
state is encoded in the coefficient vector $\bs$,
which modulates the fixed weight and bias parameters.  
Thus we may identify the coefficient $\bs$ with a specific NN (or INR)
$h(\cdot; \bs)$.
The dimension of the coefficient vector $\bs$ is $n =
\Norm{\br}{1}$. As discussed in the introduction, $n$ is
referred to as the coefficient count, and represents a type
of expressiveness of LRNRs.
For hyperbolic problems the existence of efficient LRNR approximations
can be established rigorously, and this is the main topic of
Sec.~\ref{sec:lrnr_bounds}.

Def.~\ref{def:LRNR} agrees with the LRNR definition given in \cite{rim2025lrnr} 
with a couple of generalizations.
First, we allow the rank to be different for each weight and bias.
Second, Def.~\ref{def:LRNR} allows $\bfW_i^\ell$ to be a sparse matrix.
When $\bfW_i^\ell$ in \eqref{eq:lowrank} is rank-1, it can be
expressed as 
\begin{equation} \label{eq:Wrankone}
  \bfW_i^\ell = \bu_i^{\ell} \otimes \bv_i^{\ell},
  \qquad
  \bu_i^\ell \in \RR^{M_\ell},
  \quad
  \bv_i^\ell \in \RR^{M_{\ell-1}},
\end{equation}
where $\otimes$ denotes the outer product of two vectors.
In \cite{rim2025lrnr} the
operation $\otimes$ was generalized to the Hadamard-Kronecker
products $\otimesd : \RR^{m_1} \times \RR^{m_2} \to \RR^{m_1 m_2 \times m_2}$
and ${}_d\otimes : \RR^{m_1} \times \RR^{m_2} \to \RR^{m_1 \times m_1 m_2}$
between two vectors, defined as 
  $\displaystyle{\bu \otimesd \bv := \bu \otimes \diag(\bv)}$
  and
  $\displaystyle{\bu {}_d\otimes \bv := \diag(\bu) \otimes \bv}$
for vectors $\bu \in
\RR^{m_1}$ and $\bv \in \RR^{m_2}$. This enabled $\bfW^\ell_i$
to be set as certain class of sparse matrices, e.g, the identity $\bfI =
\diag(\bu) {}_d\otimes \bv$ when $\bu$ is taken to be a vector of ones and $\bv$
a $1$-vector with a single entry of one. Here, we allow $\bfW_i^\ell$ to be
a general sparse matrix.

\subsubsection{Implementation version}

In our implementations, we work with the simpler version used in
\cite{cho2023hypernetworkbased,cho2025}: $\bfW_i^\ell$ is taken to be a rank-1
matrix \eqref{eq:Wrankone}.
In this case, the weight matrix $\bfW^\ell(\bs)$ can be rewritten in factored form,
\begin{equation} \label{eq:weight_svd}
  \bfW^\ell(\bs)
  =
  \bfU^{\ell} \diag (\bs_1^\ell) \bfV^{\ell \top},
  \with
  \quad
  \bfU^{\ell}
  :=
  \left[\bu^{\ell}_1 \mid \cdots \mid \bu^{\ell}_{r^\ell_1} \right],
  \quad
  \bfV^{\ell}
  :=
  \left[ \bv^{\ell}_1 \mid \cdots \mid \bv^{\ell}_{r^\ell_1} \right].
\end{equation}
This simple version makes the weight
matrix $\bfW^\ell(\bs)$ resemble the factored form of singular value
decompositions \cite{cho2023hypernetworkbased}, but this factorized form 
of the weight matrix
is distinct from the classical decomposition since the coefficients (the
diagonal entries) are allowed to vary. The bias can also be written,
\begin{equation} \label{eq:bias_svd}
  \bb^\ell(\bs)
  =
  \bfB^{\ell} \bs_2^\ell
  \with
  \quad
  \bfB^{\ell}
  :=
  \left[\bb^{\ell}_1 \mid \cdots \mid \bb^{\ell}_{r^\ell_2} \right].
\end{equation}
It is straightforward to see that, in this formulation, the output of the 
affine mapping at each layer
\begin{equation}
  \bz^{\ell-1} \quad \mapsto \quad \bfW^\ell(\bs) \bz^{\ell-1} + \bb^\ell(\bs)
\end{equation}
is a vector belonging to the column space of $\bfW^\ell$ and $\bb^\ell$, and
therefore has rank at most $r^\ell_1 + r^\ell_2$. 

It is also straightforward 
that the output from this mapping is stable with respect to $\bz^{\ell-1}$
and $\bs$. Since the popular choice of nonlinear activation functions are 
Lipschitz continuous, this in turn implies the stability of the LRNR as a
whole with respect to the inputs and the coefficients, as it is a composition of
stable functions. In contrast, input instability is a well-known issue in
standard deep learning models \cite{Szegedy2014IntriguingPO,BCM+13}, and the
high-rank weights incur these instabilities \cite{rim2023stability}.

\section{Approximation results on LRNR coefficient counts}
\label{sec:lrnr_bounds}

In this section, we provide approximation theoretical results regarding the LRNR
coefficient counts, which imply the existence of efficient LRNR approximations
of several important hyperbolic wave solutions.

\subsection{The LRNR coefficient count} \label{sec:lrnr_cc}

We defined the LRNR architecture in the previous section, and for a given LRNR
of rank $\br \in \RR^{2L}$, the total number of coefficients in $\bs$
\eqref{eq:coeff} equals
\begin{equation}
  n
  =
  \Norm{\br}{1}
  =
  \sum_{p = 1,2} \sum_{\ell=1}^L r^\ell_p,
\end{equation} 
that is, $r^\ell_1$ coefficients for the weight and $r^\ell_2$
coefficients for the bias,
across layers $\ell = 1, ...\, , L$ \eqref{eq:lowrank}. 

Following the
discussion in the introduction, we
define the notion of a LRNR coefficient count for a family of functions $\cM$ as
the minimal number of coefficients needed for LRNRs in $\cH_{\br, M}$ to attain
certain approximation rates with respect to the width $M$. 
  
\begin{defi}[LRNR coefficient count] \label{def:lrnr_cc}
  Recall that $\cH_{\br}$ denotes the LRNRs of depth $L$, rank $\br \in
  \RR^{2L}$ and arbitrary width, and let $\lVert
  \cdot \rVert$ be a norm over a Banach space $\cX$.
  Given a family of functions
  $\cM \subset \cX$, let us define 
  \begin{equation} \label{eq:Anorm}
    \SemiNormlr{\cM}{\mathcal{A}^{\alpha} (\cH_{\br})}
    :=
    \sup_M M^{\alpha} 
    \inf_{h \in \cH_{\br, M}}
    \left(
      \sup_{u \in \cM}
      \inf_{\bs \in \RR^n } \Norm{u - h(\cdot\,; \bs)}{}
    \right),
    \qquad
    n = \Norm{\br}{1}.
  \end{equation}
  Then the \emph{LRNR coefficient count} of $\cM$ is defined as
  \begin{equation} 
    \mathscr{C}(\cM;\, \alpha )
    =
    \min
    \left\{
      n \in \NN
      \mid
      \SemiNormlr{\cM}{\mathcal{A}^{\alpha} (\cH_{\br})} < \infty,\,
      n = \Norm{\br}{1}
    \right\}.
    \label{eq:lrnr_cc}
  \end{equation}
\end{defi}

This definition relies on the commonly used notion of approximation
spaces (see, e.g., \cite{DeVore1998}) and tests
whether
LRNRs with certain rank $\br$, which forms a special family of
neural networks, 
can approximate a given collection of approximants $\cM$
with a standard approximation rate $\alpha$
with respect to the width $M$.

Now, we bolster the viewpoint from the introduction that the LRNR
architecture is inherently well-tailored for the task of approximating
solutions to hyperbolic problems, 
by presenting some upper bounds for the
coefficient counts of solutions to representative hyperbolic problems,
namely the $d$-dimensional wave equation (Sec.~\ref{sec:dwave}), 
the $d$-dimensional advection equation (Sec.~\ref{sec:dadvect}),
and the 1-dimensional scalar conservation law (Sec.~\ref{sec:dclaw}).
Throughout this section, we set our activation as $\sigma( \cdot ) = (\cdot)_+$
the Rectified Linear Unit (ReLU).

\subsection{The $d$-dimensional wave equation} \label{sec:dwave}

Let us consider the constant-speed wave equation in dimension $d$.
Let $\Omega \subset \RR^d$ denote a bounded domain with smooth $C^1$ boundary,
and $\hOmega$ denote the extended domain given by the Minkowski sum $\hOmega =
\Omega + B(cT)$ with the open $\ell_2$-ball $B(cT)$ of radius $cT$,
\begin{equation}\label{eq:wave_eqn}
  \left\{
    \begin{aligned}
      \partial_{tt} u - c^2 \Delta u &= 0 \text{ in } \hOmega \times (0, T) \\
                          u(\cdot, 0) &= u_0 \text{ in } \hOmega,\\
              \partial_t u(\cdot, 0) &= v_0 \text{ in } \hOmega,\\
    \end{aligned}
  \right.
  \quad \text{ where } \quad
  \left\{
  \begin{aligned}
  \text{ const. } &c > 0 ,
  \\
  u_0 &\in H^1(\hOmega),
  \\
  v_0 &\in L^2(\hOmega).
  \end{aligned}
  \right.
\end{equation}
Here $L^p(\Omega)$ denotes the Banach space induced by the $L^p$-norm
$\Norm{\cdot}{L^p(\Omega)} = (\int_\Omega |\cdot|^p \dx)^{1/p}$ for $1 \le p <
\infty$, $H^1(\Omega)$
denotes the Sobolev space whose weak derivative belong to $L^2(\Omega)$
\cite{evans10}.  
There is a unique solution $u(\cdot, t) \in H^1(\Omega)$ for the smaller
spatial domain $\Omega \subset \hOmega$ for $t \in [0, T]$, and the boundary
effects can be ignored due to the finite propagation speed in wave equations
\cite{evans10}, hence we omit the boundary conditions.

\subsubsection{Motivation} 
We first motivate the key ideas through an informal discussion.
Our result for the wave equation will be based on the fact that neural
networks can be written as planar wave decompositions
in the 2-layer case. For the given spatial input
$\bx \in \RR^d$, a scalar-valued 2-layer neural network is of the form
\begin{equation} \label{eq:plane}
  \bw^2 \cdot \sigma( \bfW^1 x + \bb^1) + b^2
  =
  \sum_{i=1}^M (w^2_i \sigma( \bw^{1}_i \cdot \bx + b_i^1) + b^2 / M)
  =
  \sum_{i=1}^M g_i ( \bw_i^1 \cdot \bx ),
\end{equation}
where we define the scalar 1d functions $g_i: \RR \to \RR$ as  $g_i(\beta) :=
w_i^2 \sigma( \,\beta\, + \hat{b}_i^1) + b^2/M$ for $i = 1, ...\,, M$. 

We will describe how, the solution $u$ of
\eqref{eq:wave_eqn} can be represented as a LRNR
using this decomposition. For the sake of
exposition, suppose we are given an initial condition that has precisely the
form in \eqref{eq:plane} with zero velocity, that is,
\begin{equation}
  u_0(\bx) = \sum_{i=1}^M g_i ( \bw_i^1 \cdot \bx ),
  \quad 
  \bw_i^1 \in S^{d-1},
  \qquad
  v_0 \equiv 0. 
\end{equation}
Considering the individual terms in the sum,
the application of the multidimensional Laplacian reduces to a single dimensional one,
\begin{equation} \label{eq:intertwine}
  \Delta g_i(\bw_i^1 \cdot \bx)
  =
  \abs{\bw_i^1}^2
  g''_i(\bw_i^1 \cdot \bx)
  =
  g''_i(\bw_i^1 \cdot \bx).
\end{equation}
Hence, the solution to the wave operator $(\partial_{tt} - c^2 \Delta)$ in
\eqref{eq:wave_eqn}, when applied to each planar function $g_i(\beta)$, can be
written as the 1d wave operator $(\partial_{tt} - c^2 \partial_{\beta\beta})$.
As a result, the d'Alembert solution $\half (g_i(\beta - ct) + g_i(\beta + ct))$
solves the 1d wave equation with the initial condition $g_i(\beta)$. By linearity,
we arrive at the solution
\begin{equation} \label{eq:gi_sol}
  u(\bx, t) = \frac{1}{2}\sum_{i=1}^M 
        \left[
          g_i ( \bw_i^1 \cdot \bx - ct)
          +
          g_i ( \bw_i^1 \cdot \bx + ct)
        \right].
\end{equation}
This representation is closely related to the Lax-Phillips
representation in odd dimensions \cite{Lax64scattering} and is the basis for
Radon transform-based techniques \cite{Bonneel15slice,Rim18split,Rim18mr,Long2025}.
Inserting the 2-layer form of $g_i$ we have assumed in \eqref{eq:plane}, the
explicit form of the solution \eqref{eq:gi_sol} can be rewritten as a LRNR,
\begin{equation}
  \begin{aligned}
  u(\bx, t)
  &=
  \frac{1}{2}
  \sum_{i=1}^M 
  \left[
    w^2_i \sigma( (\bw^{1}_i \cdot \bx) + b_i^1 -ct) + b^2/M
    +
    w^2_i \sigma( (\bw^{1}_i \cdot \bx) + b_i^1 +ct) + b^2/M
  \right]
  \\
  &= \widehat{\bw}^2 \cdot \sigma( \widehat{\bfW}^1 x + \widehat{\bb}^1(\bs(t))) + \hat{b}^2,
  \end{aligned}
\end{equation}
where only the first bias term $\widehat{\bb}^1$ is dependent on $t$. Focusing on this
bias term, we write it as a function of its coefficients $\widehat{\bb}^1(\bs(t))$: Once we let
\begin{equation}
  \bfB^1 :=
    \begin{bmatrix}
      \vdots & \vdots \\
      b_i^1  
      &
    -c
      \\
      b_i^1  
      &
      c  
      \\
    \vdots & \vdots 
    \end{bmatrix},
\end{equation}
the bias $\bb^1(\bs(t))$ is seen to be a linear combination of the two column vectors 
of $\bfB^1$ with the coefficients
\begin{equation}
    \widehat{\bb}^1(\bs(t))
    =
    \bfB^1 \bs^1_{2}(t),
    \quad
    \quad
    s^1_{2, 1}(t) \equiv 1,
    \quad
    s^1_{2, 2}(t) = t.
\end{equation}
Thus, the neural network can be expressed a LRNR with $r^1_2 = 2$. 

\subsubsection{LRNR coefficient count for the $d$-dimensional wave equation}

We now turn to a formal statement and its proof. 
We define the Barron space $\cB$ \cite{E2022} as the collection of
functions with finite Barron norm $\Norm{\cdot}{\cB_k}$ for $k \in \NN_0$ 
is given by
\begin{equation}
\Norm{f}{\cB_k}
:=
\inf_{\rho \in P_f} \EE \left[ \abs{a} (\Norm{\bw}{1}  + \abs{b}) \right]
\text{ where }
P_f = \left\{ \rho : f(\bx) = \int a \sigma^k(\bw\cdot \bx + b)\rho(\da,\dee\bw,\db) \right\}.
\end{equation}
If the initial conditions in \eqref{eq:wave_eqn} are in Barron spaces,
one can bound the LRNR coefficient count, and the rate $\alpha$
is provided by existing approximation theoretical results.

\begin{thm} [Wave equation in $d$ dimensions] \label{thm:wave} Suppose
$u$ is the solution to the $d$-dimensional wave equation \eqref{eq:wave_eqn}
where $u_0 \in \cB_1$ and $v_0 \in \cB_0$. Then for the collection $\cM := \{ u(\cdot, t)
|_\Omega \mid t \in [0, T]\}$, 
\begin{enumerate}[label=(\alph*)]
  \item there is a LRNR approximation $u_M \in \cH_{\br, M}$ with depth
  $L=2$ and rank $\br = (d, 2; 1, 0)$ whose coefficients $\bs$ depend on $t$ such that
    \begin{equation}
    \begin{aligned}
    \Norm{u_M(\cdot; \bs(t)) - u(\cdot, t)}{H^1(\Omega)}
    &\lesssim
    M^{-\frac{1}{2} -\frac{1}{2d}}
    \left( \Norm{u_0}{\cB_1} + \Norm{v_0}{\cB_0} \right),
    \\
    \Norm{\partial_t u_M(\cdot; \bs(t)) - \partial_t u(\cdot, t)}{L^2(\Omega)}
    &\lesssim
    M^{-\frac{1}{2} -\frac{1}{2d}}
    \left( \Norm{u_0}{\cB_1} + \Norm{v_0}{\cB_0} \right),
    \end{aligned}
  \end{equation}

  \item the LRNR coefficient count $\mathscr{C}$ for $\cM$ satisfies the upper bound,
  \begin{equation}
    \mathscr{C}\left(\cM; \alpha=\half + \frac{1}{2d}\right)
    \le
    d + 3.
  \end{equation}
\end{enumerate}
\end{thm}

\begin{proof}
As the initial data $u_0$ and $v_0$ lie in the Barron spaces $\cB_1$ and
$\cB_0$, respectively, they can be written
\begin{equation} 
  u_0(x) = \int a \sigma(\bw \cdot \bx + b) \dee \mu(\da,\dee\bw,\db), 
  \quad
  v_0(x) = \int a \sigma^0(\bw \cdot \bx + b) \dee \nu(\da,\dee\bw,\db), 
\end{equation}
with respective optimal measures $\mu$ and $\nu$. Here $\sigma^0$ is the
derivative of the ReLU, the jump function. Thanks to this
planar wave decomposition the solution to the wave equation with this initial
condition is given by (see Sec.~\ref{sec:planewave} or \cite[Sec. 5.F]{Folland-pde1995})
\begin{equation} \label{eq:uv_sol}\begin{aligned}
  u(\bx,t) & = U(\bx,t) + V(\bx,t), \\
  U(\bx,t) & = \frac{1}{2} 
  \int a\left[ \sigma(\bw \cdot \bx + b - ct) + \sigma(\bw \cdot \bx + b + ct) \right] \,d \mu(\da,\dee\bw,\db), \\
  V(\bx,t) & = \frac{1}{2} 
  \int a\left[ \sigma(\bw \cdot \bx + b - ct) + \sigma(\bw \cdot \bx + b + ct) \right] \,d \nu(\da,\dee\bw,\db). \\
\end{aligned} \end{equation}
To obtain a neural network representation, we employ Maurey sampling
\cite{Pisier1980,Makovoz1996,Siegel2024}:
There are $a_i$, $w_i$  and $b_i$ for $i=1, ...\,, m_1$ and $i = m_1+1, ...\,,
m_1+m_2$, respectively so that the sums
\begin{equation}  \label{eq:approx_IC}
  u_{0,m_1}(\bx) = \sum_{i=1}^{m_1} a_i \sigma(\bw_i \cdot \bx + b_i),
  \quad
  v_{0,m_2}(\bx) = \sum_{i=m_1+1}^{m_1+m_2} a_i \sigma' (\bw_i \cdot \bx + b_i),
\end{equation}
approximate the initial conditions in $\hOmega$,
\begin{equation}
  \Norm{u_{0,m_1} - u_0}{H^1(\hOmega)} \lesssim m_1^{-\frac{1}{2} - \frac{1}{2d}} \Norm{u_0}{\cB_1},
  \quad
  \Norm{v_{0,m_2} - v_0}{L^2(\hOmega)} \lesssim m_2^{-\frac{1}{2} - \frac{1}{2d}} \Norm{v_0}{\cB_0},
\end{equation}
and the constants depend on $\sigma, d$ and the respective choices of error norms.

Letting $M := 2(m_1 + m_2)$, the solution with the approximate initial
conditions \eqref{eq:approx_IC} can be written in the form of \eqref{eq:uv_sol},
as a sum of Dirac measures, resulting in the discrete analogue
\begin{equation} \begin{aligned}
  u_M(\bx,t) & = U_{m_1}(x,t) + V_{m_2}(x,t), \\
  U_{m_1}(\bx,t) & = \frac{1}{2} 
    \sum_{i=1}^{m_1} a_i \left[ 
      \sigma(\bw_i \cdot \bx + b_i - ct) + \sigma(\bw_i \cdot \bx + b_i + ct) \right],\\
  V_{m_2}(\bx,t) & = \frac{1}{2} 
    \sum_{i={m_1+1}}^{m_1+m_2} a_i \left[ 
      \sigma(\bw_i \cdot \bx + b_i - ct) + \sigma(\bw_i \cdot \bx + b_i + ct) \right]. \\
\end{aligned} \end{equation}
Due to linearity, the difference $u_M - u$ between the
approximate and exact solutions satisfies the wave equation. 
Using the energy estimates, we now show that $u_M - u$ remains small over time.

The energy $E(t)$ over the cone $\hOmegat := \Omega + B(c(T-t))$ reads
\begin{equation}
  E(t)
  :=
  \half
  \SemiNorm{u_M(\cdot, t) - u(\cdot, t)}{H^1(\hOmegat)}^2
  +
  \half
  \Norm{\partial_t u_M(\cdot, t) - \partial_t u(\cdot, t)}{L^2(\hOmegat)}^2,
\end{equation}
and satisfies $E'(t) \le 0$ for $t \in [0, T]$ (a standard energy estimate;
see for example \cite[Sec. 2.4]{evans10}) hence
\begin{equation}
  E(t) \le  E(0) \le
    {m_1}^{-1 - \frac{1}{d}} \Norm{u_0}{\cB_1}^2
    +
    {m_2}^{-1 -\frac{1}{d}} \Norm{v_0}{\cB_0}^2,
    \quad
    t \in [0, T].
\end{equation}
With an application of a Poincar\'e-Friedrichs inequality in time, 
one bounds the $L^2$-norm 
\begin{equation}
  \begin{aligned}
  \Norm{u_M(\cdot, t) - u(\cdot, t)}{L^2(\hOmegat)}^2
  &\lesssim
  \Norm{u_M(\cdot, 0) - u(\cdot, 0)}{L^2(\hOmegat)}^2
  +
  \Norm{\partial_t u_M(\cdot, t) - \partial_t u(\cdot, t)}{L^2(\hOmegat)}^2
  \\
  &\lesssim
  \Norm{u_M(\cdot, 0) - u(\cdot, 0)}{L^2(\hOmegat)}^2
  +
  E(0).
  \end{aligned}
\end{equation}

Pulling these estimates together,
\begin{equation}
  \begin{aligned}
  &\Norm{u_M(\cdot, t) - u(\cdot, t)}{H^1(\Omega)}^2
  +
  \Norm{\partial_t u_M(\cdot, t) - \partial_t u(\cdot, t)}{L^2(\Omega)}^2
  \\
  &\le
  \Norm{u_M(\cdot, t) - u(\cdot, t)}{H^1(\hOmegat)}^2
  +
  \Norm{\partial_t u_M(\cdot, t) - \partial_t u(\cdot, t)}{L^2(\hOmegat)}^2
  \\
  &\lesssim
  \Norm{u_M(\cdot, t) - u(\cdot, t)}{L^2(\hOmegat)}^2
  +
  E(t)
  \\
  &\lesssim
  \Norm{u_M(\cdot, 0) - u(\cdot, 0)}{L^2(\hOmegat)}^2
  +
  E(0)
  \\
  &\lesssim
  {m_1}^{-1 - \frac{1}{d}} \Norm{u_0}{\cB_1}^2
  +
  {m_2}^{-1 -\frac{1}{d}}  \Norm{v_0}{\cB_0}^2
  \\
  &\lesssim
  M^{-1 -\frac{1}{d}}
    \left( \Norm{u_0}{\cB_1}^2 + \Norm{v_0}{\cB_0}^2 \right).
  \end{aligned}
\end{equation}
So we arrive at the uniform error estimate for $t \in [0, T]$,
\begin{equation}
  \begin{aligned}
  \Norm{u_M(\cdot, t) - u(\cdot, t)}{H^1(\Omega)}
  &\lesssim
  M^{-\frac{1}{2} -\frac{1}{2d}}
  \left( \Norm{u_0}{\cB_1} + \Norm{v_0}{\cB_0} \right),
  \\
  \Norm{\partial_t u_M(\cdot, t) - \partial_t u(\cdot, t)}{L^2(\Omega)}
  &\lesssim
  M^{-\frac{1}{2} -\frac{1}{2d}}
  \left( \Norm{u_0}{\cB_1} + \Norm{v_0}{\cB_0} \right),
  \end{aligned}
\end{equation}
and the suppressed constants depend on $\sigma, d, T$ and on the choice 
of error norms. This proves (a).

Now we reformulate the approximation $u_M$ in the form of a LRNR. 
Combining all weights into vectors and matrices $\ba \in \mathbb{R}^{M}$, 
$\bfW \in \mathbb{R}^{M \times d}$, and $\bb \in \mathbb{R}^{M}$, we obtain
\begin{equation}
  u_M(\bx,t)
  = \frac{1}{2c} 
   \ba  \,
  \sigma \left(
  \bfW 
  \bx +
  \bb 
  + ct
  \begin{bmatrix} +\bone \\ -\bone \end{bmatrix}
  \right).
\end{equation}
With a slight abuse of notation, we rewrite this as a LRNR with time-dependent
coefficients $s(t)$ of width $M$ and depth $L=2$
\begin{equation}
  u_M(\bx; \bs(t))
  = 
  \underbrace{ \frac{1}{2c} \ba }_{\bfW^2}
  \sigma \Big(
  \underbrace{ \bfW }_{\bfW^1}
  \bx +
  \underbrace{
    \left[\begin{array}{c r} \multirow{2}{*}{$\bb$ $\Big\rvert$ \hskip-10pt}  & c \bone\\ & -c \bone \\ \end{array}\right]}_{\bfB^1}
  \underbrace{\begin{bmatrix} 1 \\ t \end{bmatrix}}_{\mathclap{\bs^1_2(t)}}
  \Big).
\end{equation}
The weight matrices $\bfW^1$ and $\bfW^2$ are fixed, hence their rank
is equal to the usual rank of these matrices,
\begin{equation}
  r^1_1 = \text{rank}(\bfW^1) \le d,
  \qquad
  r^2_1 = \text{rank}(\bfW^2) = 1,
\end{equation}
and $\bb^2 \equiv 0$, so we have $r^2_2 = 0$. So the rank of this LRNR is
at most
\begin{equation}
  \br
  =
  (r^1_1, r^1_2; r^2_1, r^2_2)
  =
  (d, 2; 1, 0),
\end{equation}
hence the coefficient count is at most $n =\Norm{\br}{1} = d + 3$, for the
norm $H^1(\Omega)$ and the rate $\alpha = \half + \frac{1}{2d}$.
This proves (b).

\end{proof}

\subsection{The $d$-dimensional advection equation} \label{sec:dadvect}

Similar bounds can be obtained for the constant-speed advection equation.

\begin{thm}[Advection equation in $d$ dimensions] \label{thm:advection}
  Let $\hOmega$ and $\Omega$ 
be as in \eqref{eq:wave_eqn}, and let $u$ be the
solution to the advection equation posed in $\hOmega$
\begin{equation}
  \left\{
  \begin{aligned}
    \partial_t u + \ba \cdot \nabla u &= 0 \text{ in } \hOmega,
    \\
    u(\cdot, 0) &= u_0 \text{ in } \hOmega,
  \end{aligned}
  \right.
  \quad \text{ with } \quad
  \Norm{\ba}{2} = c,
  \quad \text{ and } \quad
  u_0 \in L^2(\hOmega).
\end{equation}
If $u_0 \in \cB_1$ then for the collection $\cM := \{u(\cdot, t) \mid t \in [0,
T]\}$ the LRNR coefficient count $\mathscr{C}$ satisfies the upper bound
\begin{equation}
  \mathscr{C}\left( \cM ; \alpha=\half + \frac{1}{2d} \right) \le d + 3.
\end{equation}
  where the underlying error norm is $\Norm{\cdot}{L^2(\Omega)}$.
\end{thm}

\begin{proof}
The proof is very similar to that of Theorem~\ref{thm:wave}, therefore we omit
it.
\end{proof}

\subsection{The 1$d$ scalar conservation laws} \label{sec:dclaw}

A notable recent result is that deeper LRNRs can efficiently handle arbitrary
shock interactions in nonlinear scalar conservation laws. This result can be
rephrased in terms of the coefficient count.

\begin{thm} [1$d$ scalar conservation laws] \label{thm:claw}
Let $u$ be the solution to the scalar conservation law posed on 
the unit interval $\Dom := (0, 1)$ for a given smooth flux $F \in
C^\infty(\RR)$ that is convex $F'' > 0$,
  \begin{equation}
    \left\{
    \begin{aligned}
      \partial_t u + \partial_x F(u) &= 0 
      \quad \text{ in } \Dom \times (0, T),
      \\
      u(\cdot, 0) &= u_0.
    \end{aligned}
    \right.
  \end{equation}
  with compactly supported initial condition $u_0 \in BV(\Omega)$ and final
  time $T$ small enough so $u(\cdot, t)$ is compactly supported in $\Omega$
  for $t \in [0, T]$.  For the collection $\cM := \{u(\cdot, t) \mid t \in
  [0, T]\}$ the LRNR coefficient count $\mathscr{C}$ satisfies the upper
  bound
  \begin{equation}
    \mathscr{C}\left( \cM ; \alpha=\half \right) \le 12,
  \end{equation}
  where the underlying error norm is $\Norm{\cdot}{L^1(\Dom)}$.
\end{thm}

\begin{proof}
  Recalling the LRNR construction used in the main theorem from
  \cite{rim2025lrnr}, there exists a LRNR $u_M \in \cH_{\br, M}$ with number of
  layers $L=5$ and $\br = (1, 2, 1, 1, 1 ; 1, 1, 2, 2, 0)$ such that
  \begin{equation}
    \Norm{ u(\cdot, t) - u_M(\cdot, t)}{L^1(\Omega)} \lesssim M^{-\half}
    \SemiNormlr{u_0}{TV} (1 + \SemiNormlr{u_0}{TV}) 
                  (1 + T \Norm{F''}{L^\infty(u_0(\Dom))}),
    \quad
    t \in [0, T]
  \end{equation}
  where $u_M$ has width $M$. So we have LRNR coefficient count $n =
  \Norm{\br}{1} = 12$ with rate $\alpha=\half$.
\end{proof}
  
Hence, these theoretical constructions prove that efficient representations of
complicated wave phenomena exist. This naturally leads to the computational question
of whether such efficient representations can be learned directly
from data: The rest of this work reports results in the positive.

\section{Data-driven meta-learning framework for wave problems}
\label{sec:lrnr_train}

In this section, the hypernetwork framework and the precise training
methods are discussed.  In previous works
\cite{cho2023hypernetworkbased,cho2025}, the meta-learning with a hypernetwork
approach was shown to be effective in the physics-informed neural networks
setting. Here, we investigate the effectiveness of a similar framework in
a data-driven setting. 

We will use a LRNR 
of depth $L$ and rank $\br \in \RR^{2L}$ 
given in Def.~\ref{def:LRNR} with the choice of rank-1 matrices in
\eqref{eq:Wrankone}.
The hypernetwork approach \cite{ha2016hypernetworks} employs an auxiliary
neural
network called the hypernetwork to generate the weights and biases of an INR.
In our case, only the coefficient variables will be output from the
hypernetwork.
We set our hypernetwork $f_{\text{hyper}}$ to output the coefficients
$\bs \in \RR^n$ (with $n = \Norm{\br}{1}$), 
and since we are interested in the dynamics,
we set the hypernetwork to take as input the time variable $t$.
That is, 
\begin{equation}
  f_{\text{hyper}}(t) = \bs,
  \qquad
  t \in [0, T],
  \qquad
  \bs \in \RR^n,
\end{equation}
and the resulting coefficients $\bs$ \eqref{eq:coeff} are used as coefficients
in the low-rank weight and bias expansions \eqref{eq:lowrank}.
  
\begin{figure}
  \centering
  \begin{tabular}{ll}
  \includegraphics[width=1.0\textwidth]{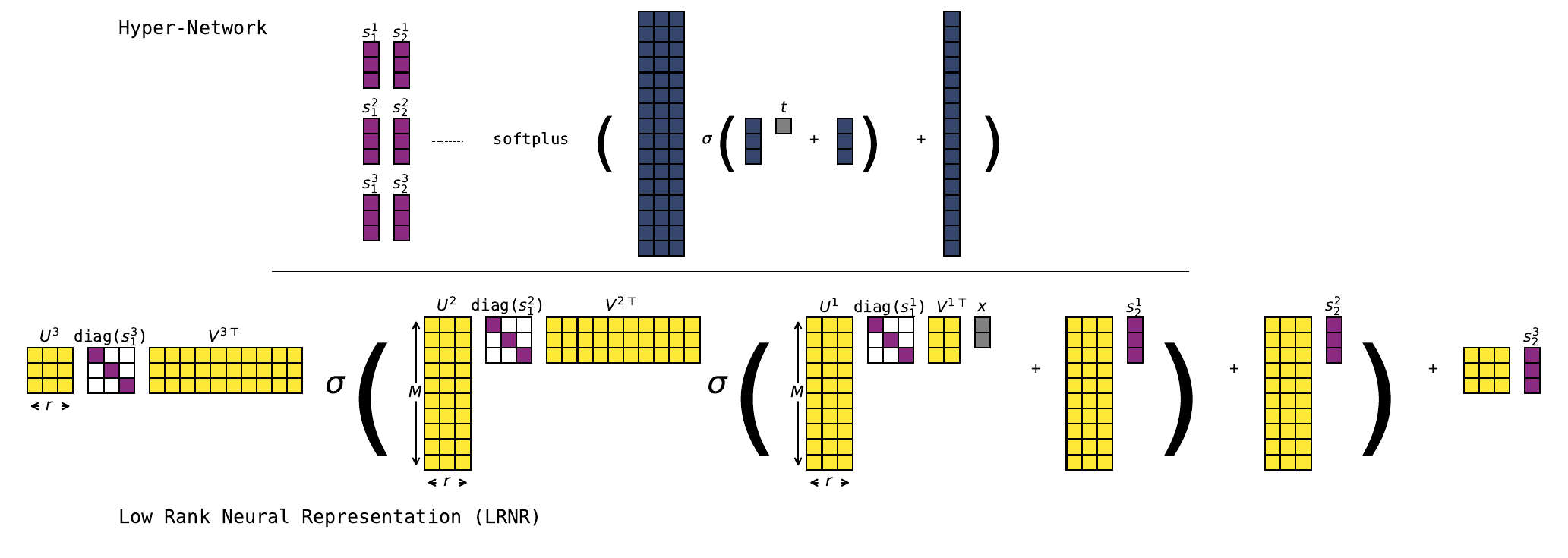}
  \end{tabular}
  \caption{A diagram depicting the hypernetwork and the low rank neural
  representation. The hypernetwork takes the time variable $t$ as input, and
  outputs the coefficients $\bs$ \eqref{eq:coeff} which are used as the 
  highlighted entries to generate the LRNR weights and biases. In this example,
  the input dimension is two, the output dimension is three, and the rank is
  three.}
  \label{fig:lrnr_diagram}
\end{figure}

The LRNR and the hypernetwork together form a meta-network, defined as
the concatenation of a hypernetwork and a LRNR: A diagram depicting the
meta-network is given in Fig.~\ref{fig:lrnr_diagram}.

Recall that the coefficient vector $\bs$ \eqref{eq:coeff} consists of entries
$\bs^\ell_{1,i}$ and $\bs^\ell_{2,i}$, where $\bs^\ell_{1,i}$ denotes the $i$-th
coefficient for the weight matrix in layer $\ell$, used as the diagonal
entries in the SVD-like factored form \eqref{eq:weight_svd}, and
$\bs^\ell_{2,i}$ denotes the $i$-th coefficient for the bias in layer $\ell$,
used to scale the basis vectors $\bb_i^\ell$ \eqref{eq:bias_svd}.
Aside from the coefficients $\bs$, the LRNR formulation has the
$\bs$-independent parameters $\bfU^{\ell},\bfV^{\ell}$ \eqref{eq:weight_svd}
and $\bfB^\ell$ \eqref{eq:bias_svd}.
Collecting all of these, the entirety of $\bs$-independent parameters is
\begin{equation}
  \bfU = (\bfU^{\ell})_{\ell=1}^L,
  \qquad
  \bfV = (\bfV^\ell)_{\ell=1}^L,
  \qquad
  \bfB = (\bfB^\ell)_{\ell=1}^L.
\end{equation}
It remains to choose the dimensions of these matrices:
We choose the same width across all layers $M_\ell \equiv M$ for $\ell=1,
...\,,L-1$ for simplicity.

With this notation, we may write the LRNR as a function of the 
spatio-temporal input, parameterized by $\bfU$, $\bfV$, $\bfB$ (which are independent
of $\bs$) and the coefficients $\bs$:
\begin{equation}
  h(\bullet\,; \bfU, \bfV, \bfB, \bs).
\end{equation}
Next, we denote by $\bfTheta$ the parameters of the hypernetwork
$f_{\text{hyper}}: \RR \to \RR^n$ 
\begin{equation}
  f_{\text{hyper}}(\bullet\,; \bfTheta).
\end{equation}
The meta-network is then the concatenation
\begin{equation}
  h(\bullet\,; \bfU, \bfV, \bfB, f_{\text{hyper}}(\bullet\,; \bfTheta)),
\end{equation}
which can be interpreted as a larger feedforward neural network with
parameters $\bfU, \bfV, \bfB$, and $\bfTheta$.
  
We emphasize that the key distinction between this formulation and other
proposals is that the hypernetwork outputs only the coefficients $\bs$, as
opposed to outputting all entries of the weights and biases as in SIREN
\cite{Sitzmann2020} or NIF \cite{Pan2023}. 
The hypernetwork outputs coefficients $\bs$ which are then multiplied by the
matrices in $\bfU, \bfV, \bfB$ that
remain constant, thus the variation in the weights and biases are 
restricted so as to preserve low-rankness in the sense of 
(\ref{eq:weight_svd}-\ref{eq:bias_svd}). In generic hypernetwork 
approaches like the SIREN hypernetwork
approach \cite{Sitzmann2020}, it is technically possible for a generic
hypernetwork (that outputs all parameters) to discover 
specific parameters satisfying
this particular
inductive bias during training; in practice this was found to be 
unlikely in our experiments. 
In the NIF approach the variation is not independent across layers due to the
bottleneck
structure in the output of the hypernetwork; thus it is not possible for
the NIF approach \cite{Pan2023} to discover the low-rank structure of a LRNR
in its general form.  We note, however, that one example given in the 
paper introducing NIF 
appears to implicitly exploit the LRNR structure,
due to the choice of small width that indirectly induces the weights to be
low-rank (see the first figure of Fig.~3 in \cite{Pan2023}).

For training, we use standard misfit cost functions, with the exception
that we normalize them as follows:
\begin{equation} \label{eq:misfit}
  \cL_{\text{misfit}}(\widehat{\by}, \by; q)
  :=
  \frac{\Norm{\widehat{\by} - \by}{q}^q}{\Norm{\by}{q}^q}.
\end{equation}
Throughout, we use $q = 1$ or $q = 2$, corresponding to normalized versions of
$\ell_1$ loss and Mean Squared Error (MSE), respectively.

\subsection{Sparsity promoting regularizer and orthogonality constraints}

A standard optimization using only the misfit function \eqref{eq:misfit} and a
PINNs penalty was shown to be effective in the first work that introduced the
hypernetwork approach for LRNRs \cite{cho2023hypernetworkbased}.  However, the
trained LRNRs did not necessarily exhibit a low-rank structure.  In the analogy
between \eqref{eq:weight_svd} and the singular value decomposition (SVD),
low-rankness would correspond to the property \eqref{eq:weight_svd} holding for
a small $r$.  However, the coefficients \eqref{eq:coeff} that were output from
the hypernetwork did not exhibit strong decay with respect to the index $i$,  
requiring $r$ to be roughly on the same scale as the width $M$.  A
straightforward approach to inducing low-rankness in this case will be to
introduce shrinkage, for example a simple $\ell_1$ regularizer, intended to
promote sparsity. However, in our experiments, this approach was not effective
in inducing sparsity in the coefficient variables. We attribute this phenomenon
to the fact that the coefficients, now playing the role of singular values in
\eqref{eq:weight_svd}, vary with time; so the dominant values also change.
  
This motivated the sparsity-promoting regularizer introduced in \cite{cho2025},
which penalizes non-decay while relaxing it to allow a sparse violation of the
penalty. It is given by
\begin{equation} \label{eq:reg_sparse}
  \cR_{\text{sparse}} (\bs; \gamma)
  :=
  \sum_{\ell=1}^L \sum_{p=1,2}
    \Norm{ (\bfGamma(\gamma) \bs^\ell_p )_+ }{\ell^1},
  \qquad
  \bfGamma(\gamma)
  = 
  \begin{bmatrix}
    -1 & \gamma & & & \\
      & -1 & \gamma & & \\
      &   & \ddots & \ddots & \\
      &   &  & -1 & \gamma \\
  \end{bmatrix}.
\end{equation}
This regularizer does not directly promote sparsity in $s$, but instead encourages 
sparsity in the transformed vector
\begin{equation}
  \left( (\gamma s^\ell_{p, j+1} - s^\ell_{p, j})_+ \right)_j
\end{equation}
for each $p$ and $\ell$. We incorporate this regularizer into our loss function.

We also apply soft orthogonality constraints to the vectors involved in
the rank-1 matrix expansions,
\begin{equation} \label{eq:reg_ortho}
  \cR_{\text{ortho}}(\bfU, \bfV, \bfB)
  :=
  \sum_{\ell=1}^L
  \left(
  \frac{1}{\# \bfU^{\ell }}
  \Norm{\bfU^{\ell \top} \bfU^{\ell} - \bfI}{F}^2
  +
  \frac{1}{\# \bfV^{\ell}}
  \Norm{\bfV^{\ell \top} \bfV^\ell - \bfI}{F}^2
  +
  \frac{1}{\# \bfB^{\ell}}
  \Norm{\bfB^{\ell \top} \bfB^\ell - \bfI}{F}^2
  \right)
\end{equation}
where $\Norm{\cdot}{F}$ denotes the Frobenius norm, and $\#$ indicates the
number of entries in a matrix. In our experiments, the inclusion of
the orthogonality penalty consistently improved the training loss decay.
  
Finally, our training loss function is obtained by adding the
regularizers (\ref{eq:reg_ortho}-\ref{eq:reg_sparse}) to the misfit
loss function,
\begin{equation} \label{eq:loss}
  \begin{aligned}
  \cL(\bfU, \bfV, \bfB, \bfTheta)
  &:=
  \EE_{x, t}
  \,
  \left[
    \cL_{\text{misfit}}(
        u(\bx;  \bfU, \bfV, \bfB, f_{\text{hyper}} (t; \bfTheta)),
        u_{\text{data}}(\bx, t)
      )
  \right]
  \\
  &\quad
  +
  \lambda_{\text{sparse}} \cR_{\text{sparse}}( f_{\text{hyper}}(t))
  +
  \lambda_{\text{ortho}} \cR_{\text{ortho}}(\bfU, \bfV, \bfB).
  \end{aligned}
\end{equation}
We use the Adam optimizer \cite{Kingma2015AdamAM} to minimize the loss.
The hyperparameters $\lambda_{\text{sparse}}$ and $\lambda_{\text{ortho}}$, 
which control the strength of the regularization terms, and the architecture 
hyperparameters $M$, $r$, and $L$ of the LRNR, are tuned using random search 
among a set of feasible choices,
as implemented in the {\tt Ray Tune} package \cite{liaw2018tune}.

\subsection{Mollifiers, adaptive sampling, and misfit switch}

We introduce several techniques to handle challenges in training the model
with the loss \eqref{eq:loss}.

\subsubsection{Mollifiers}
Due to the sharp jumps present in our compressible Euler example, standard
training methods for INRs are not effective. For instance, we implemented the
hypernetwork approach using SIREN, where a similar training procedure was
applied except that the INR was a standard (non-low-rank) neural network, and
the hypernetwork outputs all of the weights and biases of the NN in \eqref{eq:NN}
(these results are discussed in Sec.~\ref{sec:train_results}).

To alleviate this difficulty in examples with a single spatial dimension $d=1$, we
mollify the data by convolving the uniformly sampled spatial data with a box
function. We initialize the width of the support of the box function and
decrease it according to a schedule over the course of training epochs:
\begin{equation}
  \cL_{\text{misfit}}\left(
    u(\bx; \bfU, \bfV, \bfB, f_{\text{hyper}}(t; \bfTheta)),
    (\chi_{w} * u_{\text{data}})(\bx, t)
  \right),
\end{equation}
where $\chi_w$ is a $d$-dimensional box function defined as $\chi_w :=
\frac{1}{(2w)^d} \prod_{i=1}^d 1_{[-w, w]}(x_i)$.

The convolved function is a smeared-out version of the original, making it
easier to fit using gradient descent-based algorithms 
\cite{rahaman2019,basri2020,Xu2020}. We initialize the mollifier radius $w_0$ as
a user-specified hyperparameter, and at epoch $i_{\text{epc}}$, contract the
radius using a linear decay schedule:
\begin{equation} \label{eq:mollifier_width}
  w_{i_{\text{epc}}}
  =
  w_0 \left(
    \frac{n_{\text{epc}} - 2i_{\text{epc}}}{n_{\text{epc}}}
  \right)_+,
\end{equation}
so that the radius vanishes halfway through the allocated number of epochs. 
This strategy is a form of numerical continuation \cite{LeVeque2007} or
curriculum learning \cite{allgower2003,Bengio2009}.

\subsubsection{Adaptive sampling}

Another approach we deploy to resolve sharp discontinuities in the data is to
use adaptive spatial sampling. Adaptive mesh refinement (AMR) codes such as
Clawpack are designed to produce adaptive samples that concentrate near sharp
features. In higher spatial dimensions, using AMR-generated data yielded
superior results compared to applying mollifiers.

\subsubsection{Misfit switch}

We also choose the misfit loss adaptively. Taking the weighted misfit
\begin{equation} \label{eq:misfit_switch}
  \cL_{\text{misfit}} (\bullet)
  :=
  \alpha
  \cL_{\text{misfit}}(\bullet; q=1)
  +
  (1 - \alpha)
  \cL_{\text{misfit}}(\bullet; q=2)
\end{equation}
we switch from MSE to $\ell_1$ loss whenever a predefined threshold $\tau$ is
reached by starting with $\alpha=0$ then setting $\alpha=1$ in the first epoch
in which $\cL_{\text{misfit}} < \tau$, 
and reset the learning rate. This strategy is
motivated by the observation that MSE loss often results in persistent
Gibbs-like oscillations when approximating sharp jump discontinuities, whereas
$\ell_1$ loss tends to yield more accurate reconstructions. The approach is
further supported by the $L_1$ minimization formulation of convective problems
\cite{guermond2007,guermond2008}. It is a form of multi-objective weighting
\cite{Chen2018,xu2018autoloss,heydari2019softadapt,Wang2021,Wang2022}, although
the switch is applied instantaneously. The threshold $\tau$ is determined
heuristically through preliminary training runs without switching, and is set
slightly above the best MSE misfit loss attained in those runs.

\begin{figure}
  \centering
  \begin{tabular}{ccc}
    \includegraphics[width=0.32\textwidth]{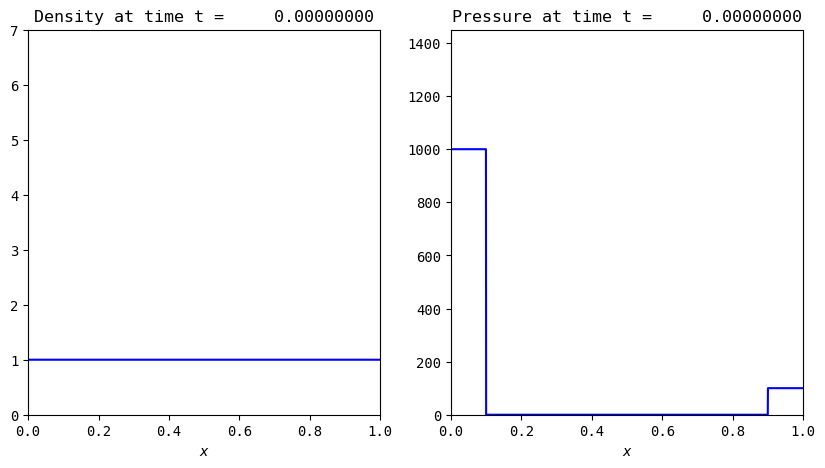}
    &
    \includegraphics[width=0.32\textwidth]{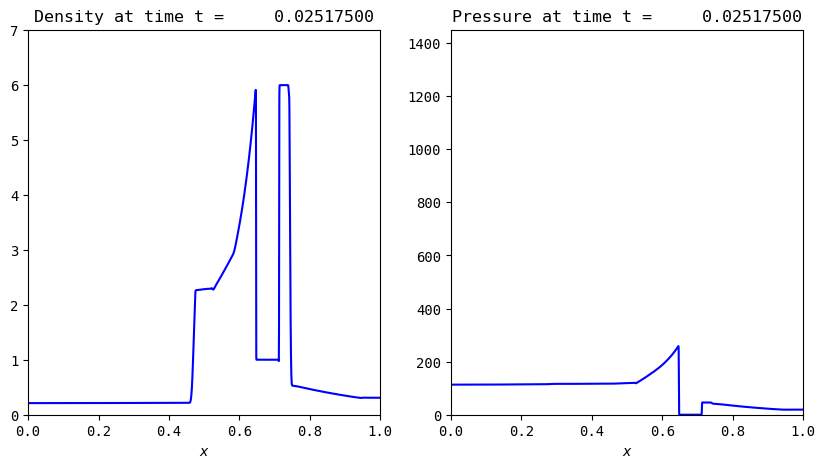}
    &
    \includegraphics[width=0.32\textwidth]{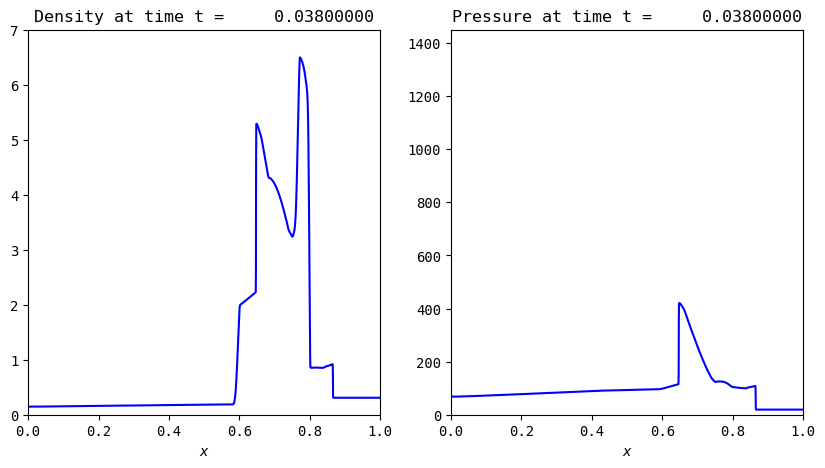}
    \\
    \includegraphics[width=0.32\textwidth]{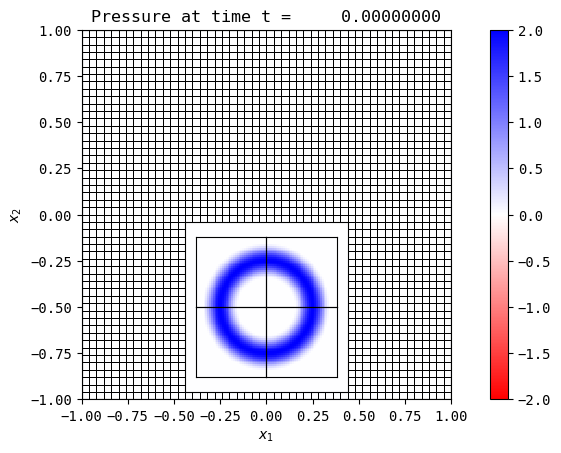}
    &
    \includegraphics[width=0.32\textwidth]{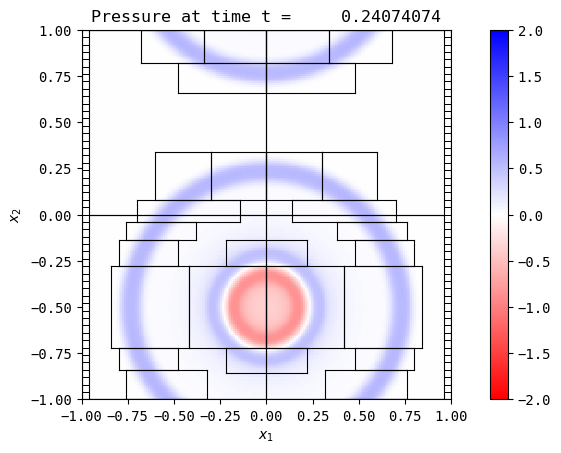}
    &
    \includegraphics[width=0.32\textwidth]{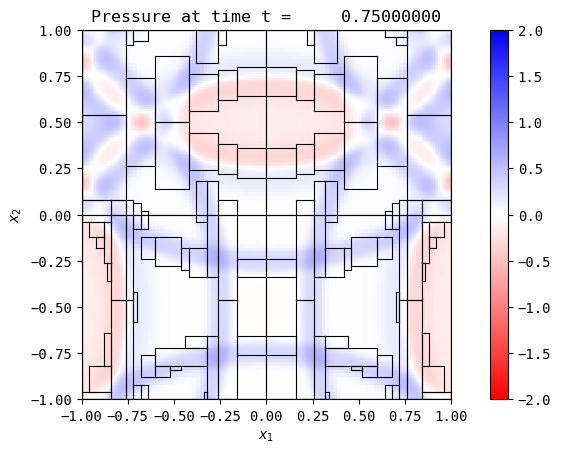}
    \\
    \includegraphics[width=0.32\textwidth]{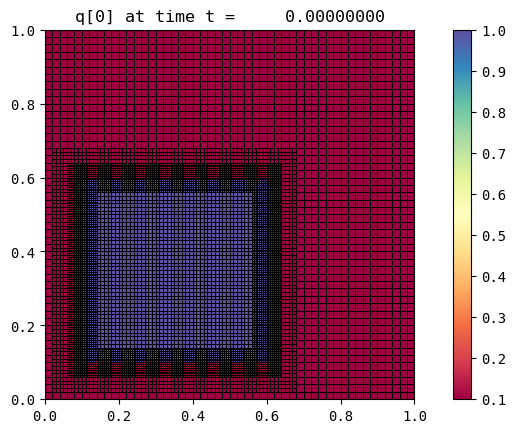}
    &
    \includegraphics[width=0.32\textwidth]{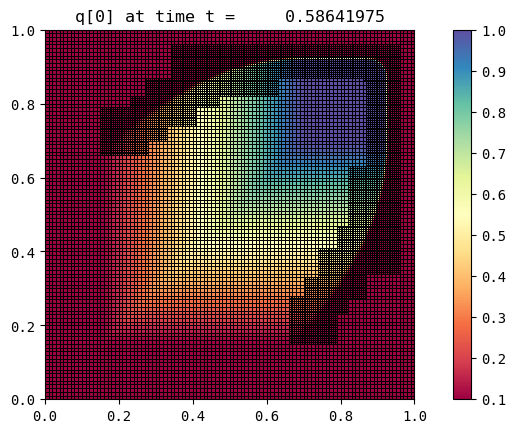}
    &
    \includegraphics[width=0.32\textwidth]{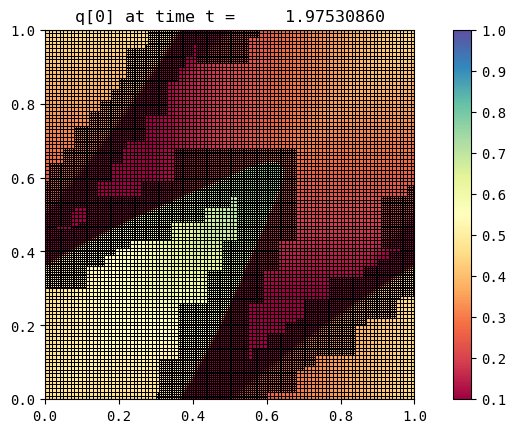}
    \\
    \includegraphics[width=0.32\textwidth]{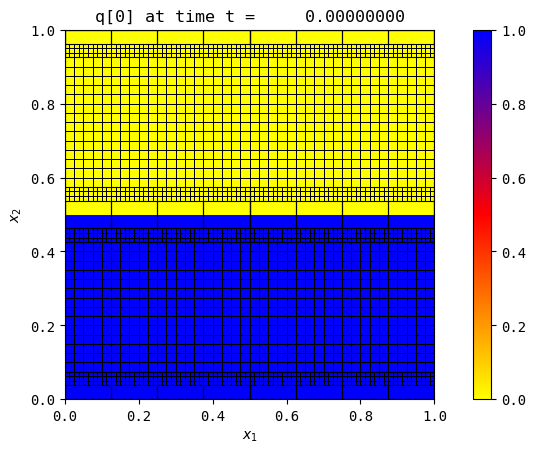}
    &
    \includegraphics[width=0.32\textwidth]{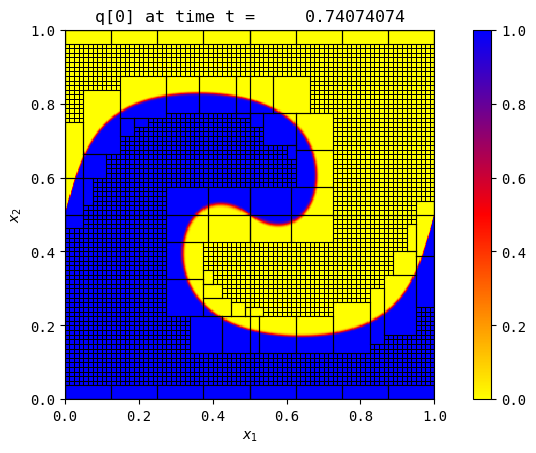}
    &
    \includegraphics[width=0.32\textwidth]{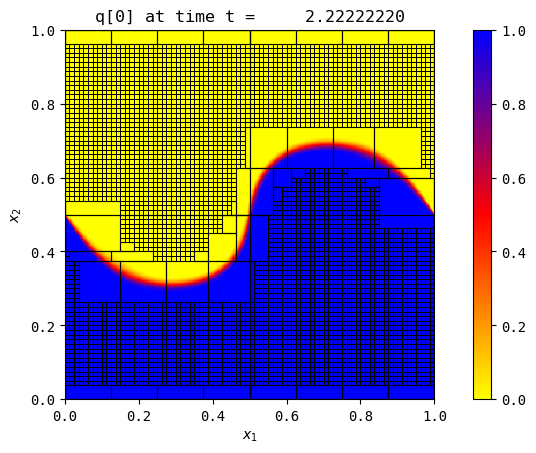}
  \end{tabular}
  
  \caption{Plots of training data from the 1d Euler example (first row), 2d
  acoustics (second row), 2d Burgers (third row), and 2d advection (fourth row).
  For the 2d examples, AMR (Adaptive Mesh Refinement) data were used, where the
  finest-grid values at each spatial location were extracted as training data.
  The grid edges indicate the boundaries of AMR patches \cite{Berger1998}. All
  data were generated using the Clawpack software package \cite{clawpack}.}

  \label{fig:examples}
\end{figure}

\subsection{Training results} \label{sec:train_results}

We report our training results below. Four representative examples are
considered, each exhibiting characteristic behaviors of hyperbolic systems: the
1d compressible Euler equations (Woodward-Colella shock collision problem
\cite{Woodward1982,Woodward1984,LeVeque2002}); the 2d acoustics equations with
periodic boundary conditions; the 2d Burgers equation with periodic boundary
conditions; and the 2d advection equation with time-dependent, spatially varying
advection velocities. For each example, we use PDE solution data consisting of
81 equally spaced time samples, generated using the Clawpack software package
\cite{clawpack}. See Fig.~\ref{fig:examples} for the solution plots displaying
the data.

\subsubsection{1d Compressible Euler example: Woodward-Colella problem}

We show the training results for the 1d Euler example.
Fig.~\ref{fig:euler1dloss} displays the training loss, regularization terms,
mollifier radius, and learning rate over time. The threshold for switching from
the (normalized) MSE loss to the (normalized) $\ell_1$ loss was set to $5 \times
10^{-4}$.  This threshold is reached quickly, and it was observed that
mollifying the data facilitates reaching the threshold more efficiently via
gradient descent.  Once the misfit loss switches \eqref{eq:misfit_switch}, the
mollifier radius continues to decrease linearly according to the prescribed
schedule \eqref{eq:mollifier_width}. Each reduction in radius causes a temporary
increase in loss due to the instantaneous change in the data, but subsequent
gradient descent iterations reduce the loss again. 

During epochs approximately from 3K to 40K, there is a visible competition
between the increasing tendency of the $\ell_1$ loss whenever the mollifier
radius is decreased, and the decreasing tendency due to gradient descent. After
this transition period, the mollification is fully phased out, and continued
training with the standard $\ell_1$ loss leads to further improvement in data
fitting.  Note that the MSE loss corresponds to the square of the $\ell_2$ loss.
Thus, a threshold value of $5 \times 10^{-4}$ for the MSE loss roughly
corresponds to an $\ell_2$ loss of $0.02$ after taking the square root.  As a
result, the apparent change in loss and accuracy at the switching point (marked
by the vertical red line) is visually exaggerated in this plot.

The orthogonality regularization tends to stabilize only after the loss function
switches from MSE to $\ell_1$. This suggests that the training process
prioritizes enforcing sparsity in the coefficient variables before gradually
promoting orthogonality in the basis vectors.

\subsubsection{Comparison with vanilla SIREN}

\begin{figure}
  \centering
  \includegraphics[width=0.85\textwidth]{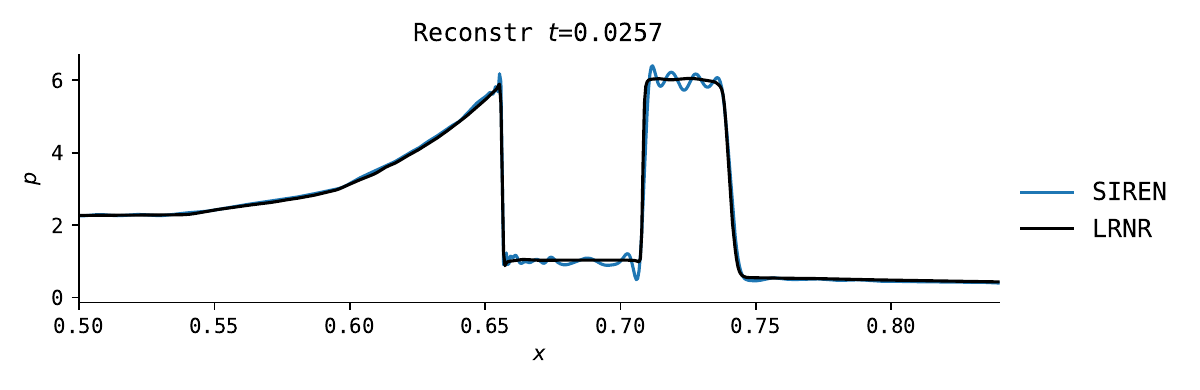}
  \caption{Comparison of reconstructions between SIREN and LRNR. Oscillations
  persist in the SIREN model trained with MSE loss, whereas the LRNR
  reconstruction exhibits damped oscillations and sharper jump discontinuities.}
  \label{fig:siren_lrnr}
\end{figure}

The trained LRNR result is compared against that of the hypernetwork-SIREN
approach trained with the MSE loss. In Fig.~\ref{fig:euler1dloss}, we plot the
$\ell_2$ misfit (the square root of the MSE loss) to enable visual comparison
with the $\ell_1$ misfit obtained from LRNR training in the later epochs. The
$\ell_2$ loss for SIREN decreases initially but diverges abruptly after
approximately 100K epochs. We display the trained SIREN output prior to this
divergence and compare it with the LRNR representation in
Fig.~\ref{fig:siren_lrnr}. The SIREN result remains oscillatory despite a
similar number of training iterations.

For the hypernetwork-SIREN, we use a SIREN INR with width 512 and depth 5. A
smaller width of 256 failed to achieve comparable loss levels. The accompanying
hypernetwork has width 256 and depth 5, with scalar modulation parameters also
tuned. Recall that the hypernetwork outputs the full set of weights and biases
for the SIREN network. Although the hypernetwork has width 256, implying that
its output lies in a 256-dimensional space, this low-rank structure is not
preserved once the outputs are reshaped into the appropriate weight and bias
matrices. In particular, singular value decomposition (SVD) of the resulting
$512 \times 512$ weight matrices shows no significant decay in singular values.

Neural implicit flow models exhibit a similar architecture to the
hypernetwork-SIREN \cite{Pan2023}, where the hypernetwork output is enforced to
be low-rank. However, these models reshape the hypernetwork outputs into INR
weights and biases in a similar manner, meaning that the resulting matrices do
not necessarily possess a low-rank structure. In contrast, the LRNR architecture
imposes an explicit low-rank matrix structure, resulting in a trained model
whose weight matrices are structurally distinct from both hypernetwork-SIREN and
neural implicit flow.

It is important to note, however, that the presented results are not intended as
a direct comparison with those of SIREN. The trained LRNR models exhibit
structural features and properties that are absent in SIRENs. Even if a SIREN
can be trained to achieve similar reconstruction accuracy, the low-dimensional
structures intrinsic to LRNR, which will be discussed in the following sections,
would still be lacking. Conversely, it is certainly possible to incorporate key
features of the SIREN architecture into LRNRs, for instance by employing sine
activation functions or introducing modulation parameters. Whether the
well-known advantages of SIREN architectures carry over to the LRNR setting
remains an open research question, which we do not explore in this work.

\subsubsection{Hyperparameters}

The hyperparameters used in our experiments are summarized in
Tab.~\ref{tab:hyperparams}. Hyperparameter selection was more consistent across
the 2d examples, all of which employed adaptive mesh refinement (AMR) data. For
the 1d case, which used a uniform spatial grid, mollification of the data was
necessary. The misfit switch threshold was set to $5 \times 10^{-4}$ for the 1d
Euler example and $5 \times 10^{-3}$ for the 2d examples. The initial learning
rate was set to $10^{-3}$, with a \texttt{ReduceLROnPlateau} scheduler applied
using a decay factor of 0.98, patience of 10 epochs, and a threshold of
$10^{-3}$.

The results were particularly sensitive to the values of
$\lambda_{\text{sparse}}$ and $\gamma$, whereas $\lambda_{\text{ortho}}$ had a
relatively minor impact. Values of $\lambda_{\text{ortho}}$ around $10^{-2}$
produced acceptable results across all examples. In addition to reducing memory
requirements, the use of AMR data in the 2d cases also eliminated the need for
mollification.

\begin{table}
  \centering
  \begin{tabular}{r|lllllllll|l}
                  & $M$ & $r$ & $L$ & $M_{\text{hyper}}$ & $L_{\text{hyper}}$ & batch & $\lambda_{\text{sparse}}$ & $\lambda_{\text{ortho}}$ & $\gamma$ & grid \\
                  \hline
                  \hline
    1d Euler     & 1200 & 150 & 5 & 25 & 4 & 16 & 3.65e-18 & 7.76e-3 & 1.0005
          & unif \\
    2d acoustics & 2000 & 190 & 4 & 10 & 3 & 1 & 2.35e-08  & 1.00e-2 & 1.0500
          & adapt \\
    2d Burgers   & 2000 & 170 & 4 & 10 & 3 & 1 & 2.71e-11  & 1.00e-2 & 1.0500  
          & adapt \\
    2d advection & 2000 & 180 & 4  & 10 & 3 & 1 & 3.48e-10 & 1.00e-2 & 1.0650 
          & adapt \\
          \hline
  \end{tabular}
  \caption{Tuned hyperparameters. $M$, $r$, and $L$ denote the width, rank, and
  depth of the LRNR, respectively. $M_{\text{hyper}}$ and $L_{\text{hyper}}$
  refer to the width and depth of the hypernetwork. “Batch” refers to the batch
  size. $\lambda_{\text{sparse}}$ is the regularization parameter for the
  sparsity-promoting regularizer, while $\lambda_{\text{ortho}}$ is the
  regularization parameter for the orthogonality regularization term. $\gamma$
  controls the coefficient value decay. The last column indicates the type of
  input data used for each example.}
  \label{tab:hyperparams}
\end{table}

\section{Hypermodes and low-rank tensor representation }

A remarkable property of neural network-based autoencoders is that even when
given complicated spatial data, the dynamics in the latent variables can be very
smooth \cite{lee19,Fries2022}.  In the d'Alembert solution for the 1d wave
equation \eqref{eq:dalembert}, we have already seen that the time-dependent
coefficients $\pm ct$ were linear in time $t$
\eqref{eq:dalembert_inner}, regardless of the regularity of the wave profiles
$f$ and $g$.
  
For LRNRs, this phenomenon is also observed in other theoretical constructions
(Thms.~\ref{thm:wave}-\ref{thm:advection}), even for entropy solutions
involving nonlinear shock interactions (Thm.~\ref{thm:claw};
\cite{rim2025lrnr}).  A particularly notable feature of these constructions is
that the coefficient variables exhibit linear (or constant) dependence on time,
despite the presence of discontinuous spatial behavior in the solution.

This observation leads us to the question of whether 
the trained LRNRs using the
hypernetwork approach exhibit similar time regularity. In
Fig.~\ref{fig:coeff_evolve}, we plot the evolution of the coefficient outputs
from the hypernetwork for the 1d Euler equations, and we observe that: (1) Many
components of the coefficients remain close to zero throughout, suggesting a
time-uniform low-rank structure; and (2) the coefficients evolve smoothly over time
and appear approximately linear. 

Motivated by these observations, we investigate two types of reductions: (1)
truncating coefficients by discarding uniformly small entries, and (2)
exploiting temporal regularity.  The first truncation can be applied
independently to each weight and bias component. We denote by $r^\ell_1$ the
truncated rank of the weight matrix $\bfW^\ell$, and by $r^\ell_2$ the truncated
rank of the bias vector $\bb^\ell$.  The resulting truncated ranks are reported
in Tab.~\ref{tab:hypermodes}. The reduction is substantial, indicating that the
sparsity-promoting regularizer was effective in inducing coefficient decay.
Beyond this straightforward reduction, however, the coefficient evolution plots
in Fig.~\ref{fig:coeff_evolve} reveal another possibility: a reduction in the
dynamics of the full coefficient vector as a whole. We explore this aspect in
the following section.

\begin{figure}
  \centering
  \includegraphics[width=1.0\textwidth]{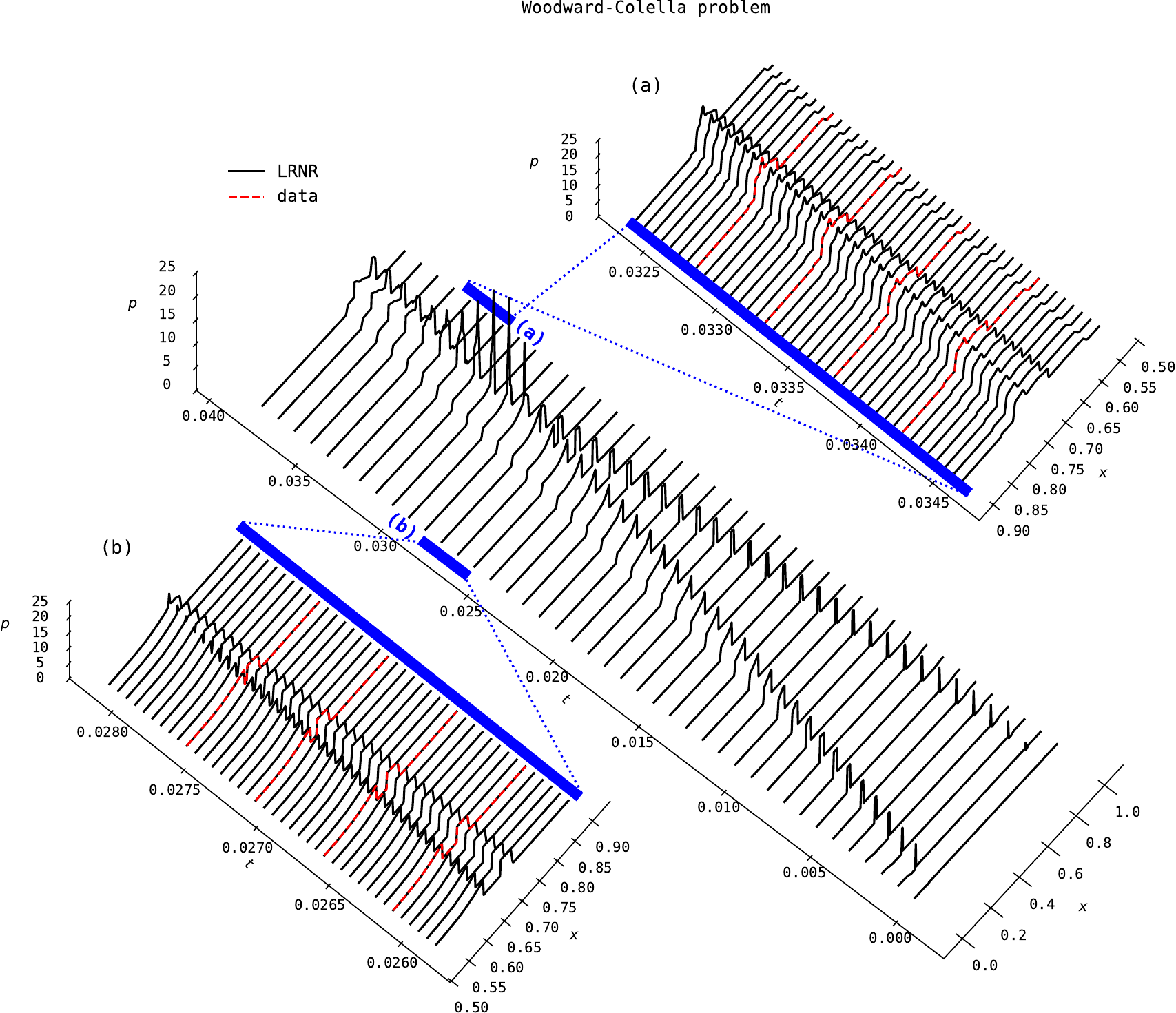}
  \caption{LRNR reconstruction of the solution to the 1d compressible Euler
  equations with periodic boundary conditions (Woodward-Colella shock problem
  \cite{Woodward1982,Woodward1984,LeVeque2002}).  (a,b) Zoom-in plots highlight the
  learned representation at interpolated times, $u(\cdot\,;
  f_{\text{hyper}}(t))$.  In (a), the rarefaction waves and shocks that emerge
  after the shock collision are individually transported with high accuracy.  In
  (b), the sharp peak formed at the shock merger is correctly placed and closely
  mimics the behavior of the finite volume solution, despite the fact that the
  precise timing of the collision must be inferred from data.}
  \label{fig:euler1d_setup}
\end{figure}

\begin{figure}
  \centering
  \includegraphics[width=1.0\textwidth]{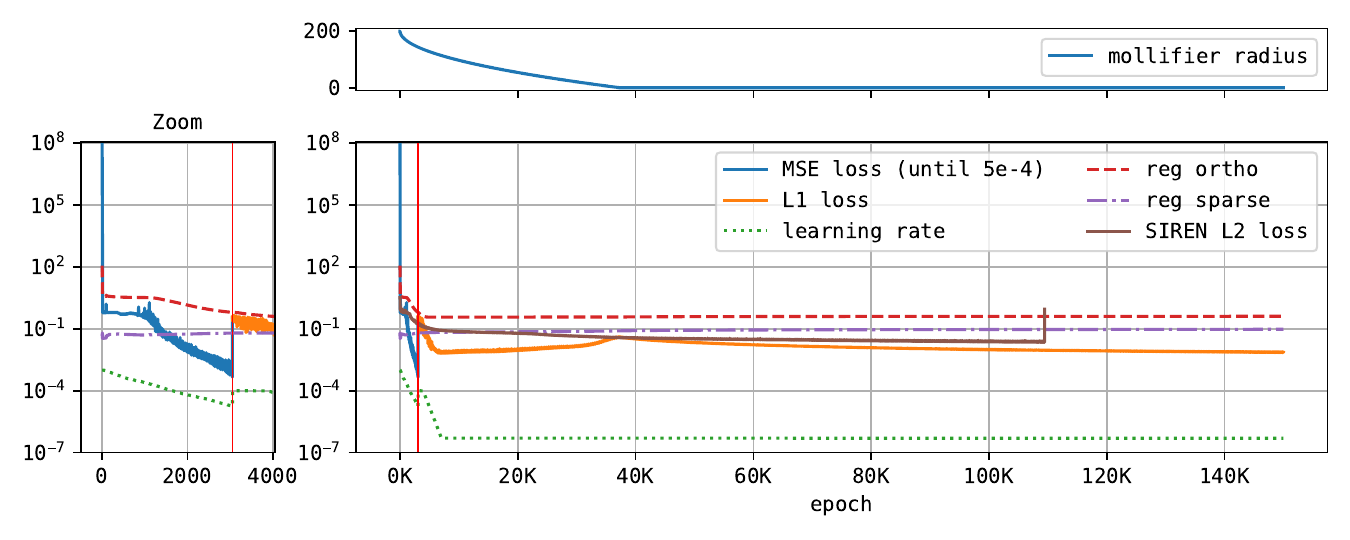}
  \caption{Scaled MSE loss is used until a threshold of {\tt 5e-4} is reached, 
  and then the misfit loss is switched to the $L_1$ loss. Mollifier radius is
  set to decrease linearly according to the schedule given above. A zoom-in near
  the epochs when the misfit function is switched is shown to the left. 
  Orthogonality and sparsity regularizer terms stay relatively flat throughout
  training, except that orthogonality is not strongly enforced until around epoch
  5K. As the mollifier radius is relaxed, the loss increases due to an increase
  in the misfit, however, once the radius is zero, a slow but steady decrease in
  the $\ell_1$ loss is observed. Spurious oscillatory behavior near the shock is
  removed during this period of training.}  
  \label{fig:euler1dloss}
\end{figure}

\subsection{Hypermodes and low-rank tensor decomposition}

We begin by noting that $f_\text{hyper}$ is relatively small
(Tab.~\ref{tab:hyperparams}), so it is feasible to explicitly identify a
low-dimensional linear subspace of its output range.  To this end, we adopt a
standard proper orthogonal decomposition (POD) approach \cite{Berkooz93}.
Sampling $f_{\text{hyper}}$ at $N$ discrete time points, we form a snapshot
matrix with the outputs of $f_{\text{hyper}}$ as columns:
\begin{equation} \label{eq:pod}
  \bfS_{\text{data}}
  =
  \left[
    f_{\text{hyper}}(t_1)
    \mid
    f_{\text{hyper}}(t_2)
    \mid
    \cdots
    \mid
    f_{\text{hyper}}(t_{N-1})
    \mid
    f_{\text{hyper}}(t_{N})
  \right].
\end{equation}
We then compute the SVD of this matrix:
\begin{equation} \label{eq:hypermodes}
  \bfS_{\text{data}} = \bfPhi \bfD \bfPsi^\top.
\end{equation}
Here, the left-singular vectors $\bfphi_i$, which form the columns of $\bfPhi$,
provide a basis for the full space of coefficient vectors spanning both weights
and biases. We refer to these vectors $\bfphi_i$ as \emph{hypermodes}.

For the $j$-th left singular vector $\bfphi_j$ in \eqref{eq:hypermodes}, we
denote the entries corresponding to the coefficients $s^\ell_{1, j}$ and
$s^\ell_{2, j}$ as $\phi^\ell_{1, ij}$ and $\phi^\ell_{2, ij}$, respectively.
After truncating the negligible singular values, the weights and biases can be
expressed as:
\begin{equation} \label{eq:tensor}
  \bfW^\ell(\bc)
  =
  \sum_{i=1}^{r_1^\ell} \sum_{j=1}^{\bar{r}}
    c_j \phi^\ell_{1,ij} \otimes \bu_i^{\ell} \otimes \bv_i^{\ell},
  \qquad
  \bb^\ell(\bc)
  =
  \sum_{j=1}^{r_2^\ell} 
    c_j \phi^\ell_{2,ij} \bb_i^\ell,
  \qquad
  \ell = 1, 2, \ldots, L.
\end{equation}
The third-order tensor term arises naturally in this formulation, 
\begin{equation}
  \bfphi^\ell_{1,i} \otimes \bu_i^{\ell} \otimes \bv_i^{\ell},
  \quad
  i = 1, \ldots, r^\ell_1,
\end{equation}
and the sum
\begin{equation}
  \sum_{i=1}^{r_1^\ell} 
    c_j \phi^\ell_{1,ij} \otimes \bu_i^{\ell} \otimes \bv_i^{\ell}
\end{equation}
constitutes a rank-$r_1^\ell$ tensor decomposition.

Using the truncated left singular matrix $\widehat{\bfPhi}$ from the POD
decomposition, we define a reduced hypernetwork output by projecting onto the
low-dimensional subspace:
\begin{equation} \label{eq:red_hypernetwork}
  \hat{f}_{\text{hyper}} (t)
  =
  \widehat{\bfPhi}\widehat{\bfPhi}^{\top} f_{\text{hyper}} (t).
\end{equation}

In our experiments, we observe a rapid decay in the singular values (a rapid decay
in the diagonal values of $\bfD$).  They are shown in
Fig.~\ref{fig:temporal_svd}.  We know that $\bfPhi$ is necessarily low rank
upon truncation, since the rank of $S_{\text{data}}$ is at most the width of the
hypernetwork.  The time-dynamics is described by the basis $\Psi$, and one sees
that they are very smooth.  The smooth dynamics can also be approximated by
polynomials; degrees up to 30 suffice to approximate all components to machine
precision

\begin{table}
  \centering
  \begin{tabular}{r|l|lllll|lllll|ll}
                  & $r$ 
                  & $r_1^1$ & $r_1^2$ & $r_1^3$ & $r_1^4$ & $r_1^5$ 
                  & $r_2^1$ & $r_2^2$ & $r_2^3$ & $r_2^4$ & $r_2^5$ 
                  & $\bar{r}$
                  \\
                  \hline
                  \hline
    1d Euler     & 150 
                 & 38 & 28 & 94 &  147 &  35 
                 & 71 & 58 & 76 &   67 & 113
                 & 5
                 \\
    2d acoustics & 190 
                 & 64 & 125 & 116 &  14 &
                 & 63 &  85 &  74 &   3 &
                 & 23
                 \\
    2d Burgers   & 170 
                 & 14 & 32  &  113  &  4  &
                 & 11 & 14  &   16  &  7  &
                 & 38
                 \\
    2d advection & 180 
                 & 85 & 160 & 158  & 5 &
                 & 29 &  37 &  20  & 9 & 
                 & 26
                 \\
          \hline
  \end{tabular}
  \caption{Truncated coefficient parameters and number of significant hypermodes \eqref{eq:hypermodes}. The truncation threshold was set to $5 \times 10^{-5}$ after normalizing so that the largest coefficient value equals 1.}
  \label{tab:hypermodes}
\end{table}

\begin{figure}
  \centering
  \includegraphics[width=0.95\textwidth]{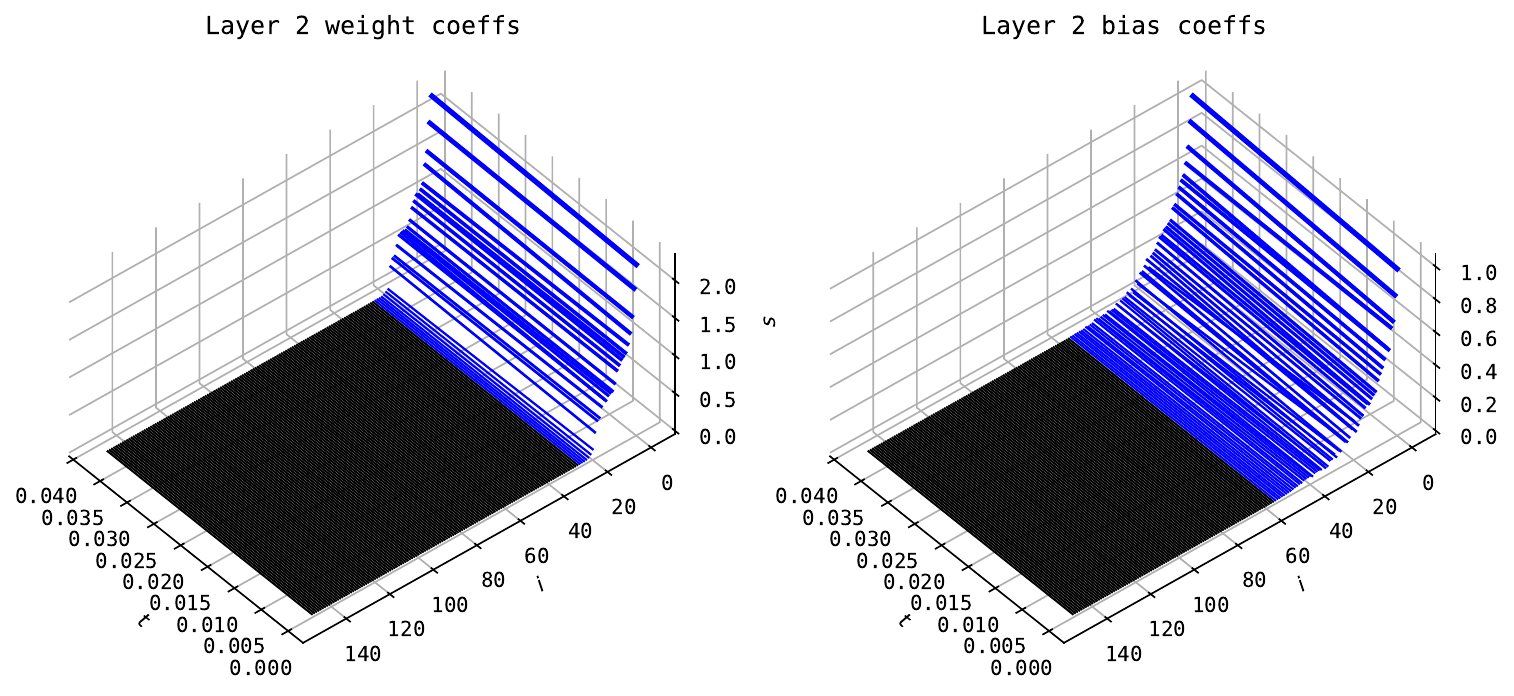}
  \caption{A plot of the coefficients in layer $\ell=2$ for the weights (left)
  and the bias (right). Many of the coefficients are zero, showing that
  truncation of certain terms in the rank-1 expansion \eqref{eq:lowrank}
  can reduce the number of coefficients. The evolution appears smooth and
  in most cases looks like straight lines, making them representable by a 
  few basis functions themselves, leading to the hypermodes
  \eqref{eq:hypermodes}.}
  \label{fig:coeff_evolve}
\end{figure}

\begin{figure}
  \centering
  \begin{tabular}{ll}
    (a) 1d Euler 
    &
    (b) 2d acoustics periodic
    \\
    \includegraphics[width=0.49\textwidth]{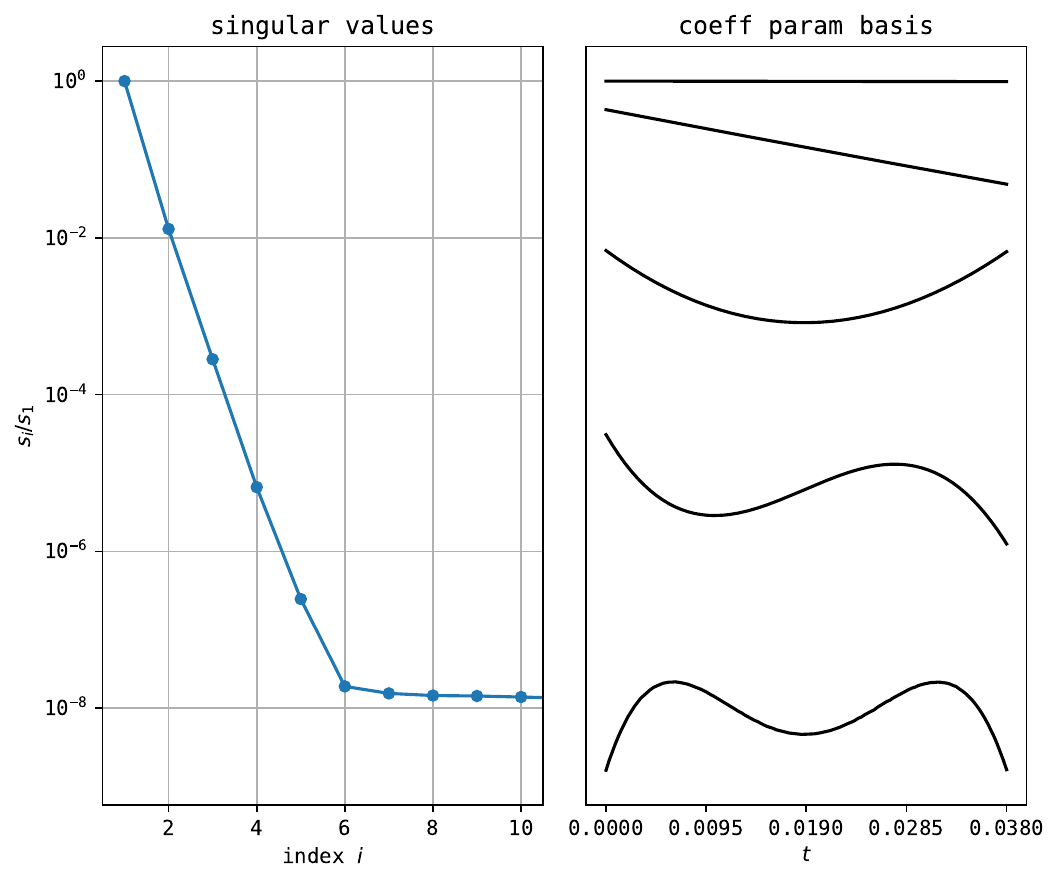}
    &
    \includegraphics[width=0.49\textwidth]{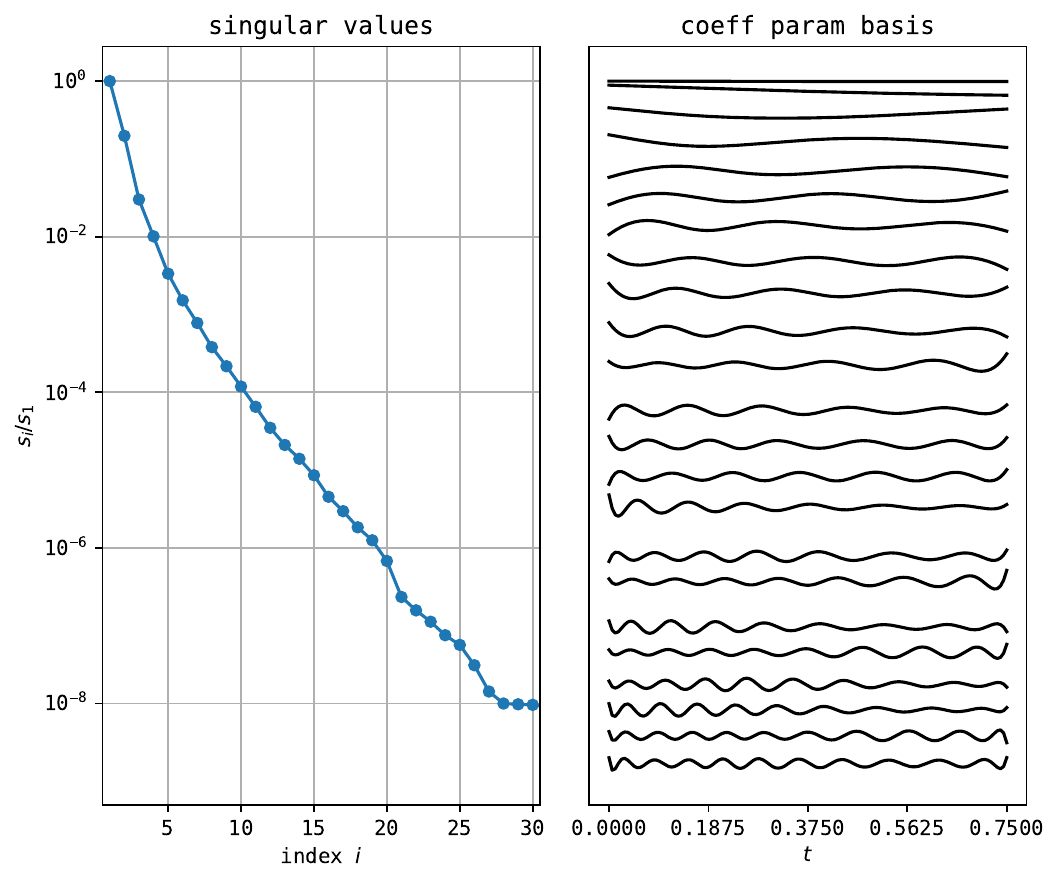}
    \\
    (c) 2d Burgers periodic
    &
    (d) 2d advection swirl
    \\
    \includegraphics[width=0.49\textwidth]{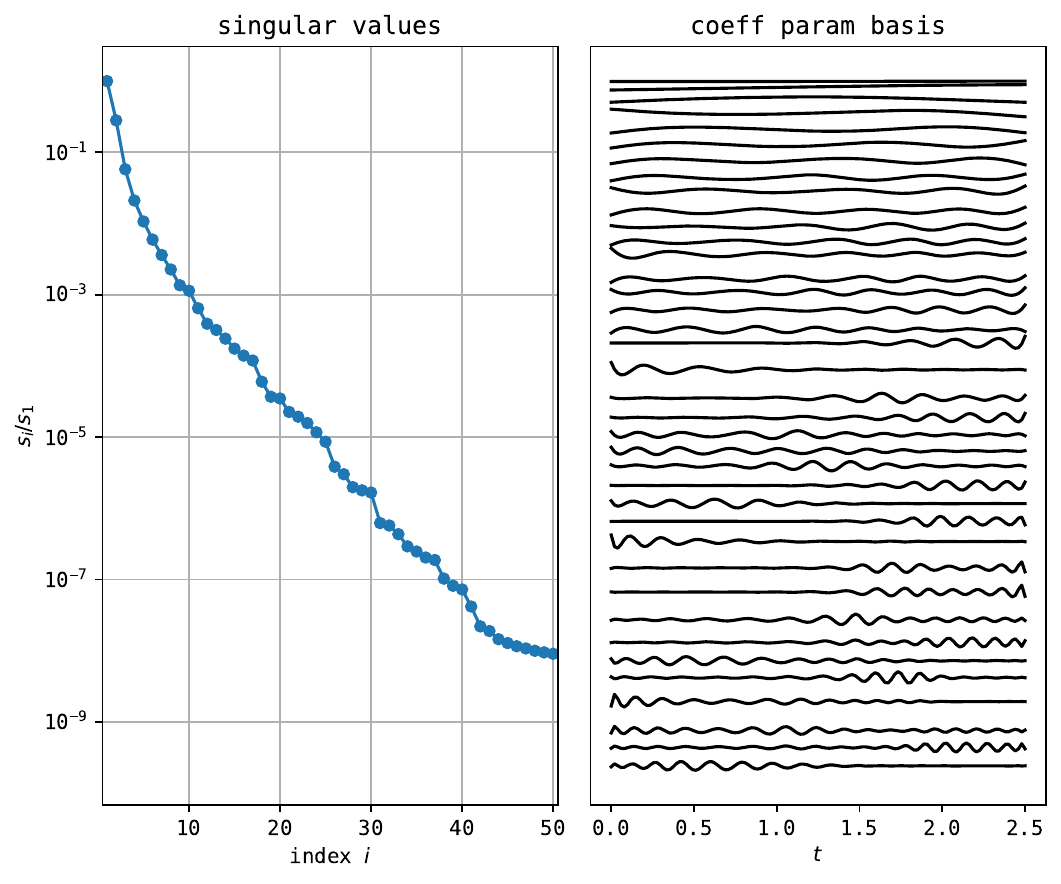}
    &
    \includegraphics[width=0.49\textwidth]{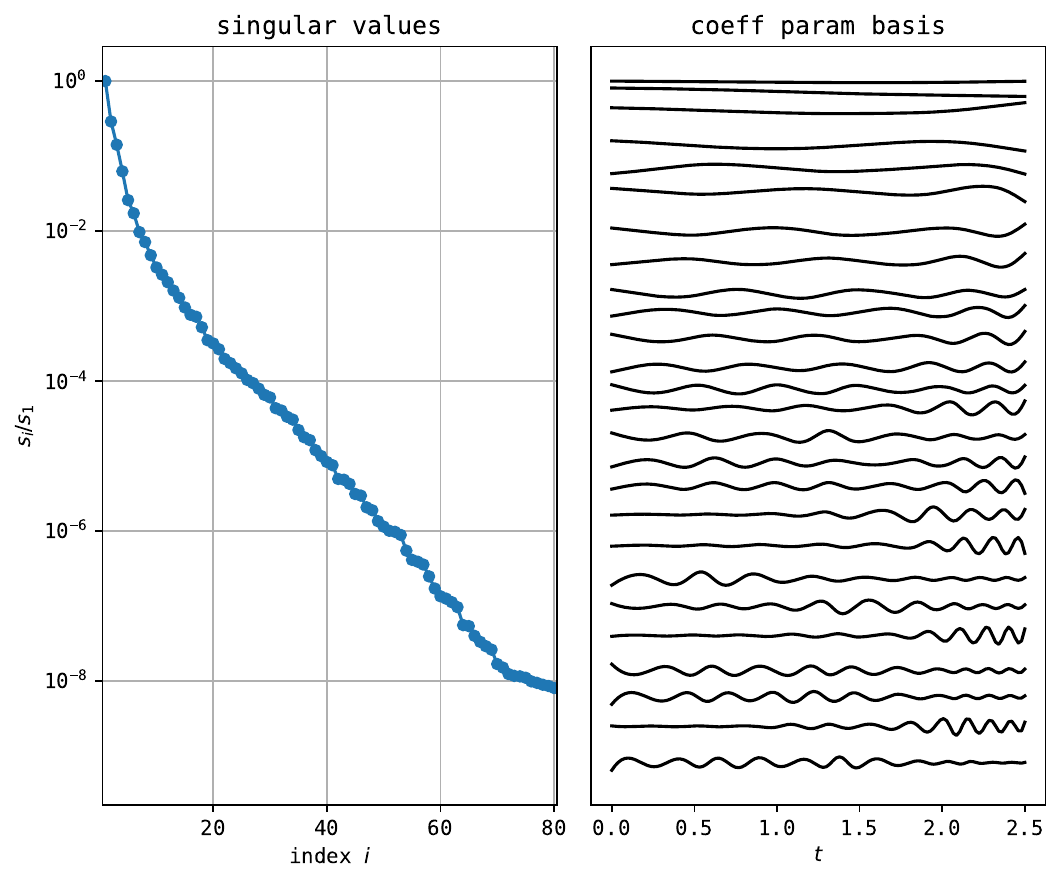}
  \end{tabular}
  \caption{Singular values of the coefficient snapshots over time for the given
  hyperbolic examples. The corresponding right-singular vectors, which form a
  shared temporal basis for the coefficient dynamics, are shown on the right.
  These temporal modes are highly regular. }
  \label{fig:temporal_svd}
\end{figure}

\subsection{Hypermode tangents}

  The tensor decomposition \eqref{eq:tensor} yields a new set of
  coordinates $\bc = (c_i)_{i=1}^{\bar{r}}$, which we refer to as hypermode
  coordinates. A key distinction from the low-rank structure built into the LRNR
  architecture is that this tensor decomposition introduces coupling across
  different layers. In contrast, the low-rank weight matrices in the LRNR
  architecture \eqref{eq:lowrank} enforce low-rank structure independently
  within each layer, without inter-layer dependencies. This is reminiscent of
  the cross-layer coupling in standard hypernetwork formulation.
  
  One can interpret the time dynamics as a curve embedded in the new
  coordinate system, represented by the dynamic hypermode coordinates $\bc(t) =
  (c_i(t))_{i=1}^{\bar{r}}$. At each time $t$, we can examine the effect of
  perturbing this curve along the hypermode directions $\bfphi_i$ and analyze
  how the LRNR responds to these perturbations:
  \begin{equation} \label{eq:tangent}
    u(\bullet\, ; \bs + \delta \bs_i ), \quad \text{where} \quad
    \delta \bs_i = \bfPhi \, \bc(t) + \eta_i \bfphi_i,
    \quad \eta_i \ll 1,
  \end{equation}
  where $\bfPhi$ is the left-singular-vector matrix computed from
  \eqref{eq:hypermodes}, and $\bfphi_i$ is the $i$-th hypermode. The scalar
  $\eta_i$ is selected to ensure that each perturbation produces a visually
  comparable effect in the output. These variations can be viewed as dominant
  modes in the nonlinearly transformed coefficient space.

  \subsubsection{1d Compressible Euler} 
\begin{figure}
  \centering
  \includegraphics[width=0.5\textwidth]{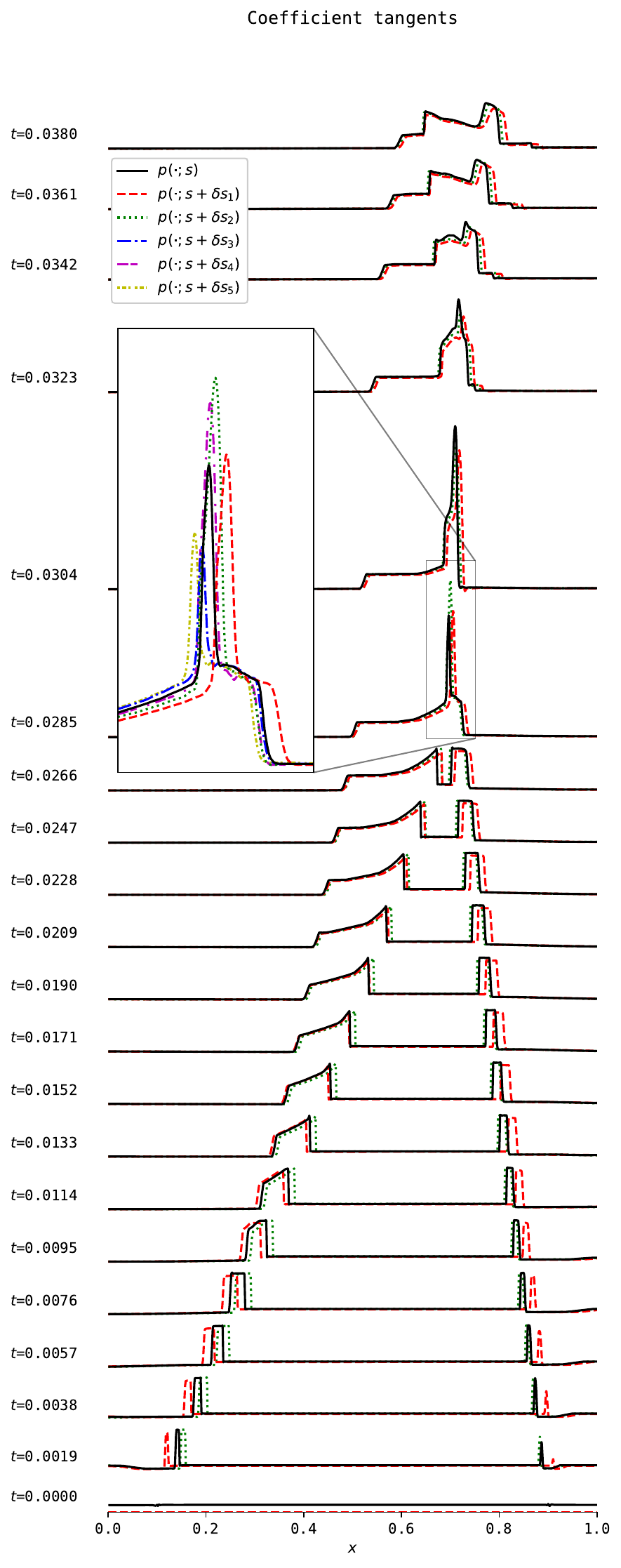}
  \caption{ Perturbations of the LRNR reconstruction along the coefficient
  tangents defined by the hypermodes \eqref{eq:hypermodes}. Shown are the
  effects of the first two hypermode perturbations, including a zoom-in view
  near time $t = 0.0285$, shortly after the shock collision. Each hypermode
  reveals plausible directions of evolution for the solution at different times.
  }
  \label{fig:euler1d_tangent}
  \end{figure}

  The perturbations along dominant hypermodes are shown for the 1d Euler
  equation in Fig.~\ref{fig:euler1d_tangent}. Notably, the first two modes
  govern the evolution of key features in the pressure profile. Perturbing along
  the first mode shifts the two shock locations and slightly increases the
  separation between the peaks. In contrast, the second mode draws the peaks
  closer together. After the collision, the two modes influence the placement of
  transient regions: for example, the first mode significantly shifts the
  left-most inclined slope, whereas the second mode leaves it largely unchanged.
  Overall, the first mode tends to shift the entire profile to the right, while
  the second mode widens the central region.

  Focusing on the time $t = 0.0285$, after the two shocks have merged into
  a single large peak, perturbations in the hypermodes induce physically
  plausible variations in the shape of the peak. The first mode shifts the peak
  and its adjacent slopes to the right. The second mode increases the peak
  height while narrowing its width, and depresses the left-side slope, and pulls
  it
  toward the peak. The third mode narrows and lowers the peak while steepening
  the left slope, leaving the right slope nearly unchanged. The fourth mode
  raises and broadens the peak, steepens the right slope, and introduces a
  discontinuity in the otherwise smooth descent. The fifth mode keeps both left
  and right slopes intact but shifts the peak significantly to the left.

\subsubsection{2d Acoustics with periodic boundary conditions}

\begin{figure}
  \centering
  \includegraphics[width=1.0\textwidth]{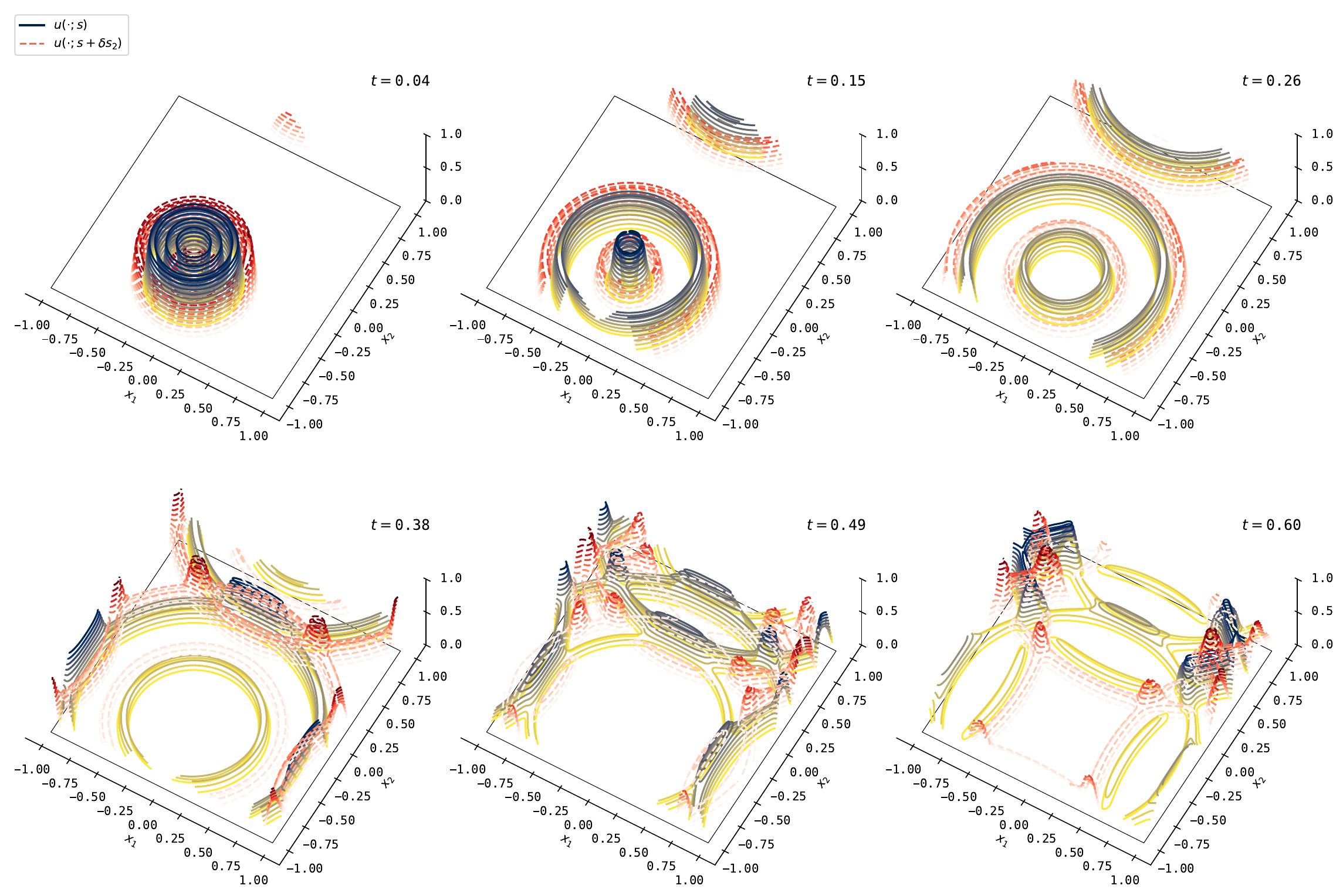}
  \caption{2d acoustics example: LRNR reconstruction (solid contours) and the
  effect of perturbing the coefficients along a selected hypermode direction
  (dashed contours) at various time instances.}
  \label{fig:acoustics2d_time}
\end{figure}

  The perturbations for the 2d acoustics equation are shown in
  Fig.~\ref{fig:acoustics2d_time}. Perturbations along the second hypermode are
  plotted using dashed lines. This mode appears to induce a time-advancing
  effect in the solution. For example, at $t = 0.04$, the perturbation
  anticipates the incoming wave from the upper boundary. At $t = 0.38$, it
  emphasizes the formation of two narrow peaks near the line $x_2 = 0.75$, while
  largely ignoring the development of the broader, oblong peak. At later times,
  the perturbation continues to act primarily on the narrow peaks and leaves the
  flatter, right-angle-aligned wavefronts, those oriented along the axes from
  the origin, mostly unchanged.

\begin{figure}
  \centering
  \includegraphics[width=1.0\textwidth]{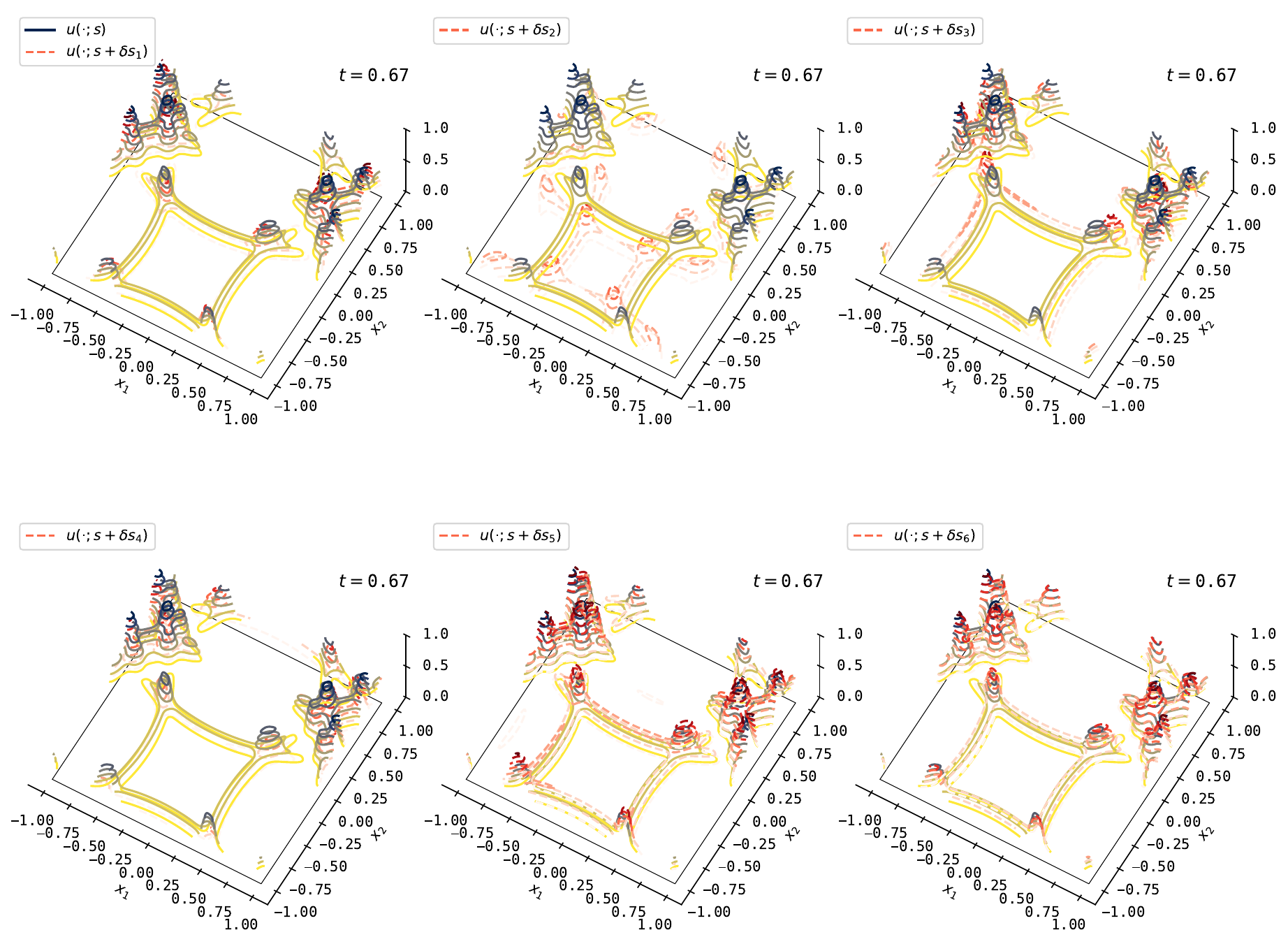}
  \caption{ 2d acoustics example: LRNR reconstruction (solid contours) and the
  effect of coefficient perturbations along various hypermode directions (dashed
  contours) at a fixed time.}
  \label{fig:acoustics2d_modes}
\end{figure}

  Fig.~\ref{fig:acoustics2d_modes} shows the effect of perturbations along
  different hypermodes at a fixed time $t = 0.67$. The first mode shifts the
  entire solution downward uniformly, indicating a global offset in amplitude.
  The second mode roughly corresponds to temporal evolution, as previously
  observed. The third mode primarily transports the connecting contours between
  the peaks. The fourth mode significantly depresses the central region while
  preserving the corner peaks. The fifth mode widens the peaks slightly. The
  sixth mode introduces minor shifts in peak positions; for instance, the two
  upper peaks near $x_2 = 0.50$ move slightly closer to the center.

\subsubsection{2d Burgers with periodic boundary conditions}

\begin{figure}
  \centering
  \includegraphics[width=1.0\textwidth]{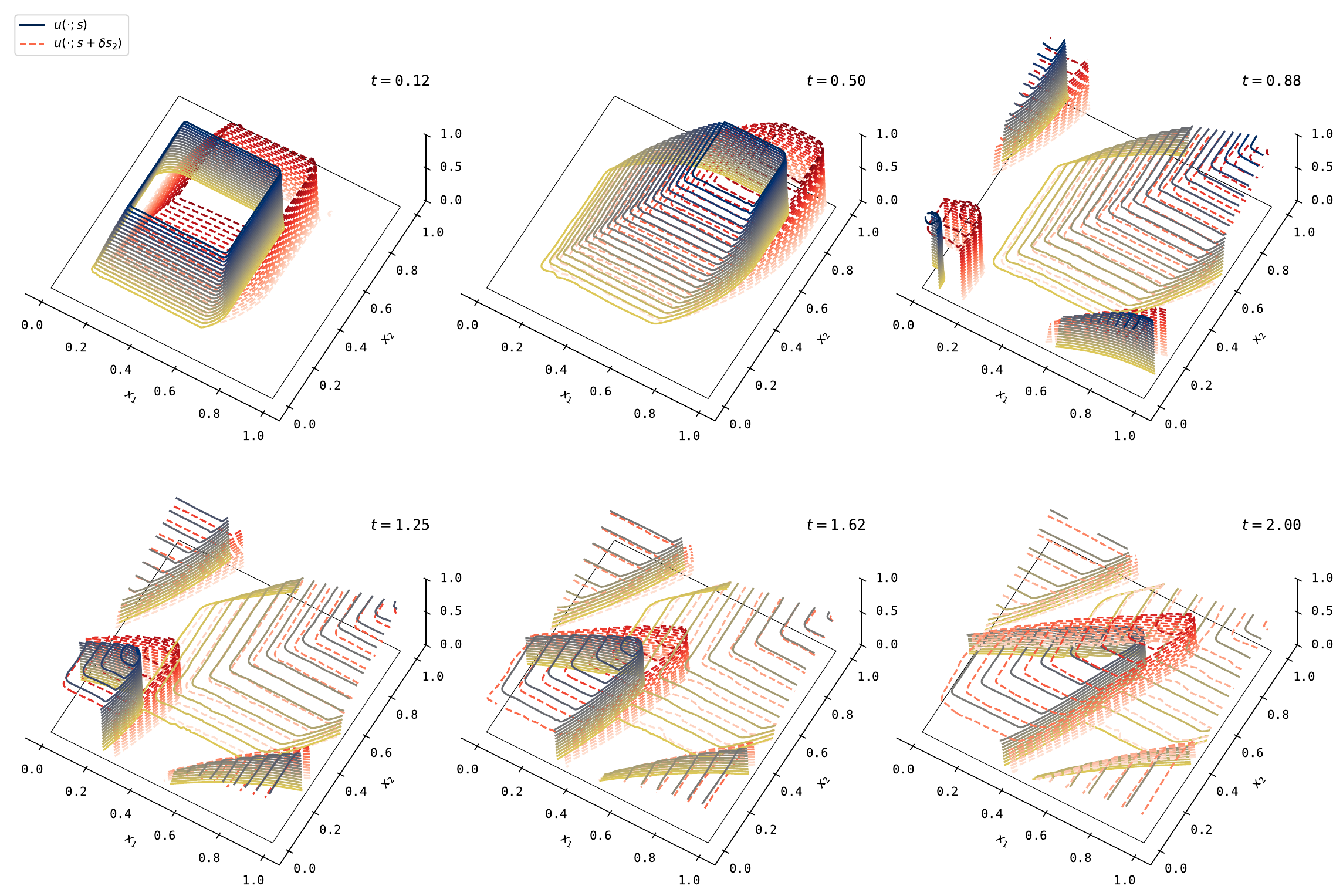}
  \caption{2d Burgers equation example: LRNR reconstruction (solid contours) and
  the effect of coefficient perturbations along a hypermode direction (dashed
  contours) at various times.}
  \label{fig:burgers2d_time}
\end{figure}

The hypermode perturbations for the 2d Burgers equation are shown in
Fig.~\ref{fig:burgers2d_time}. As in the 2d acoustics case, perturbation along
the second hypermode appears to approximate a forward time-evolved solution.
The shock at the upper-right tip becomes sharper, the rarefaction fan shifts
backward, and the incoming profiles at the upper-left and lower-right corners
move inward.

\begin{figure}
  \centering
  \includegraphics[width=1.0\textwidth]{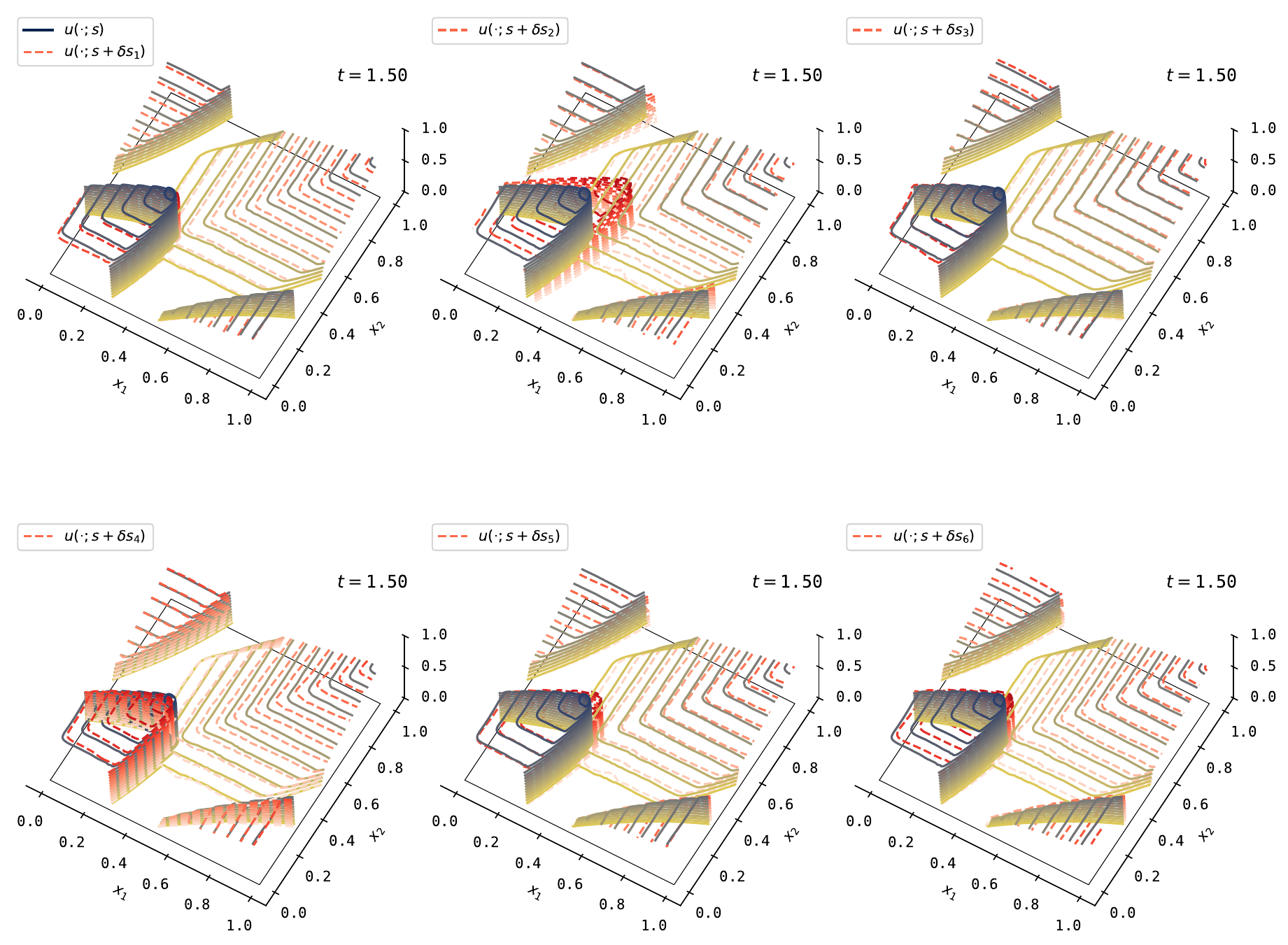}
  \caption{2d Burgers equation example: LRNR reconstruction (solid contours) and
  the effect of coefficient perturbations along various hypermode directions
  (dashed contours) at a fixed time.}
  \label{fig:burgers2d_modes}
\end{figure}

Fig.~\ref{fig:burgers2d_modes} shows the effect of different hypermode
perturbations at time $t = 1.50$. The first mode primarily affects the
rarefaction fans, moving the sloped regions backward while leaving the shocks
mostly unchanged. The second mode corresponds roughly to forward time evolution.
The third mode adjusts the central slope while preserving most of the contour
structure. The fourth mode modifies the slope and shock contours along the
diagonal from the lower-left to upper-right, without significantly affecting the
intruding jumps from the upper-left and lower-right corners. The fifth mode
alters the shape of the shock tip near $(x_1, x_2) = (0.4, 0.4)$ and moves the
slanted intruding profiles downward. In contrast, the sixth mode also modifies
the shock tip but shifts the intruding profiles upward.

\subsubsection{2d Advection with variable speed}

\begin{figure}
  \centering
  \includegraphics[width=1.0\textwidth]{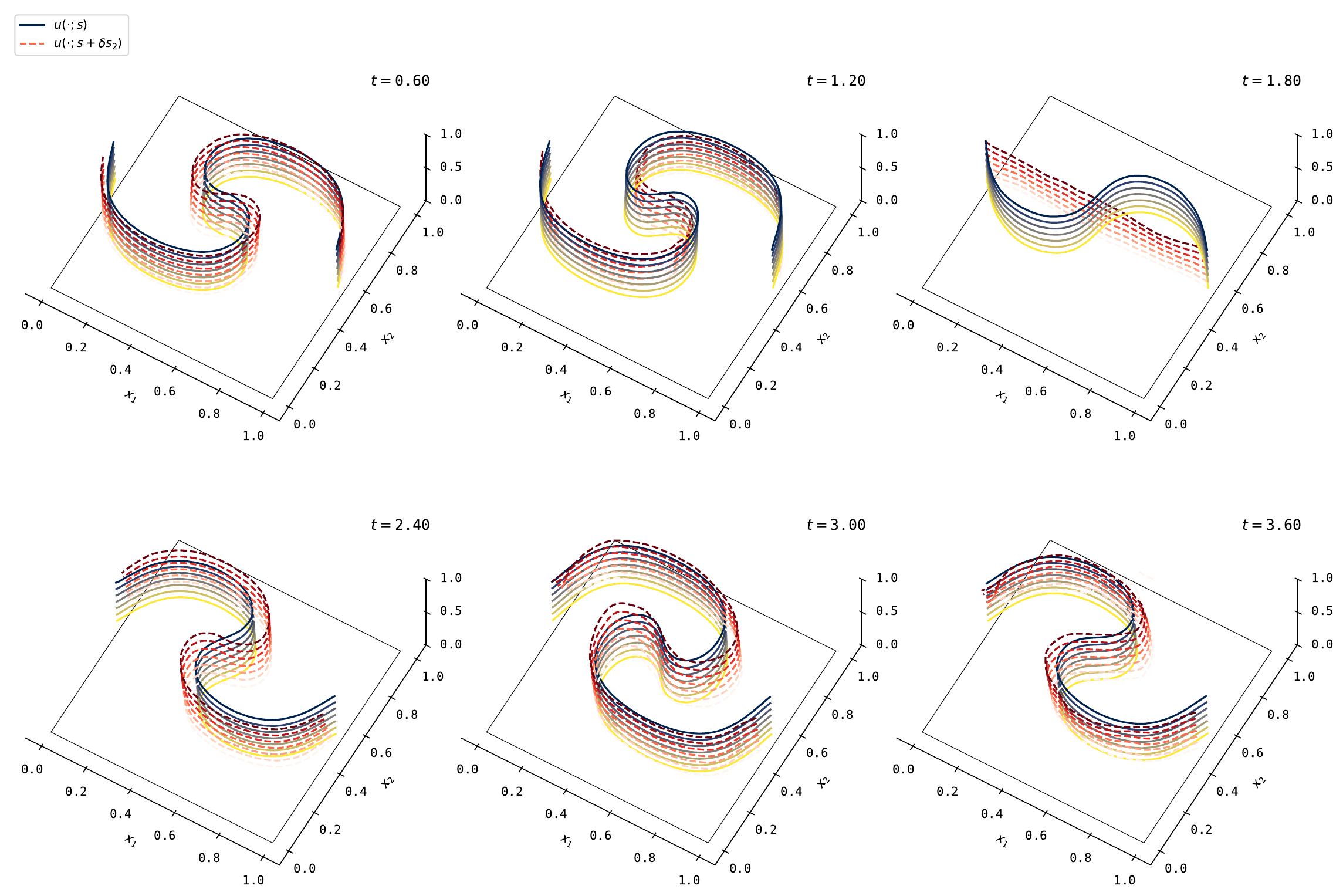}
  \caption{2d advection equation example: LRNR reconstruction (solid contours)
  and the effect of coefficient perturbations along a hypermode direction
  (dashed contours) at various times.}
  \label{fig:advection2d_time}
\end{figure}

The hypermode perturbations for the 2d advection equation are shown in
Fig.~\ref{fig:advection2d_time}. Perturbation along the second hypermode
(similar to the acoustics 2d case) appears to approximate a forward time-evolved
solution, generally twisting the curved interface in the advection direction. At
time $t = 1.80$, it even predicts the return to the initial condition, where the
interface lies along the straight line $x_2 = 0.5$.

\begin{figure}
  \centering
  \includegraphics[width=1.0\textwidth]{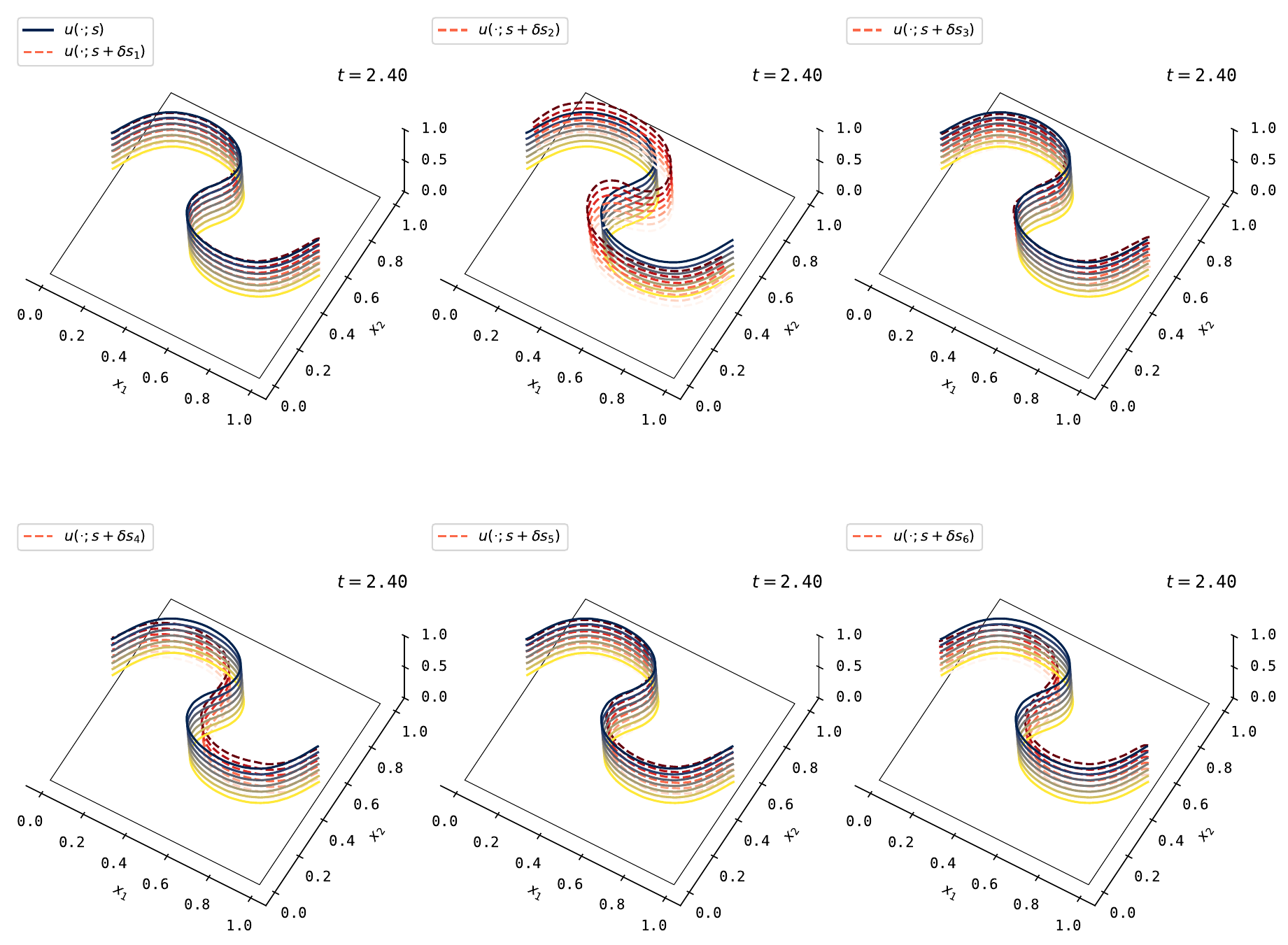}
  \caption{2d advection equation example: LRNR reconstruction (solid contours)
  and the effect of coefficient perturbations along various hypermode directions
  (dashed contours) at a fixed time.}
  \label{fig:advection2d_modes}
\end{figure}

Fig.~\ref{fig:advection2d_modes} shows the effect of different hypermode
perturbations at time $t = 2.40$. The first mode corresponds to a slight up-down
shift of the contours. The second mode again resembles forward time evolution.
The third mode primarily adjusts the vertical heights of the contours, with a
minor twist in the up-down direction. The fourth mode modifies the shape of the
twisting interface, particularly the portion in the upper-left region of the
domain. The fifth mode similarly adjusts the interface, this time focusing on
the lower-right region. The sixth mode slightly forward-evolves the solution
while also altering the curve's vertical position, pushing the upper-left
portion downward and the lower-right portion slightly upward.

\subsection{Hypermode extrapolation}

The coefficient tangent directions given by the hypermodes \eqref{eq:tangent}
appear to capture dominant directions that the solution dynamics tends to
follow. Interestingly, however, these hypermodes continue to produce salient
features even when the coefficient vectors are perturbed significantly beyond
their training-time trajectories. As the perturbation magnitude increases, the
coefficients move outside the range seen during training, resulting in wave
profiles that were never observed in the actual solution trajectory.

This behavior can be interpreted as a form of coefficient extrapolation in the
hypermode directions, extending beyond the local tangents near the solution
trajectory. Specifically, we consider traversing further along the hypermode
perturbation directions \eqref{eq:hypermodes}, as given by:
\begin{equation} \label{eq:extrap}
  u(\bullet\, ; \bs + \delta \bs_i ), \quad \text{where} \quad
  \delta \bs_i = \bfPhi \bc(t) + \eta_i \bfphi_i \bc_i'(t), \quad \eta_i \sim 1.
\end{equation}
Unlike many convolutional autoencoder models, where large extrapolations from
training data often yield outputs lacking coherent structure, these
extrapolations in the LRNR setting still produce reconstructions that retain
physically meaningful features.

\begin{figure}
  \centering
  \includegraphics[width=1.0\textwidth]{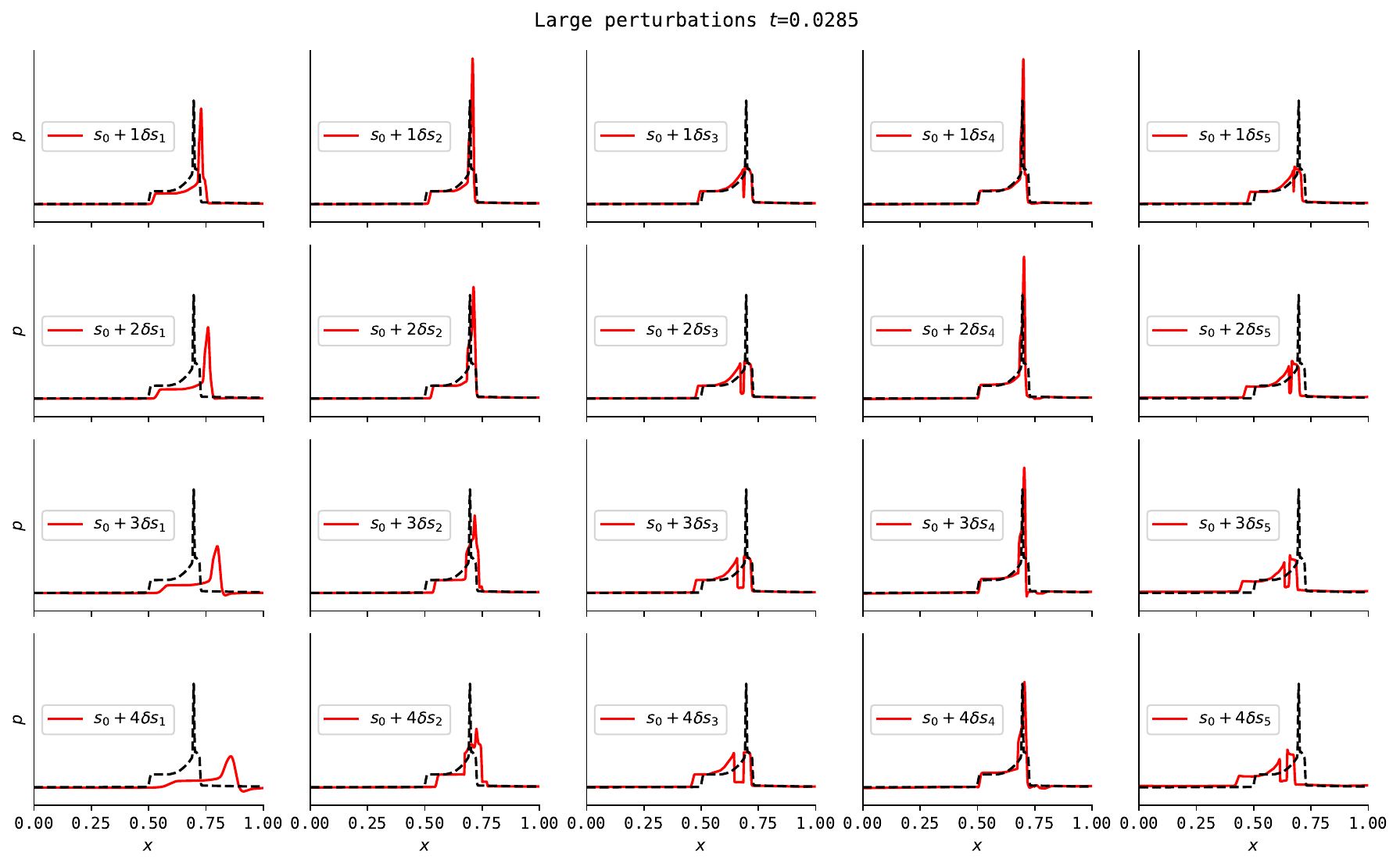}
  \caption{Coefficient extrapolation (large perturbations) along the hypermode
  directions for the 1d Euler example, starting from the reconstruction at time
  $t = 0.0285$. Each column shows the effect of perturbing along a dominant
  hypermode. The same hypermodes and base time were used in
  Fig.~\ref{fig:euler1d_tangent}.}
  \label{fig:euler1d_extrap}
\end{figure}

We illustrate that, for LRNRs, extrapolation remains meaningful even far beyond
the local tangent space. This is demonstrated in the 1d Euler example shown in
Fig.~\ref{fig:euler1d_extrap}. Extrapolation along the first hypermode largely
translates the profile to the right while simultaneously smoothing the sharp
features. Along the second hypermode, the peak height decreases, and the
previously gradual slope on the left is sharpened into a jump discontinuity. The
third mode produces a profile reminiscent of an earlier, pre-collision
state, removing the central peak and splitting the wave into two distinct
components. The fourth hypermode initially sharpens and narrows the peak, but
also introduces a kink along the left slope and transforms the right side into a
nearly linear descent. Extrapolation along the fifth mode yields a similar split
behavior to the third mode, but with the resulting profile shifted to the right
and the left peak significantly smaller than the right.
  
These extrapolated profiles bear little resemblance to any solution data on
which the LRNR was trained, yet they retain coherent structures that appear
physically plausible. Notably, even when deviating significantly from the
trajectory $t \mapsto \bc(t)$ defined by the hypernetwork outputs, the resulting
profiles do not exhibit random or noisy artifacts.

\section{FastLRNR and efficient evaluation}

A key limitation of INRs in practical applications is their slow inference
speed. For example, in SIREN hypernetworks, the INR typically consists of fully
connected layers with widths exceeding $2^9 = 512$. Even with such moderately
sized networks, evaluating the output at a single point involves computations
that scale with network width, which becomes prohibitively expensive in
performance-critical applications such as real-time rendering, as discussed in
well-known works \cite{muller2022,Kerbl2023}.  We highlight another advantage of
the LRNR architecture: its low-rank structure enables accelerated evaluation and
backpropagation \cite{cho2025}. Here, we focus on the forward evaluation
(inference) at a single spatial point, which is the fundamental operation
underlying reconstruction.

\subsection{FastLRNR construction}

  A distinguishing feature of LRNRs is that at each layer, the output from
  the affine mapping  
  \begin{equation}
  \begin{aligned}
    \by^\ell &= \bfW^\ell \bz^{\ell -1} + \bb^\ell \\
          &= \bfU^\ell \, \diag(\bs^\ell_1)\, \bfV^{\ell \top} \bz^{\ell -1} + \bfB^\ell \bs^\ell_2 \\
          &= 
          \begin{bmatrix} \bfU^\ell & \bfB^\ell \end{bmatrix}
          \begin{bmatrix} \bfzeta^\ell \\ \bs^\ell_2 \end{bmatrix},
      \qquad \qquad
      (\bfzeta^\ell := \diag(\bs^\ell_1) \bfV^{\ell \top} \bz^{\ell - 1})
  \end{aligned}
  \end{equation}
  has rank at most $r^\ell_1 + r^\ell_2$. When a scalar nonlinearity is applied to this output,
  \begin{equation}
    \bz^\ell = \sigma\left(
      \begin{bmatrix} \bfU^\ell & \bfB^\ell \end{bmatrix}
      \begin{bmatrix} \bfzeta^\ell \\ \bs^\ell_2 \end{bmatrix}
    \right),
  \end{equation}
  the result $\bz^\ell$ may also lie in a low-rank linear subspace, provided the
  nonlinearity $\sigma$ is smooth. This property, often referred to as
  \emph{affine decomposition}, has long been leveraged in classical
  reduced-order models~\cite{Hesthaven2016}, where one writes
  \begin{equation}
    \bz^\ell = \bfXi^\ell \bq^\ell, \qquad \bfXi^\ell \in \RR^{M_\ell \times \hat{r}^\ell}
  \end{equation}
  for some $\hat{r}^\ell \approx r^\ell_1 + r^\ell_2$. That is, the output
  $\bz^\ell$ lies approximately in the column space $\text{colspan}(\bfXi^\ell)$
  of dimension $\hat{r}^\ell$.  Although the basis matrix $\bfXi^\ell$ is not
  known a priori, it can be constructed empirically by sampling $\bz^\ell$ for
  different values of the spatial input $\bx$ or the coefficient vector $\bs$.

  Although $\bq^\ell$ is also not known, it can be computed by solving a
  linear system. Since the system $\bfXi^\ell \bq^\ell = \bz^\ell$ is typically
  overdetermined, we solve a reduced system by sampling a subset of the rows
  (assuming consistency):
  \begin{equation}
    \bq^\ell = \left(\bfP^{\ell \top} \bfXi^\ell \right)^{-1} \bfP^{\ell \top} \bz^\ell
  \end{equation}
  where $\bfP^\top$ is a sampling matrix that selects (or truncates) rows from
  the product. This leads to the reduced reconstruction:
  \begin{equation}
    \bz^\ell 
    = 
    \bfXi^\ell \left(\bfP^{\ell \top} \bfXi^\ell \right)^{-1} \bfP^{\ell \top} \sigma(\by^\ell)
    = 
    \bfXi^\ell \left(\bfP^{\ell \top} \bfXi^\ell \right)^{-1} \sigma(\bfP^{\ell \top} \by^\ell).
  \end{equation}
  Taking into account the low-rank structure of the weight matrix in the next layer, we compute
  \begin{equation}
    \bfV^{(\ell + 1)\top} \bz^\ell
    = 
    \bfV^{(\ell + 1)\top} \bfXi^\ell 
      \left(\bfP^{\ell \top} \bfXi^\ell \right)^{-1} \sigma(\bfP^{\ell \top} \by^\ell).
  \end{equation}
  In practice, the inverse is typically avoided and replaced with a linear
  solve. Notably, this formulation eliminates dependence on the full width
  $M_\ell$ in the matrix products.
  Note that the matrix dimensions are:
  \begin{equation}
  \begin{aligned}
    \bfV^{(\ell + 1)\top} &\in \RR^{r^{\ell+1}_1 \times M_\ell}, 
    & \bfXi^\ell &\in \RR^{M_\ell \times \hat{r}^\ell}, \\
    \bfP^{\ell \top} &\in \RR^{\hat{r}^\ell \times M_\ell}, 
    & \begin{bmatrix} \bfU^\ell & \bfB^\ell \end{bmatrix} &\in \RR^{M_\ell \times (r^\ell_1 + r^\ell_2)}.
  \end{aligned}
  \end{equation}
  We define compressed matrices:
  \begin{equation}
  \begin{aligned}
    \widehat{\bfV}^{(\ell + 1)\top} &:= 
    \bfV^{(\ell + 1)\top} \bfXi^\ell \left(\bfP^{\ell \top} \bfXi^\ell \right)^{-1}
    \in \RR^{r^{\ell+1}_1 \times \hat{r}^\ell}, \\
    \widehat{\bfU}^{\ell} &:= \bfP^{\ell \top} \bfU^\ell \in \RR^{\hat{r}^\ell \times r^\ell_1}, \\
    \widehat{\bfB}^{\ell} &:= \bfP^{\ell \top} \bfB^\ell \in \RR^{\hat{r}^\ell \times r^\ell_2}.
  \end{aligned}
  \end{equation}
  Thus, the reduced representation defines a smaller LRNR whose width is now given by $\hat{r}^\ell$ and $r^\ell_1 + r^\ell_2$. We call this compressed model \textbf{FastLRNR}, denoted by $\hat{u}$. It is expressed as
  \begin{equation}
  \begin{aligned}
    \bz^{\ell}
    &= 
    \sigma\left( 
      \widehat{\bfU}^\ell \, \diag(\bs^\ell_1) \, 
      \widehat{\bfV}^{\ell \top} \bz^{\ell-1}
      + \widehat{\bfB}^\ell \bs^\ell_2 \right), \\
    \bz^L 
    &= 
    \widehat{\bfU}^L \, \diag(\bs^L_1) \, \bz^{L-1} + \bb^L.
  \end{aligned}
  \end{equation}
Recall that $\bs_1^\ell$, $\bs_2^\ell$ are truncated coefficients projected onto hypermodes via the reduced hypernetwork $\hat{f}_{\text{hyper}}$.

\subsection{Point evaluation}

We consider the application of FastLRNR construction above for
pointwise evaluation.  Specifically, we aim to evaluate the output of the
network at a spatial point $\bx$ at various times $t$
\begin{equation} \label{eq:eval}
  u(\bx ; \hat{f}_{\text{hyper}}(t)) \quad \text{for} \quad t \in [0, T].
\end{equation}

To compute $u$, all intermediate hidden states $\bz^\ell$ must be evaluated
layer by layer. The dominant computational cost arises from matrix-vector
multiplications, particularly the multiplications by the weight matrices in
each layer. This makes the overall evaluation complexity heavily dependent on
the width and depth of the network.

Throughout this procedure, we fix a spatial point $\bx$ within the domain
and estimate the basis $\bfXi^\ell$ for each hidden state $\bz^\ell$ by sampling
its values over different time instances $t$. Specifically, we form the
snapshot matrix
\begin{equation}
  \left[
    \bz^\ell(\bx; f_{\text{hyper}}(t_1))
    \mid 
    \bz^\ell(\bx; f_{\text{hyper}}(t_2))
    \mid 
    \cdots
    \mid 
    \bz^\ell(\bx; f_{\text{hyper}}(t_{N}))
  \right],
\end{equation}
and compute its singular value decomposition (SVD) to obtain the basis matrix
$\bfXi^\ell$.
Next, we apply the Empirical Interpolation Method (EIM)
\cite{barrault04,chanturantabut10} to determine the interpolation (or
sampling) matrix $\bfP^\ell$. With these components in place, we construct the
FastLRNR model, which we then use to efficiently approximate the evaluation
\eqref{eq:eval}.

\begin{figure}
  \centering
  \begin{tabular}{ll}
    (a) 1d Euler 
    &
    (b) 2d acoustics periodic
    \\
  \includegraphics[width=0.5\textwidth]{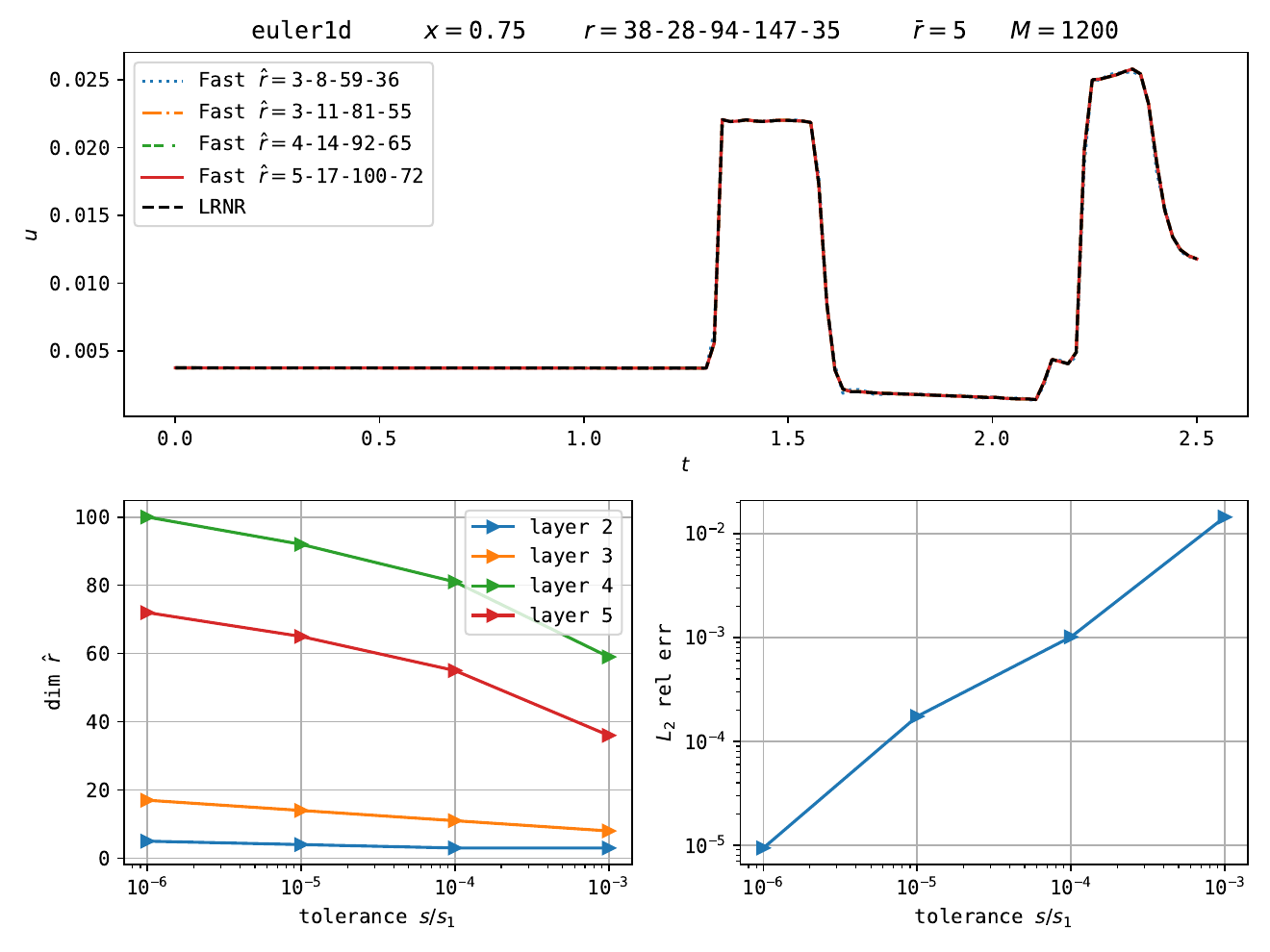}
  &
  \includegraphics[width=0.5\textwidth]{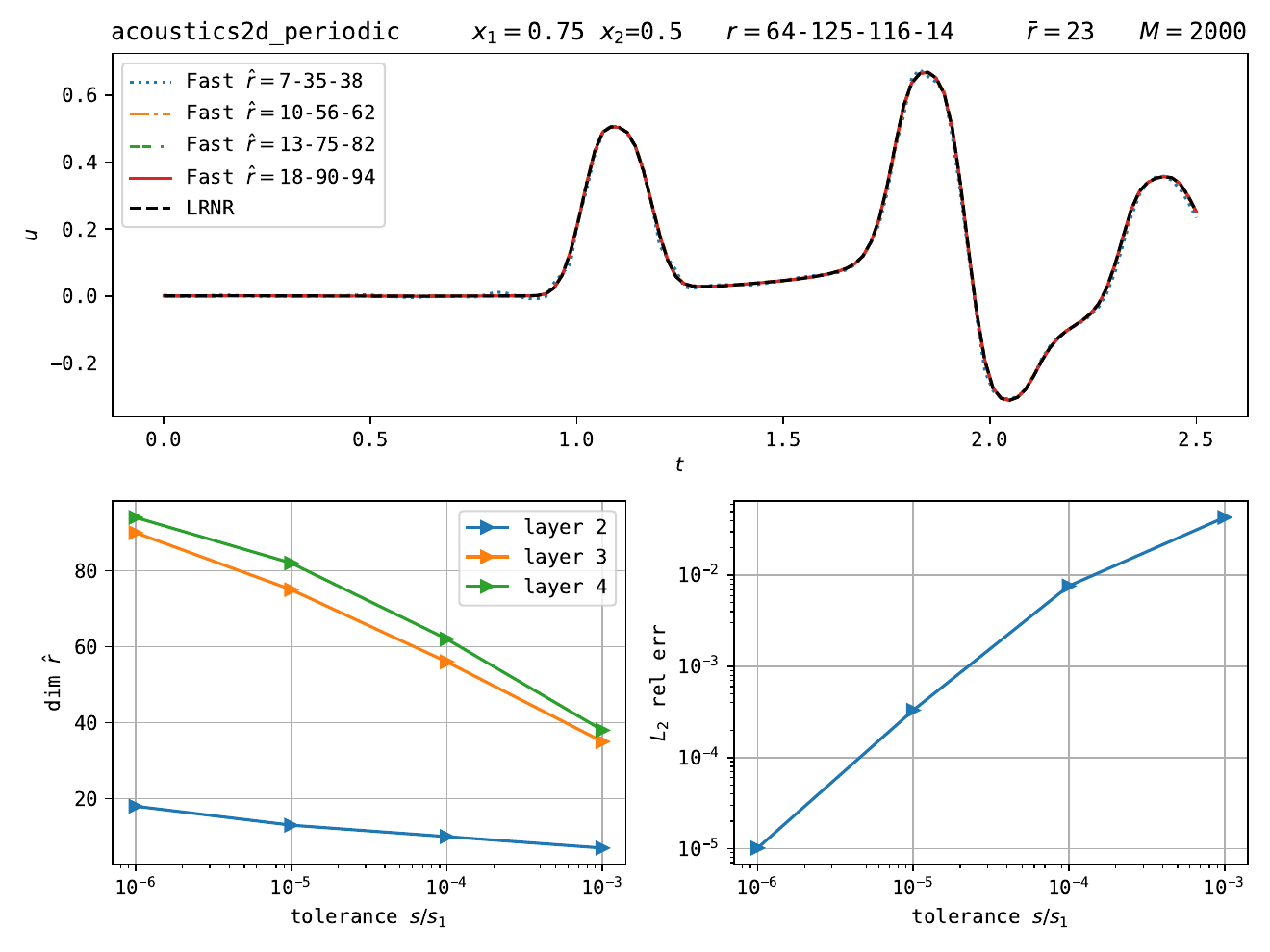}
  \\
  (c) 2d Burgers periodic
  &
  (d) 2d advection swirl
  \\
  \includegraphics[width=0.5\textwidth]{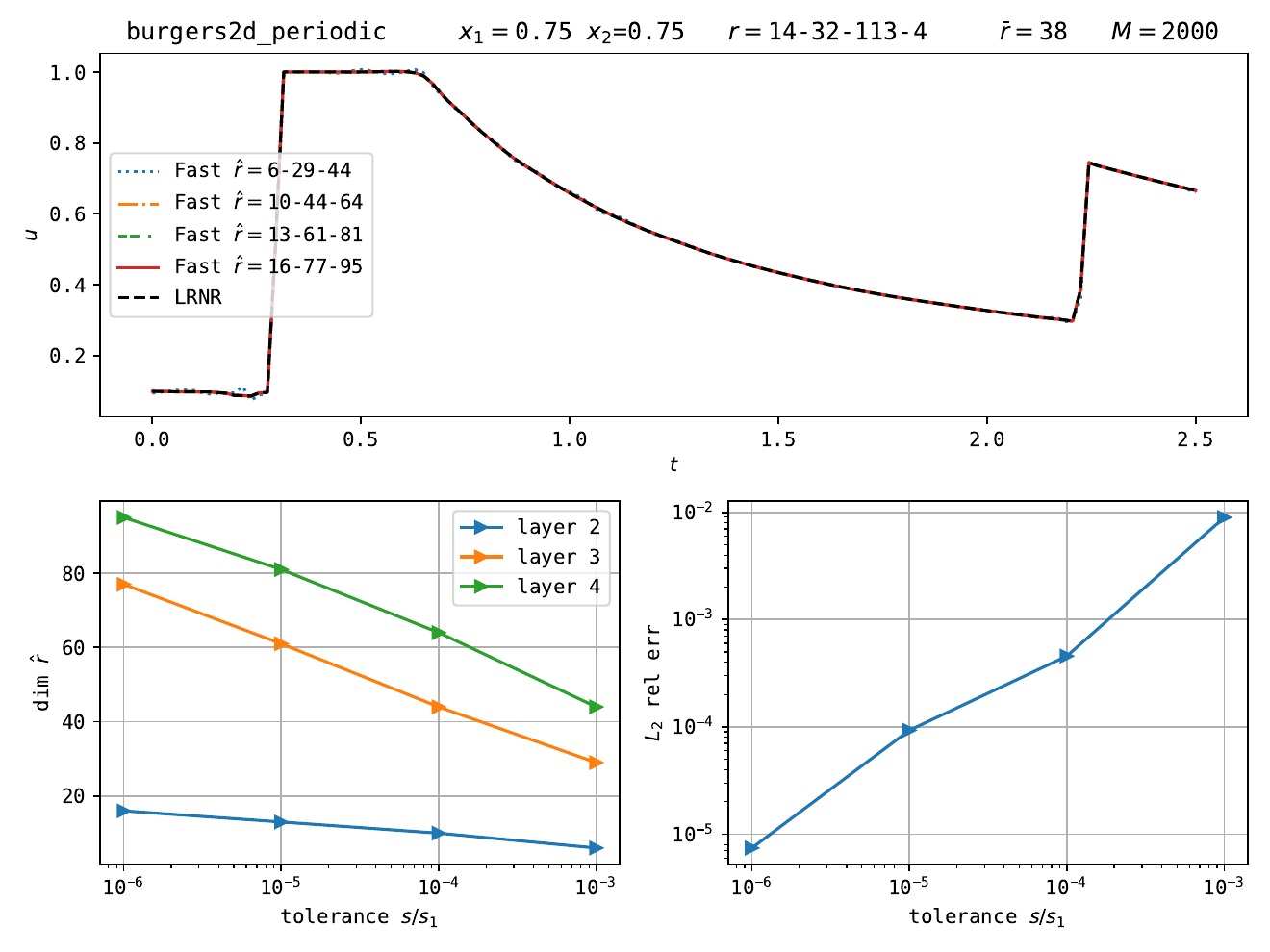}
  &
  \includegraphics[width=0.5\textwidth]{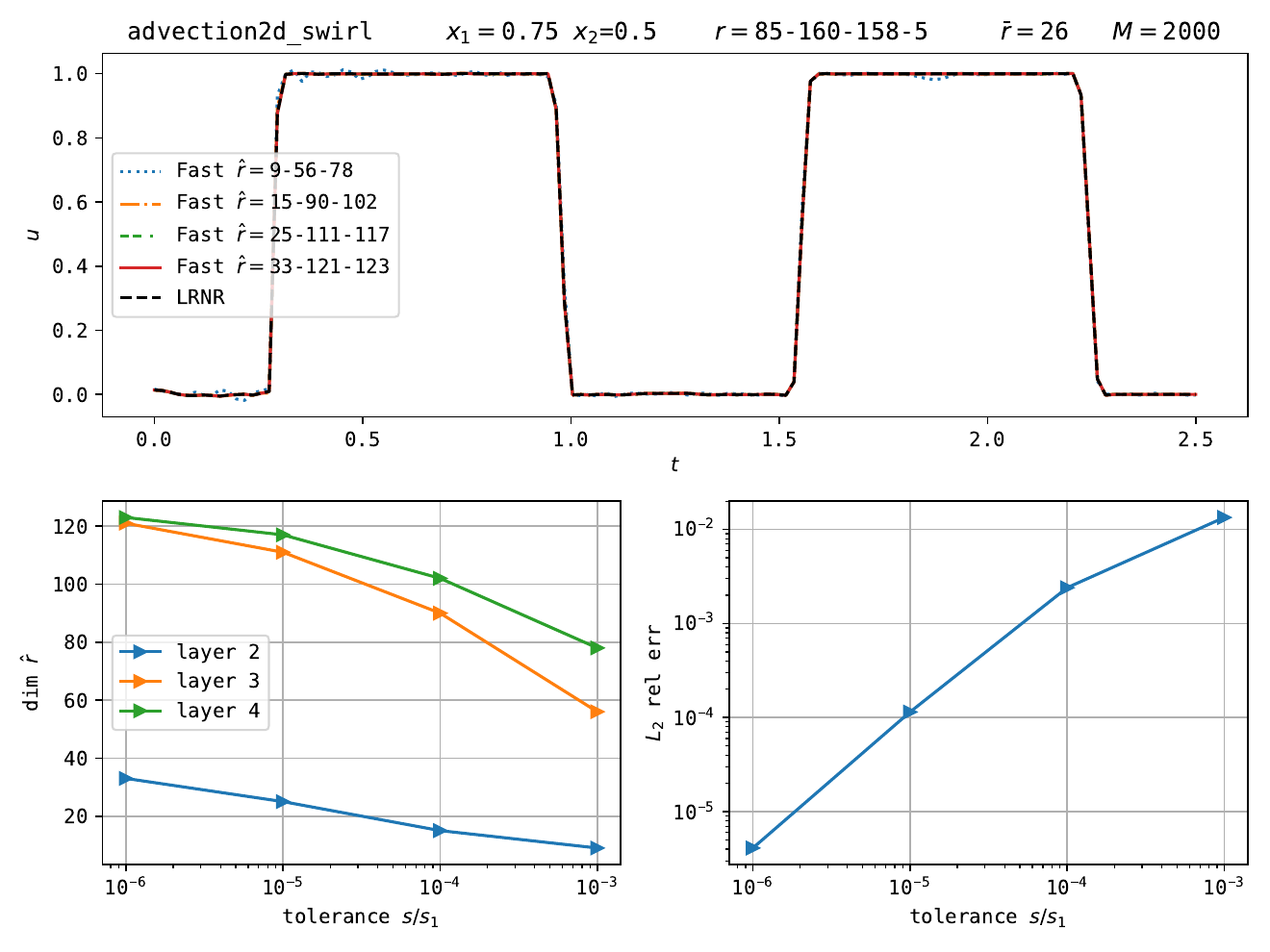}
  \end{tabular}
  \caption{FastLRNR evaluation results for varying EIM ranks. As the EIM rank
  increases, the $\ell_2$ relative error consistently decreases, reaching below
  $10^{-4}$ when compared to full LRNR evaluation. The EIM rank $\hat{r}^\ell$
  remains comparable to the rank of the weight matrices $r_1^\ell$ in the
  examples.}
  \label{fig:eval}
\end{figure}

The results for our examples are shown in Fig.~\ref{fig:eval}. The
reconstructed solutions from FastLRNR agree well with those of the full model,
despite FastLRNR using a significantly smaller architecture. As a result, it
can be evaluated much more efficiently, with the number of floating-point
operations scaling with the reduced width of the FastLRNR. According to recent
work in the INR literature, such network sizes approach the regime required
for real-time rendering applications \cite{muller2022,Kerbl2023}.

The FastLRNR construction relies on the fact that the hidden units $\bz^\ell$
above lie in a low-rank subspace, a property that is unique to LRNRs:
Low-rankness in the hidden units is not observed in other techniques enforcing
low-rankness in the weights \cite{jaderberg2014speeding,hu2022lora,deepseekai2025}, or
in other hypernetwork approaches \cite{ha2016hypernetworks,Sitzmann2020,Pan2023}.

We remind the reader that FastLRNR can also be used to approximate
backpropagation operations for solving partial differential equations
\cite{cho2025}, thereby enabling its integration with numerical solvers. This
opens the door for FastLRNRs to be employed not only for efficient inference
but also as components in physics-informed and hybrid computational
frameworks.

\section{Spatial extrapolation in periodic boundary conditions}

For linear low-dimensional representations, handling incoming boundary
conditions, particularly where wave profiles enter the domain, remains a
challenging problem, primarily due to the slow decay of the Kolmogorov
$n$-width. However, in two of our examples, the meta-learning framework appears
to emulate periodic boundary conditions without introducing visible artifacts. A
plausible explanation for how LRNRs manage boundaries is as follows: the
low-rank structure of LRNRs promotes an economical reconstruction of the
solution, effectively prioritizing physically plausible wave propagation. This
behavior aligns with the nature of hyperbolic waves, where information is
transported along characteristic curves. The resulting reconstruction
approximately respects causality and maintains information flow within the
domain of influence.

To investigate how the trained LRNRs approximate boundary values, we evaluate
the LRNR over a spatially extended domain. Since the LRNR was not exposed to any
data outside the original spatial domain during training, this evaluation
amounts to spatial extrapolation. The results for the 2d Burgers and acoustics
equations are shown in Fig.~\ref{fig:boundary}. Remarkably, the LRNR continues
to exhibit coherent and physically plausible features even beyond the training
domain.

\subsection{2d acoustics equation}

\begin{figure}
  \centering
  \begin{tabular}{ll}
  \includegraphics[width=1.0\textwidth]{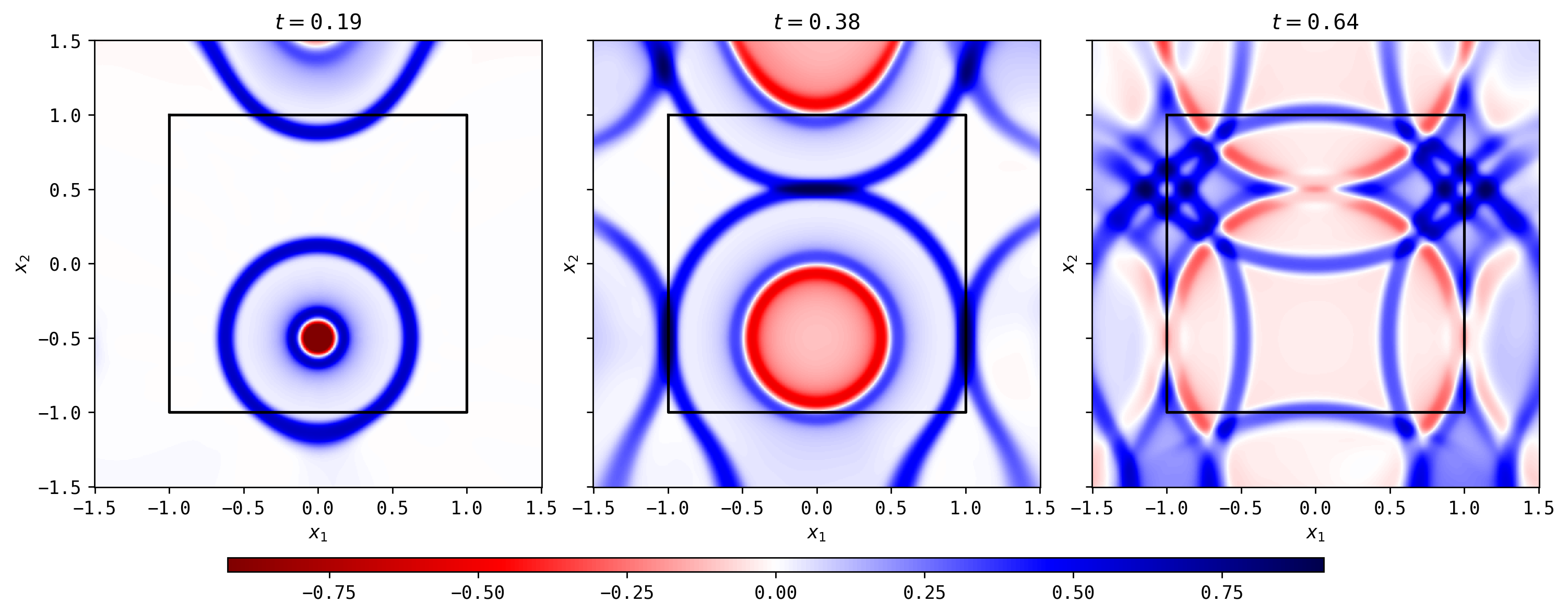}
  \\
  \includegraphics[width=1.0\textwidth]{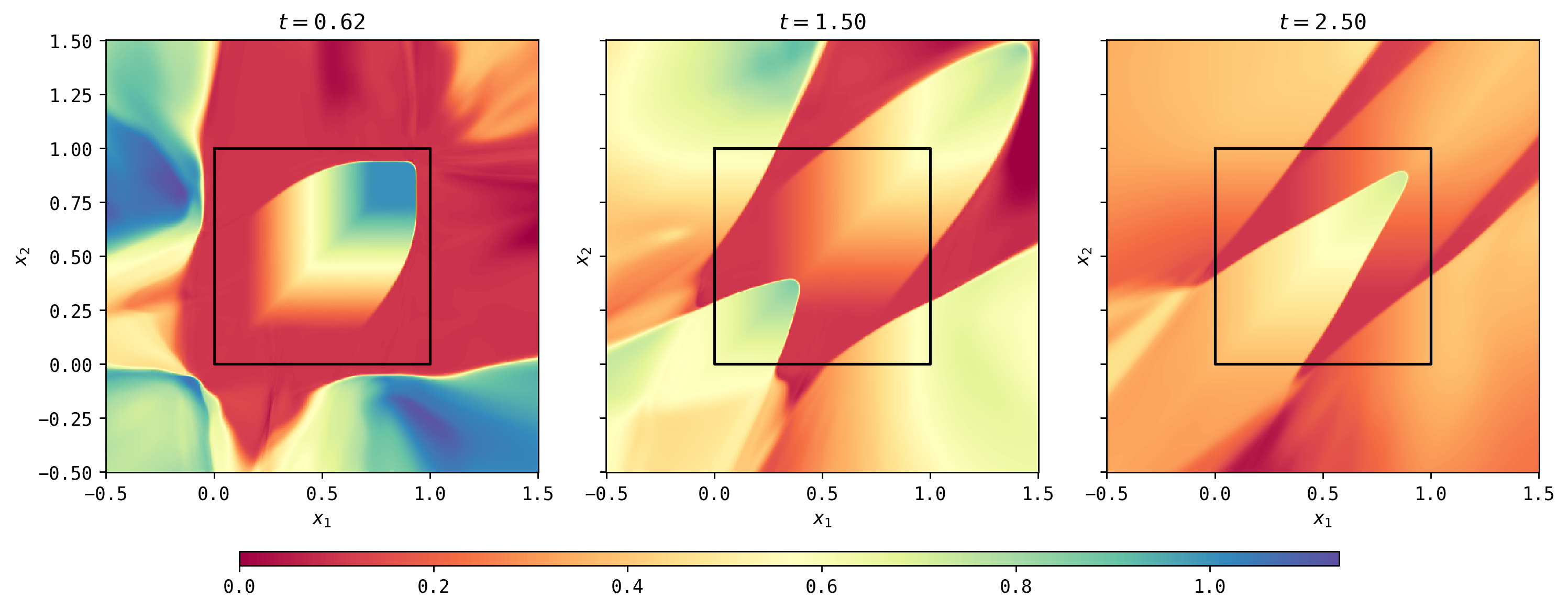}
  \end{tabular}
  \caption{Evaluation of LRNR reconstruction of the 2d acoustics (top) and 2d Burgers
  (bottom) equations with periodic boundary conditions, evaluated outside the square
  domain on which the training data was provided.}
  \label{fig:boundary}
\end{figure}

Focusing on the 2d acoustics equations (Fig.~\ref{fig:boundary}, top), we
observe that at time $t = 0.19$, the compactly supported initial condition is
extended beyond the spatial domain in a pseudo-periodic manner. While the
extension is not exact, the extrapolated waveforms propagate in a fashion
similar to the interior solution, which is accurately represented. As these
extrapolated waves enter the domain, any initial inaccuracies are corrected near
the boundary: although the waveforms appear slightly distorted outside, they
recover physically plausible shapes upon entry.

At time $t = 0.38$, the outgoing waves continue to retain their structure for
some distance, although distortion increases as they propagate
further from the domain. By time $t = 0.64$ (near the final time $t = 0.75$),
the LRNR no longer maintains coherent incoming waveforms outside the domain. For
instance, the solution just beyond the lower boundary shows no discernible wave
features. Similarly, the upper-left and upper-right corners display little
recognizable structure. This behavior aligns with the hyperbolic wave
perspective, where the LRNR appears to implicitly respect the domain of
influence by maintaining accurate waveforms only to the extent that they impact
the solution within the spatial domain.

\subsection{2d Burgers equation}

Similar observations hold for the 2d Burgers' equation. However, a key
distinction from the acoustics case is that the solution here propagates
strictly along characteristic curves. This difference is apparent already at the
initial time $t = 0.62$: in the upper-right corner outside the square domain, no
physically meaningful profiles are present, and the extrapolated behavior
resembles what is typically observed when INRs are queried outside their
training domain. We hypothesize that this is due to the learned directionality
of wave propagation. The trained LRNR appears to have implicitly learned the
characteristics of the PDE and thus omits wave structures in regions that lie
outside the domain of influence, i.e., regions from which no characteristics
reach the interior of the domain.

In contrast, in other directions, the LRNR places extrapolated structures that
resemble severely distorted periodic extensions of the solution inside the
domain. At time $t = 1.50$, these distorted features propagate along the general
direction of wave motion and gradually morph into accurate wave profiles as they
enter the domain. Between these propagating structures, we observe regions
containing irregularities or “wrinkles,” which seem to be occluded once they
are inside the domain.

Meanwhile, the outgoing shock retains its general form as it leaves the domain;
however, its tip becomes smoothed, deviating from the physically expected sharp
profile and rendering it unphysical. By $t = 2.50$, the intruding wavefronts at
the upper-left and lower-right parts of the domain have evolved into straight
rarefaction slopes that continue outside the spatial domain. As the domain of
influence contracts over time, the LRNR representation outside the domain
exhibits increasingly random features, consistent with the loss of physical
relevance in those regions.

Overall, the LRNR reconstruction appears to implicitly respect the finite
propagation speed and the domain of influence characteristic of hyperbolic
systems.

\section{Conclusion} 

This work introduced the Low-Rank Neural Representation (LRNR), a novel
architecture that enables efficient, interpretable, and physically meaningful
modeling of hyperbolic wave phenomena. By imposing a layer-wise low-rank
structure and leveraging a hypernetwork to capture temporal dynamics, LRNRs
achieve compact representations while preserving the rich behaviors of nonlinear
wave interactions, including shocks and rarefactions. We demonstrated that these
representations not only interpolate well but also extrapolate meaningfully
along learned hypermodes, revealing dominant modes of variation even far from
the training data. The compressed FastLRNR variant further enables rapid evaluation
and backpropagation, making it practical for use in real-time and
PDE-constrained settings. Additionally, we showed that LRNRs implicitly respect
causality and  handle domain boundaries through the learned propagation
dynamics.  These findings suggest that LRNRs provide a compelling
architecture for scientific machine learning, particularly for systems where
structure, interpretability, and computational efficiency are required.

\section{Acknowledgements} 

The authors acknowledge the Research Infrastructure Services
(RIS) group at Washington University in St. Louis for providing computational
resources and services needed to generate the research results delivered within
this paper (URL: \href{https://ris.wustl.edu}{\texttt{https://ris.wustl.edu}}).

K. Lee acknowledges support from the U.S. National Science Foundation under
grant IIS 2338909.  N. Park was supported by Samsung Research Funding \&
Incubation Center of Samsung Electronics under Project Number SRFC-IT2402-08,
and by Samsung Electronics Co., Ltd. (No. G01240136, KAIST Semiconductor
Research Fund (2nd)).  Part of this work was done while D. Rim was enjoying the
hospitality of University of Washington in Seattle as a Visiting Scholar.
He thanks R. J. LeVeque, Jesse Chan, Kui Ren, and Gunther Uhlmann for helpful
discussions.

\appendix

\section{Planar wave decompositions} \label{sec:planewave}

Consider the free space wave equation in $d$ dimensions, which seeks $u: \RR^d
\times \RR_+ \to \RR$ that satisfies
\begin{equation} \label{eq:freespace}
  \left\{
  \begin{aligned}
    (\partial_{tt} - c^2 \Delta) u &= 0
    & \text{ in } \RR^d \times \RR_+,
    \\
    u(\cdot, 0 ) &=  u_0,
    \\
    \partial_t u(\cdot, 0 ) &=  v_0.
  \end{aligned}
  \right.
\end{equation}
We decompose the problem into two parts, one with zero initial velocity and the
other zero initial value, which reads
\begin{equation}
  \left\{
  \begin{aligned}
    (\partial_{tt} - c^2 \Delta) u &= 0,
    \\
    u(\cdot, 0 ) &=  u_0,
    \\
    \partial_t u(\cdot, 0 ) &=  0,
  \end{aligned}
  \right.
  \qquad \qquad
  \left\{
  \begin{aligned}
    (\partial_{tt} - c^2 \Delta) u &= 0,
    \\
    u(\cdot, 0 ) &=  0,
    \\
    \partial_t u(\cdot, 0 ) &=  v_0,
  \end{aligned}
  \right.
\end{equation}
then denote the solutions to the two problems by $u^{(1)}$ and $u^{(2)}$,
respectively, with $u = u^{(1)} + u^{(2)}$. Now, consider the planar initial
conditions
\begin{equation}
  u_0(\bx) = g(\bw \cdot \bx),
  \qquad
  v_0(\bx) = h(\tilde{\bw} \cdot \bx),
  \qquad
  \bw, \tilde{\bw} \in S^{d-1},
\end{equation}
for some 1d functions $g, h: \RR \to \RR$.
Letting the solution ansatz be
\begin{equation}
  u^{(1)}(\bx, t) = G(\bw \cdot \bx, t),
  \qquad
  u^{(2)}(\bx, t) = H(\tilde{\bw} \cdot \bx, t),
\end{equation}
one reduces the wave equation to their 1d analogues due to the 
intertwining relation \eqref{eq:intertwine},
\begin{equation}
  \left\{
  \begin{aligned}
    (\partial_{tt} - c^2 \partial_{\beta\beta})G &= 0,
    \\
    G(\cdot, 0 ) &=  g,
    \\
    \partial_t G(\cdot, 0 ) &=  0,
  \end{aligned}
  \right.
  \qquad \qquad
  \left\{
  \begin{aligned}
    (\partial_{tt} - c^2 \partial_{\beta\beta}) H &= 0,
    \\
    H(\cdot, 0 ) &=  0,
    \\
    \partial_t H(\cdot, 0 ) &=  h.
  \end{aligned}
  \right.
\end{equation}
The solution to these 1d problems are given by the d'Alembert solution \eqref{eq:dalembert},
\begin{equation} \label{eq:planar_dalembert}
  \begin{aligned}
  u^{(1)} (\bx, t)
  &=
  \half
  \left(
    g(\bw \cdot \bx + ct) + g(\bw \cdot \bx - ct)
  \right),
  \\
  u^{(2)} (\bx, t)
  &=
  \frac{1}{2c}
  \int_{\tilde{\bw} \cdot \bx - ct}^{\tilde{\bw} \cdot \bx + ct}
  h(\beta) \, \dee \beta
  =
  \frac{1}{2c}
  \left(
    \tilde{h}(\tilde{\bw} \cdot \bx + ct) 
  - \tilde{h}(\tilde{\bw} \cdot \bx - ct)
  \right),
  \end{aligned}
\end{equation}
where $\tilde{h}(\beta) := \int_0^\beta h(\zeta) \dee \zeta$ denotes
the antiderivative. Then, the solution to the original free space problem
\eqref{eq:freespace} is the sum $u = u^{(1)} + u^{(2)}$.
\vskip6pt

\noindent \emph{Superposition of planar waves.}
In the case the initial condition was given by the superposition
\begin{equation}
  u_0(\bx) = \sum_{i=1}^\infty g_i(\bw_i \cdot \bx),
  \qquad
  v_0(\bx) = \sum_{i=1}^\infty h_i(\tilde{\bw}_i \cdot \bx),
  \qquad
  \bw_i, \tilde{\bw}_i \in S^{d-1},
\end{equation}
the solution is obtained by applying the solution \eqref{eq:planar_dalembert}
termwise,
\begin{equation}
  \begin{aligned}
  u (\bx, t)
  &=
  \half
  \sum_{i=1}^\infty
  \left(
    g_i(\bw_i \cdot \bx + ct) + g_i(\bw_i \cdot \bx - ct)
  \right)
  \\
  &+
  \frac{1}{2c}
  \sum_{i=1}^\infty
  \left(
    \tilde{h}_i(\tilde{\bw}_i \cdot \bx + ct) - \tilde{h}_i(\tilde{\bw}_i \cdot \bx - ct)
  \right),
  \end{aligned}
\end{equation}
with $\tilde{h}_i(\beta) := \int_0^\beta h_i (\zeta) \dee \zeta$. 

More generally, we can consider the planar waves that are continuously
parametrized.  That is, consider the functions $g$ and $h$ indexed by the real
tuple $(a, b)$, of the form
\begin{equation}
  g_{a, b}(\bw\cdot \bx) = a \sigma(\bw \cdot \bx + b),
  \qquad
  h_{a, b}(\bw\cdot \bx) = a \sigma'(\bw \cdot \bx + b),
\end{equation}
where $\sigma$ and $\sigma'$ are the ReLU function and its derivative,
respectively, and the initial conditions can be written as the integrals
\begin{equation}
  \begin{aligned}
  u_0 (\bx, t)
  &=
  \int g_{a, b}(\bw \cdot \bx) \, \dee \mu( \da, \dee \bw, \db)
  =
  \int a \sigma(\bw \cdot \bx + b) \, \dee \mu( \da, \dee \bw, \db),
  \\
  v_0 (\bx, t)
  &=
  \int h_{a,b}(\bw \cdot \bx) \, \dee \nu( \da, \dee \bw, \db)
  =
  \int a \sigma'(\bw \cdot \bx + b) \, \dee \nu( \da, \dee \bw, \db),
  \end{aligned}
\end{equation}
for given probability measures $\mu$ and $\nu$ over $\RR \times S^{d-1} \times
\RR$.  The wave solution is then 
\begin{equation}
  \begin{aligned}
  u (\bx, t)
  &=
  \half
  \int
  a
  \left[
    \sigma (\bw \cdot \bx + b + ct) + \sigma (\bw \cdot \bx + b- ct)
  \right]
  \dee \mu( \da, \dee \bw, \db)
  \\
  &+
  \frac{1}{2c}
  \int
  a
  \left[
    \sigma (\bw \cdot \bx + b + ct) 
    - \sigma (\bw \cdot \bx + b- ct)
  \right]
  \dee \nu( \da, \dee \bw, \db).
  \end{aligned}
\end{equation}

\end{document}